\documentclass[acmtog,nonacm]{acmart}

\usepackage{multirow}
\usepackage{multicol}
\usepackage{colortbl}  
\usepackage{hhline} 
\usepackage{gensymb}
\usepackage{subcaption}

\definecolor{best}{rgb}{0.96, 0.57, 0.58}
\definecolor{second}{rgb}{0.98, 0.78, 0.57}

\AtBeginDocument{%
  }

\setcopyright{acmlicensed}
\copyrightyear{2025}
\acmYear{2025}
\acmDOI{XXXXXXX.XXXXXXX}

\begin{document}

\title{Dens3R: A Foundation Model for 3D Geometry Prediction}

\author{Xianze Fang}
\affiliation{%
  \institution{Alibaba Group}
  \country{China}
}
\authornote{Equal Contribution}
\email{fangxianze.fxz@taobao.com}

\author{Jingnan Gao}
\affiliation{%
  \institution{Shanghai Jiao Tong University}
  \country{China}
}
\authornotemark[1]
\email{gjn0310@sjtu.edu.cn}

\author{Zhe Wang}
\affiliation{%
  \institution{Alibaba Group}
  \country{China}
}
\email{xingmi.wz@taobao.com}

\author{Zhuo Chen}
\affiliation{%
  \institution{Shanghai Jiao Tong University}
  \country{China}
}
\email{ningci5252@sjtu.edu.cn}

\author{Xingyu Ren}
\affiliation{%
  \institution{Shanghai Jiao Tong University}
  \country{China}
}
\email{rxy_sjtu@sjtu.edu.cn}

\author{Jiangjing Lv}
\affiliation{%
  \institution{Alibaba Group}
  \country{China}
}
\authornote{Project Leader}
\email{jiangjing.ljj@taobao.com}

\author{Qiaomu Ren}
\affiliation{%
  \institution{Alibaba Group}
  \country{China}
}
\email{renqiaomu.rqm@taobao.com}

\author{Zhonglei Yang}
\affiliation{%
  \institution{Alibaba Group}
  \country{China}
}
\email{zhonglei.yzl@taobao.com}

\author{Xiaokang Yang}
\affiliation{%
  \institution{Shanghai Jiao Tong University}
  \country{China}
}
\email{xkyang@sjtu.edu.cn}

\author{Yichao Yan}
\affiliation{%
  \institution{Shanghai Jiao Tong University}
  \country{China}
}
\authornote{Corresponding Author}
\email{yanyichao@sjtu.edu.cn}

\author{Chengfei Lv}
\affiliation{%
  \institution{Alibaba Group}
  \country{China}
}
\email{chengfei.lcf@taobao.com}




\newcommand{\xy}[1]{{\color{red}#1}}

\begin{abstract}
Recent advances in dense 3D reconstruction have led to significant progress, yet achieving accurate unified geometric prediction remains a major challenge. Most existing methods are limited to predicting a single geometry quantity from input images. However, geometric quantities such as depth, surface normals, and point maps are inherently correlated, and estimating them in isolation often fails to ensure consistency, thereby limiting both accuracy and practical applicability. This motivates us to explore a unified framework that explicitly models the structural coupling among different geometric properties to enable joint regression. In this paper, we present Dens3R, a 3D foundation model designed for joint geometric dense prediction and adaptable to a wide range of downstream tasks. Dens3R adopts a two-stage training framework to progressively build a pointmap representation that is both generalizable and intrinsically invariant. Specifically, we design a lightweight shared encoder-decoder backbone and introduce position-interpolated rotary positional encoding to maintain expressive power while enhancing robustness to high-resolution inputs. By integrating image-pair matching features with intrinsic invariance modeling, Dens3R accurately regresses multiple geometric quantities such as surface normals and depth, achieving consistent geometry perception from single-view to multi-view inputs. Additionally, we propose a post-processing pipeline that supports geometrically consistent multi-view inference. Extensive experiments demonstrate the superior performance of Dens3R across various dense 3D prediction tasks and highlight its potential for broader applications.
\end{abstract}



\begin{teaserfigure}
  \includegraphics[width=\linewidth]{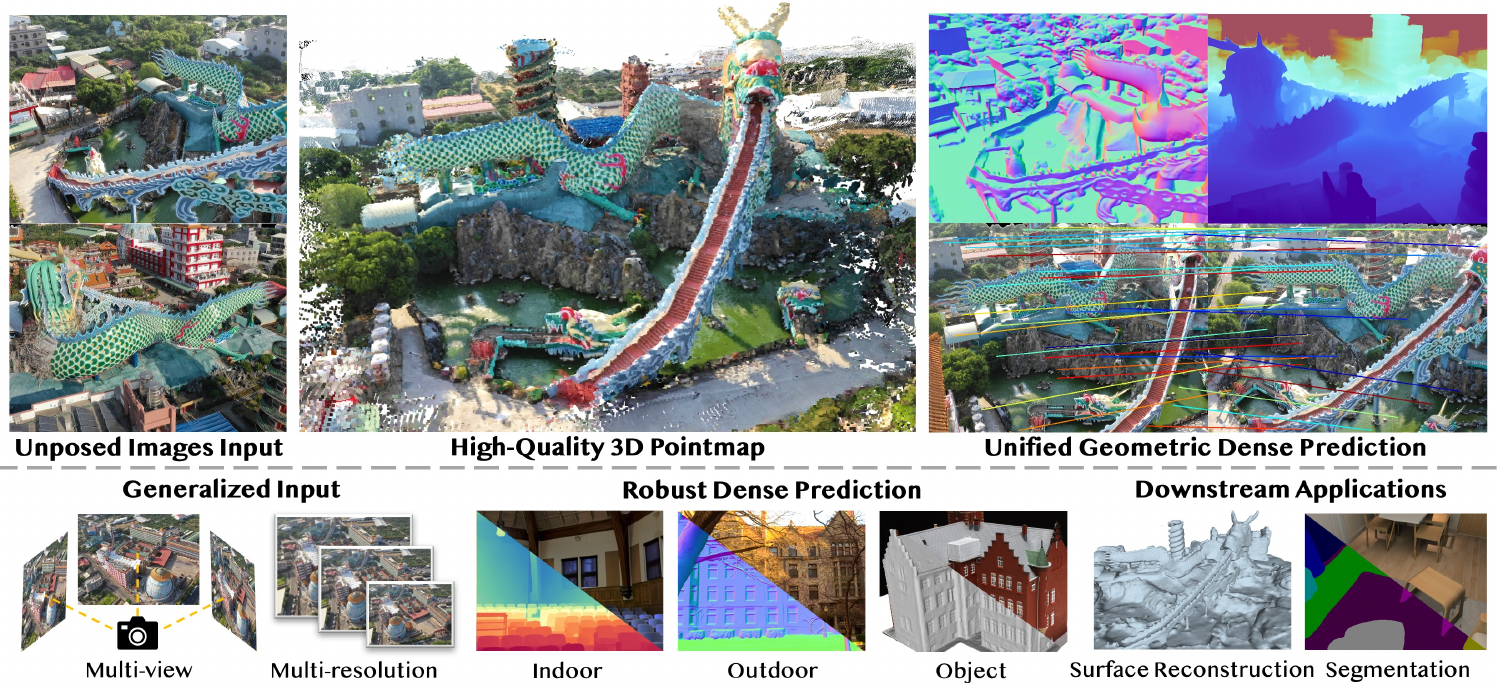}
  \caption{Dens3R is a feed-forward visual foundation model that takes unposed images as input and outputs high-quality 3D pointmap with unified geometric dense prediction. Dens3R also accepts generalized inputs, supporting both multi-view and multi-resolution inputs. As a versatile backbone, Dens3R achieves robust dense prediction under several scenarios and can be easily extended to downstream applications.}
  \label{fig:teaser}
\end{teaserfigure}


\maketitle

\section{Introduction}
Recovering 3D geometric structures from static images is a long-standing and fundamental problem in computer vision.
Classical approaches, such as Structure-from-Motion (SfM) and Multi-View Stereo (MVS), demonstrate strong performance in controlled settings and have been widely adopted in a broad range of 3D reconstruction applications. However, in unconstrained scenarios—where camera intrinsics, extrinsics, or viewpoint information are unavailable—achieving accurate and dense geometric prediction remains highly challenging. These conditions demand more generalizable and robust solutions capable of handling diverse and unstructured visual inputs.

Existing methods for dense geometric prediction primarily fall into two categories. The first category mostly adopts generative models, utilizing strong image priors from pre-trained diffusion models or large-scale training datasets for dense prediction. For example, GenPercept~\cite{genpercept} is used for depth prediction, and StableNormal~\cite{stablenormal} for normal estimation. This raises a key issue: while image generation tasks typically benefit from their inherent ambiguity and multi-modal output characteristics, geometric prediction is fundamentally different. Geometric prediction is essentially a deterministic task that needs to closely reflect the structural information of the underlying scene. Moreover, the pixel continuity and spatial smoothness required by geometric representations are difficult to naturally obtain through standard diffusion sampling mechanisms without structural constraints. Therefore, the direct application of diffusion models in geometric regression tasks faces significant challenges, especially in such tasks where a strict one-to-one correspondence between input and output needs to be maintained. Based on this, we adopt a regression-oriented framework to construct geometric mapping models in a more efficient and interpretable way. Furthermore, the aforementioned methods mainly handle only one geometric quantity prediction and cannot generalize to output multiple geometric quantities in a single forward pass. The second category includes DUSt3R~\cite{dust3r} and its follow-up works~\cite{mast3r,moge,vggt}. These methods use regression models that can regress 3D point map representations with geometric properties, applied to dense prediction, including image pair matching and depth estimation. However, these methods typically focus on a single prediction task, and other geometric quantities suffer severe performance degradation due to representation influences.

This raises a natural question: can we build a unified model that simultaneously regresses multiple geometric quantities with high quality? We observe that existing methods like DUSt3R, when handling dense geometric regression tasks, overlook a crucial geometric information—surface normals. Traditionally, normals have been used to add high-frequency details to rough geometric structures to enhance rendering quality. However, our research finds that introducing normal information during geometric prediction can significantly improve the accuracy of point maps, resulting in more detailed and structurally consistent 3D representations. This is mainly because:
1) From the perspective of normal prediction, the inherent image pair matching capability in dense vision backbone networks helps alleviate monocular ambiguity and improve the stability and accuracy of normal prediction;
2) From the feature modeling perspective, normals possess good intrinsic invariance, which simplifies the mapping learning process and aids in model convergence and generalization.
This modeling approach enables the model to simultaneously predict multiple geometric quantities (such as depth, normals, and point maps) from a single view, effectively reducing dependence on multi-view supervision and simplifying the training process.
However, training such a multi-task, multi-output 3D foundation model still faces significant challenges. Geometric quantities are tightly coupled, and how to coordinate these relationships to achieve optimal overall performance requires carefully designed training strategies and architectural support.

In this paper, we present \textbf{Dens3R}, a foundation model for high-quality geometric prediction. To this end, we design a two-stage training framework that gradually builds a versatile pointmap representation, which generalizes well to various downstream tasks.
Specifically, we first construct a dense vision backbone network with multi-task prediction capabilities. This network adopts a shared encoder-decoder architecture, which significantly reduces model parameters while maintaining expressive power. To accommodate high-resolution inputs, we introduce position-interpolated rotary positional encoding, which effectively mitigates prediction degradation caused by increased input resolution.
For the training strategy, we propose a novel two-staged approach. In the first stage, the model leverages image pair matching features to learn scale-invariant point maps, capturing consistent spatial geometric structures across viewpoints. Subsequently, to fully exploit the one-to-one mapping property in normal estimation, we extend the pointmap representation into an intrinsically invariant form. This allows the model to independently attend to each viewpoint, thereby improving the accuracy of normal prediction. The learned geometric structures also assist in estimating other geometric quantities, such as depth, thereby simplifying their training processes.
Finally, we design a simple and efficient post-processing pipeline that supports multi-view inputs during inference, which enhances the geometric consistency of the model in real-world applications.
In summary, we make the following contributions:
\begin{itemize}
    \item We introduce \textbf{Dens3R}, a dense 3D visual foundation model that demonstrates \textbf{high-quality performance} in various 3D tasks including pointmap reconstruction, depth estimation, normal prediction and image matching under several benchmark evaluations.
    \item We design a novel training strategy with the \textbf{intrinsic-invariant pointmap} and shared Encoder-Decoder visual backbone to incorporate surface normals in unconstrained image-based dense 3D reconstruction, simplifying the training complexity of other 3D quantities and achieving better results without requiring excessive computation resources.
   \item We employ a \textbf{position-interpolated rotary positional encoding} to preserve prediction accuracy at higher resolutions and support multi-resolution inputs.
    \item Extensive experiments on various benchmarks showcase our high-quality predictions of 3D geometric quantities, which further enable a wide range of applications.
\end{itemize}

\section{Related Works}
\subsection{Monocular depth and normal prediction}
Monocular depth prediction has been extensively investigated and demonstrates strong capability in providing geometric priors for a multitude of downstream tasks like image understanding and 3D reconstruction. 
The earliest pioneering researchers~\cite{DBLP:conf/cvpr/BhatAW21,DBLP:journals/corr/abs-2302-12288,DBLP:conf/nips/EigenPF14,DBLP:conf/iccv/000600CYWCS23,DBLP:journals/pami/HuYZCLCWYSS24,DBLP:conf/cvpr/PiccinelliYSSLG24} addressed this issue by estimating depth with a metric scale. These methods usually rely heavily on data from specific sensors, which restricts the applicability and deteriorates the performance when confronted with complex scenes. Subsequently, deep learning approaches involve predicting relative depth either through direct regression~\cite{DBLP:conf/nips/ChenFYD16,DBLP:conf/cvpr/ChenQFKHD20,DBLP:conf/iccv/GodardAFB19,DBLP:conf/cvpr/LiS18,DBLP:journals/pami/RanftlLHSK22,DBLP:conf/cvpr/YangKHXFZ24,depth_anything_v2} or via generative modeling based on diffusion priors~\cite{geowizard,DBLP:journals/corr/abs-2403-13788,DBLP:conf/cvpr/KeOHMDS24,DBLP:journals/corr/abs-2312-14733}.
While monocular depth estimation has made significant strides, accurate 3D shape reconstruction from depth maps remains fundamentally dependent on precise camera intrinsic parameters. 
\begin{figure*}
    \centering
    \includegraphics[width=\linewidth]{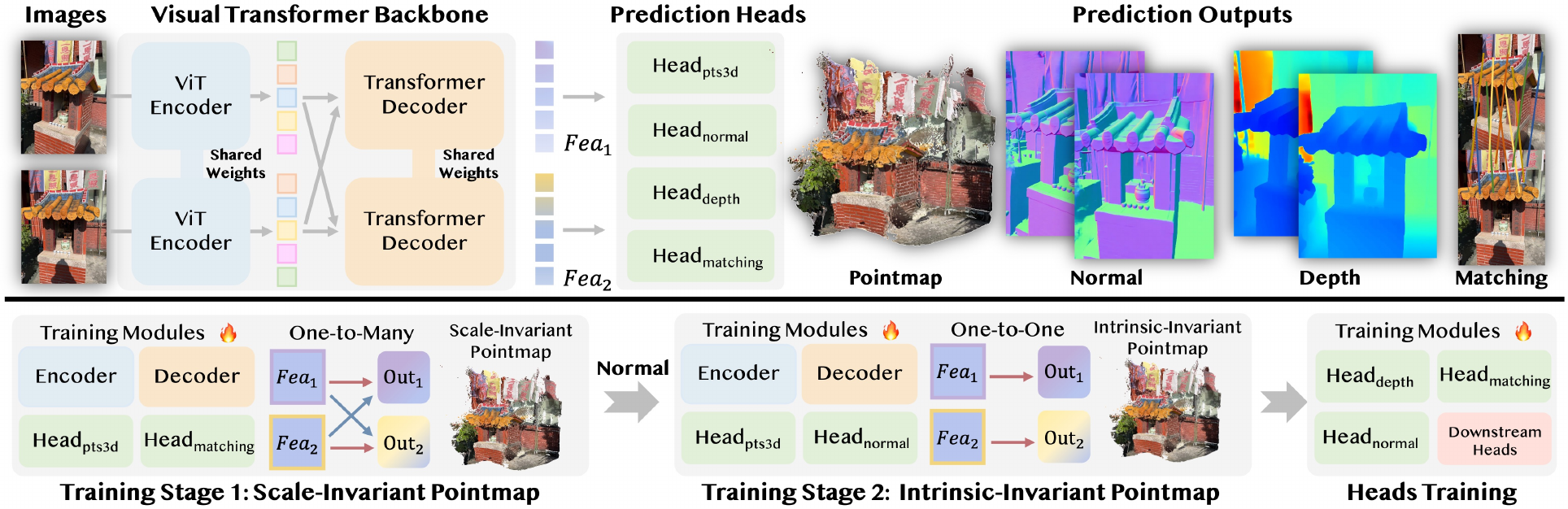}
    \caption{Overview of Dens3R. We propose Dens3R, a dense visual transformer backbone featuring a shared encoder-decoder architecture and multiple task-specific heads for geometric prediction. To train this foundation model, we adopt a two-stage strategy. In Stage 1, we learn a scale-invariant pointmap by enforcing cross-view mapping consistency across multiple viewpoints. In Stage 2, we incorporate surface normals and leverage one-to-one correspondence constraints to transform the representation into an intrinsic-invariant pointmap. Built upon this unified backbone, additional geometric prediction heads and downstream task branches can be seamlessly integrated to support a wide range of applications.}
    \label{fig:Framework}
\end{figure*}
Meanwhile, normal maps serve as a supervision for neural scene representation, bridging 2D and 3D worlds. The accurate estimation of the normal map can open up broader applications like material decomposition and relighting. On one hand, regression-based methods~\cite{DBLP:conf/iccv/EftekharSMZ21,DBLP:conf/cvpr/BansalRG16,DBLP:conf/cvpr/WangFG15,DBLP:conf/iccv/RanftlBK21} utilize large-scale training datasets for robust estimation. DSINE~\cite{DBLP:conf/cvpr/BaeD24} proposes to
leverage the per-pixel ray direction and try to model the inductive biases for surface normal estimation correctly. On the other hand, diffusion-based methods~\cite{DBLP:conf/cvpr/LongGLLDLMZHTW24,geowizard,stablenormal} adapt the pretrained diffusion model as a geometric cues predictor. Geowizard includes a geometry switcher to disentangle mixed-sourced data into distinct sub-distributions for normal prediction. StableNormal repurposes the diffusion model for deterministic estimation tasks and can estimate sharp normals steadily. Nevertheless, these normal estimation methods often suffer from monocular ambiguity, leading to inaccurate and inconsistent results for complex scenes. In contrast, our method allows the communication between 3D geometric representation and normal prediction without known camera poses. This not only resolves the ambiguity but also achieves accurate 3D reconstruction with accurate normals.

\subsection{Image Pair Matching in 3D}
Dense matching~\cite{DBLP:conf/cvpr/EdstedtAWF23,DBLP:conf/cvpr/EdstedtSBWF24,DBLP:conf/cvpr/EfeIA21,DBLP:conf/wacv/MelekhovTSPRK19,DBLP:conf/cvpr/TruongDT20,DBLP:conf/cvpr/TruongDGT21,DBLP:journals/pami/TruongDTG23,DBLP:conf/cvpr/Zhu023a,DBLP:conf/cvpr/SarlinDMR20,DBLP:conf/cvpr/SunSWBZ21} has been proved to be effective in many scenarios and results in top performance in many benchmarks. However, these approaches cast matching as a 2D problem, which restricts the application for visual localization. Thus anchoring image correspondence in 3D space is essential when these 2D-based methods fall short. Early methods~\cite{DBLP:conf/cvpr/BhalgatHZ23,DBLP:conf/cvpr/HeYFY20a,DBLP:conf/eccv/WangZHS20,DBLP:conf/iccv/YaoJP19,DBLP:conf/cvpr/YifanDACZ22,DBLP:conf/cvpr/ZhouSL21,DBLP:conf/eccv/ToftTSKB20} leverage epipolar constraints in order to improve accuracy or robustness. Recently, researchers~\cite{DBLP:conf/iclr/0001LKYRT24,DBLP:conf/iccv/Wang0N23} leverage diffusion models for pose estimation and demonstrate promising results by incorporating 3D geometric constraints into estimation formulation. MASt3R~\cite{mast3r} retrieves correspondences via 3D reconstruction from uncalibrated images by explicitly training local features for pairwise matching. However, MASt3R only grounds image-pair matching and overlooks other geometric predictions like depth and normal, while our method achieves unified geometric predictions and better matching.

\subsection{Dense Unconstrained Geometric Representations}
Neural scene reconstructions~\cite{nerf,neus,3dgs,mipnerf,Martin-BruallaR21,Barron_2023_ICCV,DBLP:conf/nips/YarivGKL21,scaffoldgs,Yu2023MipSplatting,neus2} usually require the camera intrinsic parameters and poses for optimization. The reconstruction quality of these methods is highly dependent on the accuracy of the camera intrinsics and poses. Later methods~\cite{smart2024splatt3r,noposplat,coponerf} propose to optimize the scene without known camera poses, but these methods usually take longer time and sacrifice reconstruction quality. To bypass estimation of camera parameters and poses, DUSt3R~\cite{dust3r} proposes to directly map two input images in a single forward pass, leading to a more straightforward geometry representation. Subsequently, Spann3R~\cite{wang2024spann3r} and Fast3R~\cite{fast3r} augment DUSt3R to process an ordered set of images. MoGe~\cite{moge} further proposes affine-invariant pointmaps for monocular geometry estimation. VGGT~\cite{vggt} utilizes 3D pointmaps and multiple prediction heads to predict geometric quantities from multi-view images input. However, the aforementioned methods overlook the normal attribute and fall short in prediction for complex scenarios. In contrast, our model takes advantage of pointmap representation and employs several prediction heads including the normal head to achieve unified geometric predictions.

\section{Method}
This work aims to utilize a single model to predict various geometric data from unconstrained images, including 3D pointmaps, depth maps, normal maps, and image-pair matching. To this end, we built a backbone network based on dense visual transformers and designed input configurations that adapt to multi-resolution and multi-view requirements (Sec.~\ref{sec:backbone}). Since achieving accurate results through direct training with a single model is challenging, we adopted a two-stage training approach. In the first stage, we train the backbone and heads to obtain scale-invariant pointmaps. In the second stage, we fine-tune the backbone on this foundation to obtain intrinsic-invariant pointmaps (Sec.~\ref{sec:training}). Finally, we further fine-tune the prediction heads for each downstream task to adapt to different application scenarios. Meanwhile, extending the model inputs to multi-view images in the inference stage significantly improves the overall inference quality. (Sec.~\ref{sec:inference}).

\subsection{Model Formulation}
\label{sec:backbone}
\noindent\textbf{Shared Backbone.} 
Motivated by recent advances in 3D vision~\cite{dust3r,mast3r,vggt,lvsm}, we aim to build a foundation model capable of predicting diverse geometric quantities across different scenes and tasks. To this end, we adopt a dense visual transformer as the backbone, learning from rich 3D annotated data.
Given an image pair of image sequence $(I_i)^2_{i=1} \in \mathcal{R}^{3\times H\times W}$, Dens3R's dense visual transformer is a function $f$ that maps the input to a corresponding set of 3D quantities per frame:
\begin{equation}
     (C_i, P_i, D_i, N_i, M_i)_{i=1}^2 = f((I_i)^2_{i=1}),
\end{equation}
where $C_i \in \mathcal{R}^9 $ is the camera parameters including both intrinsics and extrinsics, $D_i \in \mathcal{R}^{H \times W}$ is the depth map, $N_i \in \mathcal{R}^{3 \times H \times W}$ is the normal map, and $M_i \in \mathcal{R}^{C\times H \times W}$ is the image-pair-matching with $C$-dimensional features. 

The overall architecture is illustrated in the upper part of Fig.~\ref{fig:Framework}. Similar to prior DUSt3R-based approaches~\cite{dust3r,mast3r,vggt,moge}, we first employ a shared-weight encoder to process input image sequences and extract image features $Fea_{i}$, which are then fed into the decoder. Unlike previous works, our approach introduces a novel weight-sharing mechanism within the decoders, allowing the backbone to better capture spatial relationships across viewpoints and to model the holistic 3D scene structure. Given the need to predict a wider range of geometric outputs, this design also significantly reduces memory and computational overhead, keeping the training and inference efficient. Moreover, the shared-weight strategy facilitates high-resolution input processing while effectively preventing memory overflow.

\noindent\textbf{Multi-resolution Input.}
Existing methods represented by DUSt3R perform excellently at fixed resolutions (such as 512), but their prediction accuracy significantly decreases when processing higher-resolution inputs. The main challenge for this issue lies in the rotary positional encoding (RoPE) used in their ViT structure, which becomes unstable when inferring images beyond the training resolution range.  Inspired by context window extension techniques in LLMs~\cite{ropellm}, we incorporate the position-interpolated RoPE into the ViT as a simple yet effective improvement. We adapt the idea from context window to image resolution in the image domain, addressing the instability at higher resolutions. 
Considering the smooth properties of trigonometric functions in RoPE, interpolation is more stable than direct extrapolation when handling high resolutions. 
Specifically, let the original RoPE be $R$, the input sequence length be $L$, and for any RoPE embedding vector $x$, we obtain a new encoding representation $R'$ through interpolation. That is:
\begin{equation}
    R'(x,m) = R(x,\frac{mL}{L'}),
\end{equation}
where $m$ is the position index and $L'$ is the longer sequence.
This position-interpolation encoding strategy significantly enhances the model's robustness under high-resolution inputs, effectively avoiding the performance degradation caused by RoPE extrapolation.

\subsection{Foundation Model Training}
\label{sec:training}
The main challenge in training 3D geometric foundation models lies in the coupling among multiple prediction outputs, where mutual interference often leads to performance degradation. Existing methods typically focus on only one or two geometric tasks, resulting in poor generalization to others. To this end, we propose to build upon a unified geometric representation since all geometric representations are inherently interconvertible. We adopt a two-stage training paradigm, progressively learns a strong geometric prior, which can be efficiently transferred to a variety of 3D geometry prediction tasks via lightweight fine-tuning.

\noindent\textbf{Scale-Invariant Pointmap Training.}
In the first stage, we train the ViT backbone (Encoder-Decoder), pointmap head, and matching head to obtain a scale-invariant pointmap $P_i$.
Following MASt3R's \cite{mast3r}, we adopted (1) local 3D regression loss $\mathcal{L}_{\text {pts\_loc }}$, (2) Global 3D Regression Loss $\mathcal{L}_{\text {pts\_glb }}$, (3) Pointmap Normal Loss $\mathcal{L}_{\text {pts\_n}}$, (4) Pixel Matching Loss $\mathcal{L}_{\text {match}}$. The details are as follows:

\noindent\underline{(1) Local 3D Regression Loss $\mathcal{L}_{\text {pts\_loc }}$.} 
For a predicted camera, we use the local 3D regression loss to quantify the pointmap in its own coordinate frame. We apply a mask derived from the ground-truth data to the pointmap and only evaluate the valid points when calculating the loss. We also employ a normalization factor to handle the scale ambiguity between ground-truth and the predicted pointmaps. We set the factor $z_v$ as the average distance of all valid points in $v_{th}$ camera coordinate frame to the origin: 
\begin{equation}
\begin{aligned}
    &z_v=\left\| P^{1,v}_{masked} \right\| + \left\| P^{2,v}_{masked} \right\|,v\in\{1,2\},\\
    &\bar z_v=\left\| \bar P^{1,v}_{masked} \right\| + \left\| \bar P^{2,v}_{masked} \right\|,v\in\{1,2\},
\end{aligned}
\end{equation}
where $\bar z_v$ is the corresponding factor of the ground-truth. Then the local 3D regression loss can be formulated as:
\begin{equation}
    \mathcal{L}_{\text {pts\_loc }}=\left\|\frac{1}{z_v} P^{v, v}_{masked}-\frac{1}{\bar{z_v}} \bar{P}_{masked}^{v, v}\right\|,v\in \{1,2\},
\end{equation}
where $P^{n,m}$ denotes the pointmap from camera $n$ expressed in the coordinate frame of camera $m$.

\noindent\underline{(2) Global 3D Regression Loss $\mathcal{L}_{\text {pts\_glb }}$.}
The global 3D regression loss is applied to quantify the pointmap expressed in another camera's coordinate frame. This loss function simultaneously optimizes for two objectives. It not only constrains the network to fit the pointmap shape of the image, but also aligns the pointmap to another paired image. The global regression loss is formulated as:
\begin{equation}
    \mathcal{L}_{\text {pts\_glb }}=\left\|\frac{1}{z_t} P_{masked}^{v, t}-\frac{1}{\bar{z_t}} \bar{P}_{masked}^{v, t}\right\|,v,t\in \{1,2\},v\neq t,
\end{equation}
where $z_t$ and $\bar z_t$ is the normalization factor of the pointmap and the ground-truth.

\noindent\underline{(3) Pointmap Normal Loss $\mathcal{L}_{\text {pts\_n}}$.}
 To train an intrinsic-invariant pointmap from the scale-invariant pointmap, we use a pointmap normal loss to encourage the pointmap learn smooth surface and sharp edge, making the pointmap perceives the normal information and the intrinsic-invariant property. Suppose $N^{v,v}$ is the ground-truth view-space normal expressed in its own camera coordinate frame and  $N^{v,t}$ is the ground-truth normal expressed in another camera coordinate frame, the pointmap normal loss is the absolute error loss between the transformed normal and the ground-truth normal:
\begin{equation}
    \mathcal{L}_{\text {pts\_n}}=\mathcal{L}_1(N^{v,v},\hat{N}^{v,v})+\mathcal{L}_1(N^{v,t},\hat{N}^{v,t}),v,t\in \{1,2\},v\neq t,
\end{equation}
where the $\hat{N}^{v,v}$ is the normal transformed from the local pointmap and $\hat{N}^{v,t}$ is the normal transformed from the global pointmap. 

\noindent\underline{(4) Pixel Matching Loss $\mathcal{L}_{\text {match}}$.}
We utilize the pixel matching loss proposed in MASt3R~\cite{mast3r} to learn accurate image-matching. This loss is based on the infoNCE~\cite{Oord_Li_Vinyals_2018} loss and ensures that each pixel's descriptor in the first image match at most one pixel's descriptor in another image. Suppose $\hat{\mathcal{M}}=(i,j)$ is the set of ground-truth correspondences where the $i_{th}$ pixel in the first image matches the $j_{th}$ pixel in another, the loss can then be formulated as:
\begin{equation}
    \begin{aligned}
     \mathcal{L}_{\text {match }}=-\sum_{(i, j) \in \hat{\mathcal{M}}} \log &\frac{s_\tau(i, j)}{\sum_{k \in \mathcal{P}^1} s_\tau(k, j)}+\log \frac{s_\tau(i, j)}{\sum_{k \in \mathcal{P}^2} s_\tau(i, k)}, \\
    s_\tau(i, j)&=\exp \left[-\tau D_i^{1\top} D_j^2\right],
\end{aligned}
\end{equation}
where $\tau$ is a hyper-parameter, and $D_i$ and $D_j$ are the corresponding descriptors in each image.

\begin{figure}
    \centering
    \includegraphics[width=\linewidth]{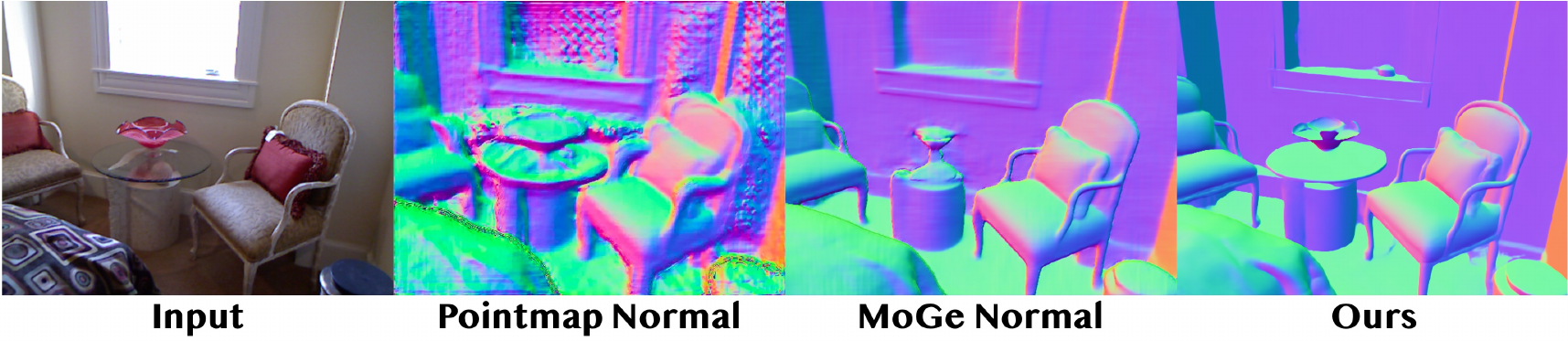}
    \caption{Normal comparison. We demonstrate that the normal derived directly from the scale-invariant pointmap and MoGe both are not accurate enough.}
    \label{fig:normal_exp}
    \vspace{-4mm}
\end{figure}

With the above losses, we summarize the training objective as:
\begin{equation}
\mathcal{L}_{stage1} = \mathcal{L}_{\text {pts\_loc }} + \eta_{1}\mathcal{L}_{\text {pts\_glb }} + \eta_{2}\mathcal{L}_{\text {pts\_n}} + \eta_{3}\mathcal{L}_{\text {match}},
\label{eq:stage1}
\end{equation}
where the loss weights $\eta_{1}$, $\eta_{2}$, and $\eta_{3}$ are set as $1.0$, $0.1$ and $0.075$, respectively.
After training, we obtained a scale-invariant pointmap capable of capturing rich spatial information. However, as shown in Fig.~\ref{fig:normal_exp}, the accuracy of normals obtained directly from the pointmap at this stage is still not ideal.

\noindent\textbf{Intrinsic-Invariant Pointmap Training.}
Although the point-based representation learned in the first stage achieves good performance, it remains limited in its ability to generalize to other tasks—particularly surface normal estimation. Existing methods often struggle with monocular ambiguity in normal prediction, leading to inaccurate and inconsistent results.

To this end, we expanded the pointmap representation in the second stage, proposing an \textbf{intrinsic-invariant pointmap}. This representation allows the model to form consistent geometric understanding of the same structure under different viewpoints, effectively improving the stability and generalization capability of normal estimation. Specifically, we introduce high-quality normal supervision based on the first stage's point map, and jointly fine-tune the encoder-decoder module, point map prediction head, and newly added normal prediction head to achieve end-to-end optimization. In terms of supervision mechanism, we adjusted the initial "one-to-many" mapping (one image corresponding to multiple view supervisions) to a "one-to-one" mapping, enabling the model to independently optimize normal prediction under a single viewpoint. This strategy not only significantly reduces the ambiguity brought by multi-view supervision but also simplifies the training process and improves training efficiency and stability. 

We observe that the commonly-used confidence loss in previous works~\cite{dust3r,mast3r,vggt} tends to cause models to \textbf{ignore complex scenarios} such as reflective surfaces and low textured areas. However, naively removing the loss without additional constraints leads to degraded performance, since previous models rely heavily on confidence weighting for point-view regression. In contrast, by utilizing the deterministic nature of normals, we obviate the need to rely on additional views, which further enables stable and accurate prediction.

For detailed implementation, we explicitly connect normal to the pointmap representation, that is
\begin{equation}
    P_i^n = P_i \oplus n,
\end{equation}
where $\oplus$ represents feature concatenation operation. The normal prediction head is connected after the initial point map training is completed, allowing the model to consistently output coherent normal mappings from the same input image, thereby internalizing this intrinsic invariance in the point map and maintaining geometric consistency across different views.

In the second stage, we add a normal loss $\mathcal{L}_{\text {n}}$ for finetuning.

\noindent\underline{(5) Predicted Normal Loss $\mathcal{L}_{\text {n}}$.}
Apart from the intrinsic-invariant pointmap, we also design a normal head to predict the view-space normal of each frame in input image pairs. We also use the $\mathcal{L}_1$ loss to supervise the normal prediction:
\begin{equation}
    \mathcal{L}_{\text {n}}=\mathcal{L}_1(N^{v,v},\bar{N}^{v,v}),v\in \{1,2\},
\end{equation}
where $N$ is the ground-truth normal and $\bar N$ is the direct prediction of the normal prediction head.
The complete training objective for training stage 2 is as follows:
\begin{equation}
\mathcal{L}_{stage2} = \mathcal{L}_{\text {pts\_loc }} + \lambda_{1}\mathcal{L}_{\text {pts\_glb }} + \lambda_{2}\mathcal{L}_{\text {pts\_n}} + \lambda_{3}\mathcal{L}_{\text {n}},
\label{eq:stage2}
\end{equation}
where the loss weights $\lambda_{1}$, $\lambda_{2}$, and $\lambda_{3}$ are set as $1.0$, $0.1$ and $1.0$, respectively.

To further improve the performance of Dens3R on high-resolution inputs, we introduce a coarse-to-fine training strategy. Specifically, we first fine-tune the model on 512 resolution images to establish a stable geometric prior, and then fine-tune it on 1024 resolution images to further improve the prediction accuracy. In addition, combining high-resolution data also significantly improves the fidelity of point-based representations, ultimately enhancing the overall quality of dense 3D prediction.

\begin{figure*}
    \centering
    \includegraphics[width=\linewidth]{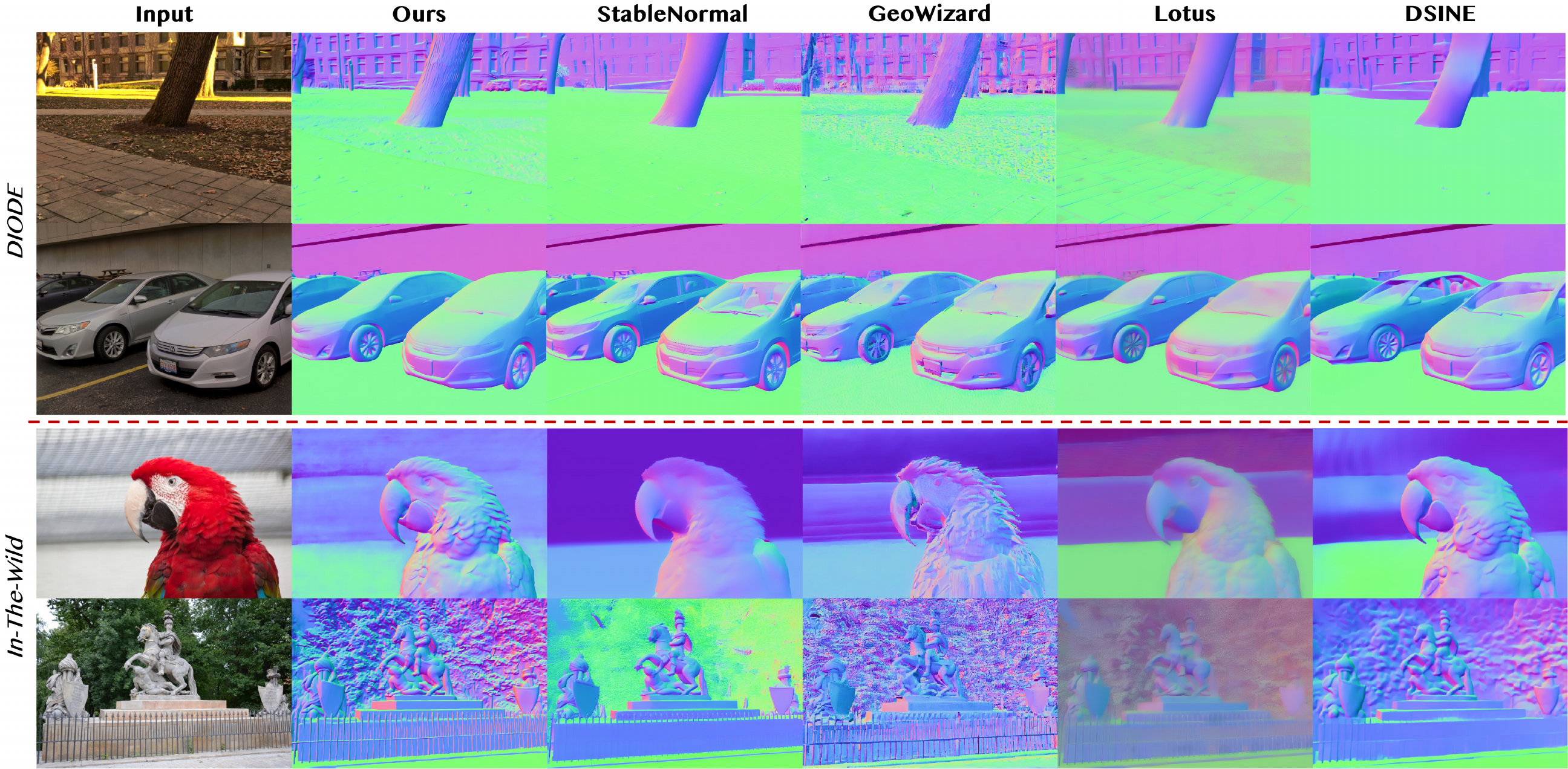}
    \caption{Qualitative comparison of normal prediction. Dens3R generates more accurate and detailed normal maps than previous methods. Our method is capable of predicting accurate normals for reflective surfaces and in backgrounds. For both object-centric and unbounded scenes, our method also accomplishes to predict stable and intricate surface normals.}
    \label{fig:normal}
\end{figure*}

\subsection{Model Inference}
\label{sec:inference}
\noindent\textbf{Heads Training.}
After training, we fine-tune it for downstream tasks by adding task-specific prediction heads on top of the frozen backbone network. Training only these new heads enables extension to various tasks such as depth estimation, normal estimation, matching estimation, and even segmentation and object detection. 

\begin{table*}
\resizebox{0.9\linewidth}{!}{
  \begin{tabular}{c|ccc|ccc|ccc|ccc|ccc}
    \toprule
    & \multicolumn{3}{c|}{NYUv2} & \multicolumn{3}{c|}{ScanNet} & \multicolumn{3}{c|}{IBims-1} & \multicolumn{3}{c|}{Sintel} & \multicolumn{3}{c}{DIODE-outdoor} \\
    Method & Mean & Med & $\delta_{11.25^\circ}$ & Mean & Med & $\delta_{11.25^\circ}$ & Mean & Med & $\delta_{11.25^\circ}$ & Mean & Med & $\delta_{11.25^\circ}$ & Mean & Med & $\delta_{11.25^\circ}$ \\
    \midrule
    DSINE & 18.6 & 9.9 & 56.1 & 18.6 & 9.9 & 56.1 & 18.8 & 8.3 & 64.1 & \cellcolor{second}34.9 & 28.1 & \cellcolor{second}21.5 & \cellcolor{second}22.0 & \cellcolor{second}14.5 & \cellcolor{second}39.6 \\
    Lotus* & \cellcolor{second}17.5 & \cellcolor{second}8.6 & \cellcolor{second}58.7 & 18.1 & \cellcolor{second}8.8 & \cellcolor{second}58.2 & 19.2 & \cellcolor{second}5.6 & 66.2 & 35.7 & 28.0 & 20.5 & 24.7 & 15.9 & 32.9 \\
    GeoWizard & 20.4 & 11.9 & 47.0 & 21.4 & 13.9 & 37.1 & 19.7 & 9.7 & 58.4 & 41.6 & 34.3 & 11.8 & 27.0 & 19.8 & 24.0 \\
    StableNormal & 19.7 & 10.5 & 53.0 & \cellcolor{second}18.1 & 10.1 & 56.0 & \cellcolor{second}17.2 & 8.1 & \cellcolor{second}66.7 & 35.0 & \cellcolor{second}27.0 & 19.5 & 26.9 & 16.1 & 36.1 \\
    Ours & \cellcolor{best}16.1 & \cellcolor{best}7.4 & \cellcolor{best}62.5 & \cellcolor{best}16.9 & \cellcolor{best}7.1 & \cellcolor{best}64.0 & \cellcolor{best}16.0 & \cellcolor{best}4.3 & \cellcolor{best}72.2 & \cellcolor{best}30.7 & \cellcolor{best}21.4 & \cellcolor{best}28.9 & \cellcolor{best}20.8 & \cellcolor{best}12.8 & \cellcolor{best}43.0 \\
    \bottomrule
\end{tabular}}
  \caption{Quantitative comparison of normal prediction. We report the mean and median angular errors with each cell colored to indicate the \colorbox{best}{best} and the \colorbox{second}{second}. Dens3R achieves accurate normal prediction for both indoor and outdoor scenes. *Note that we utilize Lotus-G for a fair comparison.}
  \label{tab:normal_result}
  \vspace{-6mm}
\end{table*}

\noindent\textbf{Multi-view Inputs.} 
To enable Dens3R to efficiently process multi-view inputs during inference, we design a simple yet effective post-processing step. This step ensures both computational efficiency in multi-view data processing and the consistency and accuracy of results. Specifically, based on Dens3R's high-precision image pair matching predictions, we establish geometric mappings between different viewpoints by constructing and optimizing a dense correspondence network across views. This approach effectively guides the model to understand geometric consistency between multiple viewpoints and accurately captures spatial relationships between views. Additionally, while maintaining high computational efficiency, it significantly improves the performance and stability of multi-view processing.

\section{Experiments}
\subsection{Normal and Matching Prediction }
We evaluate our Dens3R on several surface normal prediction datasets that include both indoor and outdoor scenes.
We compare our method with regression-based methods such as DSINE~\cite{DBLP:conf/cvpr/BaeD24} and diffusion-based methods like StableNormal~\cite{stablenormal}, GeoWizard~\cite{geowizard} and Lotus~\cite{lotus}. 
Quantitative results are shown in Tab.~\ref{tab:normal_result}, where Dens3R outperforms other methods across multiple benchmarks.
Qualitative comparisons are provided in Fig.~\ref{fig:normal}, also demonstrating that Dens3R generates more accurate and detailed normal maps.
On the DIODE dataset, our method produces more accurate normals for reflective surfaces (\textit{e.g.}, car window) and finer details in backgrounds and tree structures. 
On in-the-wild scenes, Dens3R handles both object-centric and unbounded scenarios, producing more stable and intricate surface normals.
Thanks to the intrinsic-invariant pointmap and the proposed training strategy, our method effectively reduces the ambiguity from monocular estimation, enabling more accurate and detailed predictions across various settings.

For the image-matching task, we evaluate our method on the ZEB benchmark as shown in Tab.~\ref{tab:matching_result}. We compare our method with previous dense image-matching methods and MASt3R~\cite{mast3r}. It can be seen that our method yields higher accuracy and surpasses previous methods across nearly all datasets, demonstrating our superior performance across various evaluation protocols.

\begin{table}
\resizebox{0.98\linewidth}{!}{%
  \begin{tabular}{l|c|cccccccc|cccc}
    \toprule
    \multirow{2}*{Method} & Mean & \multicolumn{8}{c|}{Real AUC$@5\degree\uparrow$} & \multicolumn{4}{c}{Simulate AUC$@5\degree\uparrow$} \\
     ~  & AUC$@5\degree\uparrow$ & GL3 & BLE & ETI & ETO & KIT & WEA & SEA & NIG & MUL & SCE & ICL & GTA \\
    \midrule

    SIFT  & 31.8 & 43.5 & 33.6 & 49.9 & 48.7 & 35.2 & 21.4 & 44.1 & 14.7 & 33.4 & 7.6 & 14.8 & 43.9 \\

    SuperGlue  & 34.3 & 43.2 & 34.2 & 58.7 & 61.0 & 29.0 & 28.3 & 48.4 & 18.8 & 34.8 & 2.8 & 15.4 & 36.5 \\

    LoFTR  & 39.1 & 50.6 & 43.9 & 62.6 & 61.6 & 35.9 & 26.8 & 47.5 & 17.6 & 41.4 & 10.2 & 25.6 & 45.0 \\

    DKM  & 51.2 & 63.3 & 53.0 & 73.9 & 76.7 & 43.4 & 34.6 & 52.5 & 24.5 & 56.6 & 32.2 & 42.5 & 61.6 \\
    ROMA  & 53.2 & \cellcolor{best}61.8 & \cellcolor{second}53.8 & \cellcolor{best}76.7 & \cellcolor{best}82.7 & 43.2 & 36.7 & 53.2 & 26.6 & 60.7 & 33.8 & 45.4 & 64.3 \\
    MASt3R  & \cellcolor{second}59.9 &  57.8 &  52.3 &  66.2 &  78.1 &  \cellcolor{second}46.2 &  \cellcolor{second}52.8 &  \cellcolor{second}70.5 &  \cellcolor{second}43.7 &  \cellcolor{second}70.1 &  \cellcolor{best}53.9 &  \cellcolor{second}60.1 &  \cellcolor{second}67.7 \\

    Ours &  \cellcolor{best}64.5 & \cellcolor{second}61.3 & \cellcolor{best}59.2 & \cellcolor{second}74.7 & \cellcolor{second}81.1 & \cellcolor{best}55.6 & \cellcolor{best}57.4 & \cellcolor{best}71.7 & \cellcolor{best}50.4 & \cellcolor{best}71.3 & \cellcolor{second}53.7 & \cellcolor{best}66.3 & \cellcolor{best}71.7 \\
    
    \bottomrule
  \end{tabular}}
    \caption{Benchmark on image matching on ZEB dataset. We report the AUC values with each cell colored to indicate the \colorbox{best}{best} and the \colorbox{second}{second}. Our method outperforms previous methods in almost all the metrics.}
  \label{tab:matching_result}
  \vspace{-4mm}
\end{table}

\subsection{Pointmap and Depth Prediction}
\begin{figure*}
    \centering
    \includegraphics[width=0.975\linewidth]{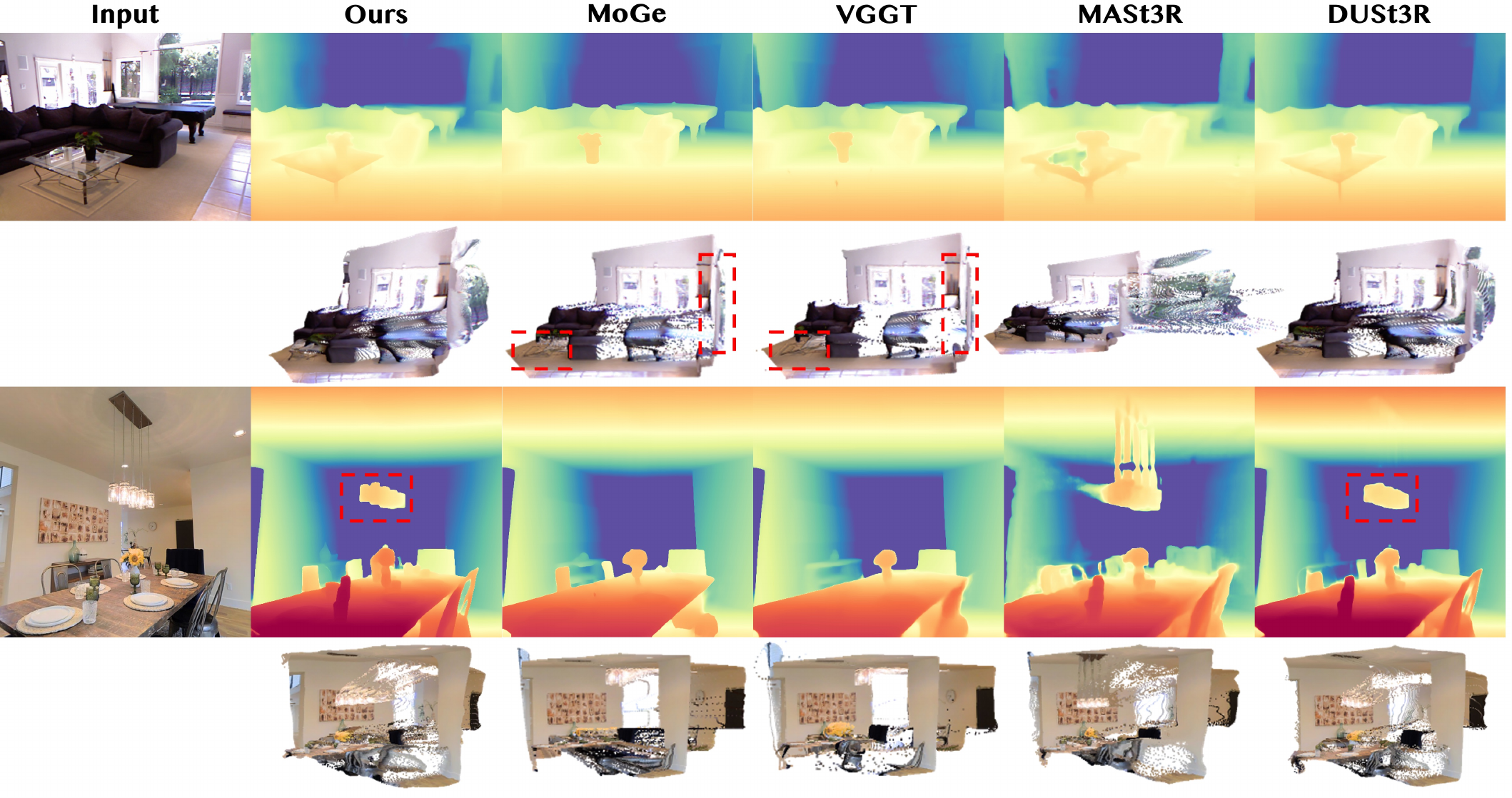}
    \caption{Qualitative comparison of depth maps and pointmaps. We compare our method with previous DUSt3R-based methods and demonstrate high-quality depth prediction results. Dens3R also reconstructs more stable and accurate pointmap than previous methods.}
    \label{fig:depth}
\end{figure*}

For monocular depth prediction, we evaluate our model on several datasets containing both indoor and outdoor scenes. We compare our method with MoGe~\cite{moge}, VGGT~\cite{vggt}, MASt3R~\cite{mast3r} and DUSt3R~\cite{dust3r}. 
The quantitative results are presented in Tab.~\ref{tab:depth_result}. It can be seen that our method achieves accurate results in depth estimation. The qualitative comparison of the depth prediction is shown in Fig.~\ref{fig:depth} with the corresponding pointmaps. Our method achieves high-quality pointmap prediction and depth estimation with the intrinsic-invariant pointmap and the novel training strategy. As for pointmap prediction, MoGe and VGGT often fail to recover depth for reflective surfaces and tend to produce flattened pointmaps in background regions.
In contrast, our method accomplishes to predict accurate depth with high-quality pointmaps. Moreover, our method yields more stable and high-quality predictions than MASt3R. It is also noteworthy that our method generates more accurate depth maps than DUSt3R, which can be reflected from the depth predictions for the Chandeliers.

\begin{table}
\resizebox{0.98\linewidth}{!}{
\begin{tabular}{c|ccccc|ccccc|ccccc}
\hline
\multirow{2}{*}{Method}                                        & \multicolumn{5}{c|}{NYUv2}                                                                                                                                                                          & \multicolumn{5}{c|}{DIODE-indoor}                                                                                                                                                                   & \multicolumn{5}{c}{DIODE-outdoor}                                                                                                                                                                   \\
                                                                        &REL$\downarrow$ & RMSE$\downarrow$ & $\delta_{1}\uparrow $ & $\delta_{2}\uparrow $ & $\delta_{3}\uparrow $                                 & REL$\downarrow$ & RMSE$\downarrow$ & $\delta_{1}\uparrow $ & $\delta_{2}\uparrow $ & $\delta_{3}\uparrow $                  & REL$\downarrow$ & RMSE$\downarrow$ & $\delta_{1}\uparrow $ & $\delta_{2}\uparrow $ & $\delta_{3}\uparrow $                  \\ \hline
GenPercept & 0.052 & 0.214 & 96.7 & 99.3 & 99.8  & 0.107 & 0.924 & 89.1 & 96.0 & 98.1 &0.727 & 5.571 & 67.3 & 84.2 & 90.6 \\
Lotus* & 0.053                                    & 0.262                                     & 96.5                                  & 99.1                                  & 99.7                                  & 0.111                                    & 1.123                                     & 88.7                                  & 96.0                                  & 98.4                                  & 0.488                                    & 9.960                                     & 47.1                                  & 63.3                                  & 71.8                                  \\
DepthAnythingV2   & 0.049                                    & 0.204                                     & 97.3                                  & 99.3                                  & 99.8                                  & 0.091                                    & 0.878                                     & 92.5                                  & 97.3                                  & 98.6                                  & 0.705                                    & 5.525                                     & 67.8                                  & 83.4                                  & 89.7                                  \\
 DUSt3R & 0.046 & 0.197 & 97.1 & 99.3 & 99.8 & 0.083 & \cellcolor{second}0.375 & 92.0 & 97.7 & 99.0 & 0.451 & 5.217 & 67.7 & 84.3 & 90.7\\
 VGGT & \cellcolor{second}0.038 & 0.194 &  \cellcolor{best}98.0 &  \cellcolor{best}99.4 & 99.8 & \cellcolor{best}0.064 & 0.404 & \cellcolor{second}93.1 & \cellcolor{best}98.0 & \cellcolor{best}99.2 & \cellcolor{second}0.400 & \cellcolor{second}4.861 & 70.6 & 84.9 & 90.6\\
MoGe                                              & \cellcolor{best}0.035   & \cellcolor{best}0.167    & \cellcolor{second}97.9 & \cellcolor{best}99.4 & \cellcolor{best}99.9 & 0.080                                    & 0.879                                     & 92.6                                  & 97.3                                  & 98.7                                  & 0.578                                    & 5.177                                     & \cellcolor{best}72.8                                  & \cellcolor{second}86.7                                  & \cellcolor{second}91.9                                  \\
Ours                                              & 0.042                                    & \cellcolor{second}0.189                                     & 97.5                                  & \cellcolor{second}99.3                                  & \cellcolor{second}99.8                                  & \cellcolor{second}0.072   & \cellcolor{best}0.372    & \cellcolor{best}93.7 & \cellcolor{second}97.5 & \cellcolor{second}98.8 & \cellcolor{best}0.387   & \cellcolor{best}4.740    & \cellcolor{second}72.2 & \cellcolor{best}87.0 & \cellcolor{best}92.3 \\ \hline
\end{tabular}}
  \caption{Quantitative comparison on monocular depth prediction. We report the relative point error (REL), root mean square error (RMSE) and the percentage of inliers $\delta_{1}, \delta_{2}, \delta_{3}$ with each cell colored to indicate the \colorbox{best}{best} and the \colorbox{second}{second}. *Note that we utilize Lotu-G disparity model for comparison.}
  \label{tab:depth_result}
  \vspace{-6mm}
\end{table}

\subsection{Ablation Study}
We present high-quality geometric predictions for high-resolution inputs and various scenarios in Fig.~\ref{fig:2k} and Fig.~\ref{fig:sup_vis}. More evaluations and comparisons are also provided in the \textit{Supplementary Materials}. We then  conducted comprehensive ablation studies for our key components: the position-interpolated rotary positional encoding, the intrinsic-invariant training and the coarse-to-fine training strategy.

\begin{table}[h]
    \centering
\resizebox{0.9\linewidth}{!}{%
 \begin{tabular}{c|cc|cc|cc}
\toprule
Method      & \multicolumn{2}{c|}{NYUv2}       & \multicolumn{2}{c|}{ScanNet}     & \multicolumn{2}{c}{IBims}        \\
            & Mean $\downarrow$ & $\delta_{11.25^\circ}\uparrow$ & Mean $\downarrow$ & $\delta_{11.25^\circ}\uparrow$ & Mean $\downarrow$ & $\delta_{11.25^\circ}\uparrow$ \\
\midrule
w/o IIT     & 17.8 & 50.6 & 18.6 & 49.4 & 20.2 & 56.8 \\
w/o C2F     & 17.6 & 50.5 & 17.8 & 58.8 & 18.6 & 63.9 \\
Ours        & \textbf{16.1} & \textbf{62.5} & \textbf{16.9} & \textbf{64.0} & \textbf{16.0} & \textbf{72.2} \\
\bottomrule
\end{tabular}
}
    \caption{Normal quantitative metrics for ablation. We demonstrate that both the intrinsic-invariant training and coarse-to-fine strategy contributes to accurate normal predictions.}
  \label{tab:ablation}
   \vspace{-6mm}
\end{table}

\begin{figure*}[t]
  \centering
  \begin{subfigure}[t]{0.48\linewidth}
    \centering
    \includegraphics[width=0.95\linewidth]{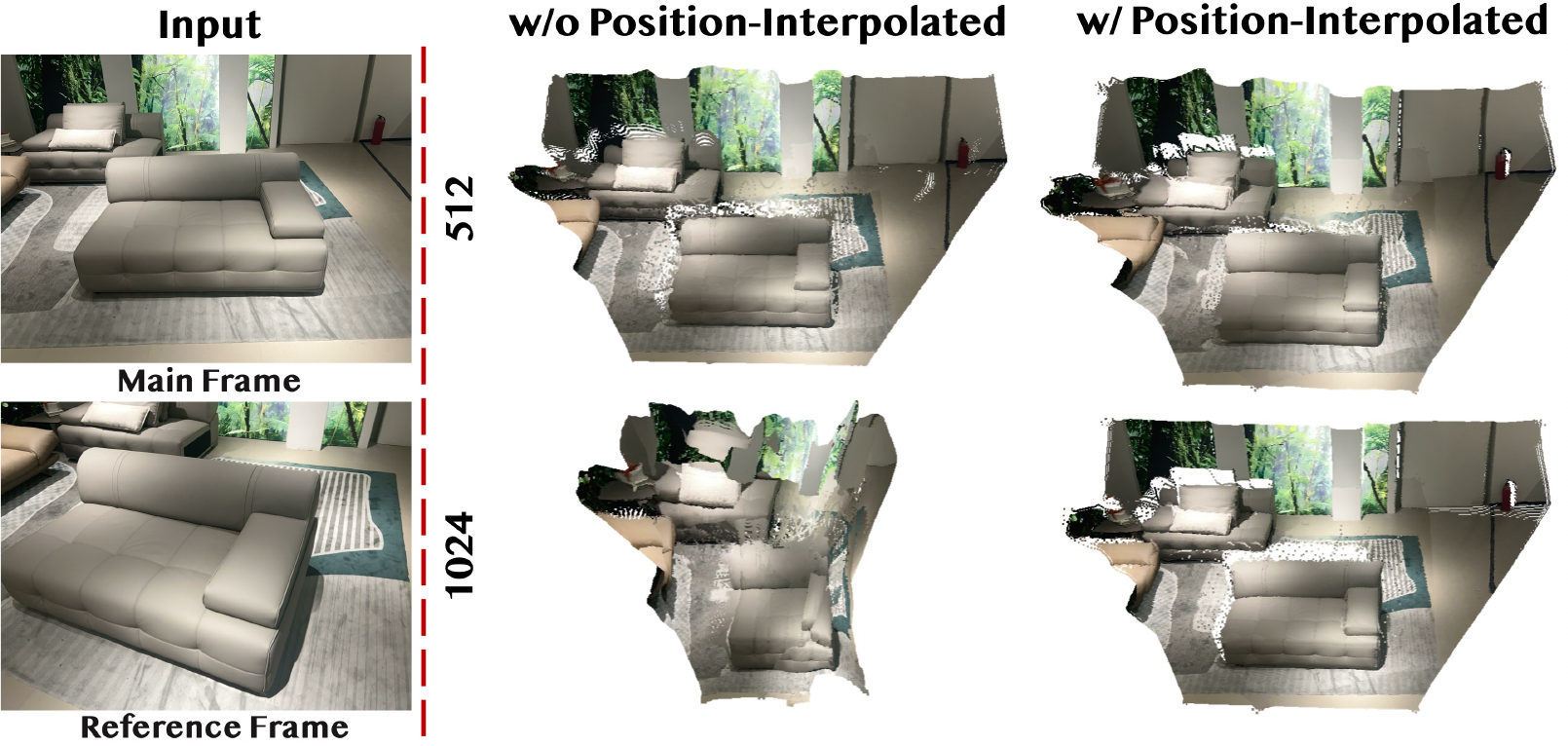}
    \caption{High-resolution inference comparison. Our method supports high-resolution input and generates accurate and well-structured pointmaps.}
    \label{fig:infer_highres}
  \end{subfigure}
  \hfill
  \begin{subfigure}[t]{0.48\linewidth}
    \centering
    \includegraphics[width=\linewidth]{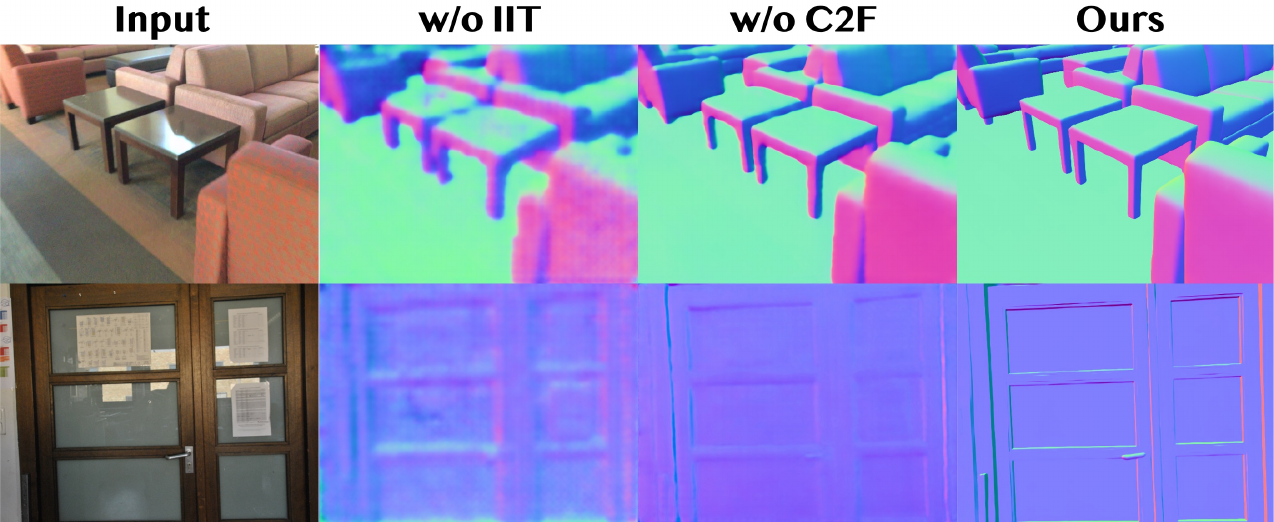}
    \caption{Normal comparison for ablation. The intrinsic-invariant training enables accurate normal prediction and the coarse-to-fine training enhances details.}
    \label{fig:ab}
  \end{subfigure}

  \begin{subfigure}[t]{0.48\linewidth}
    \centering
    \includegraphics[width=\linewidth]{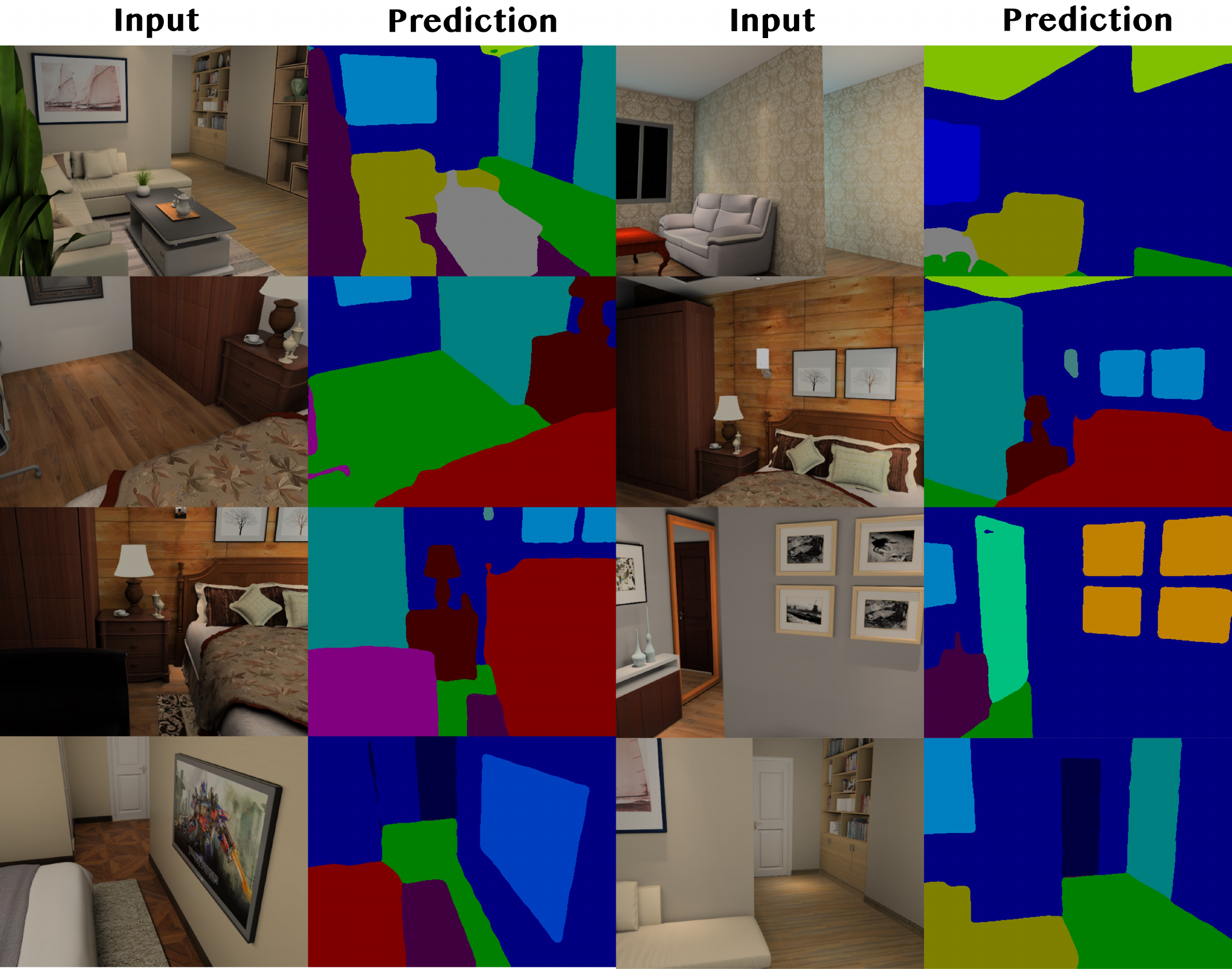}
    \caption{Segmentation results. Our model can be easily extended to segmentation tasks by training a new prediction head with the backbone frozen. }
    \label{fig:seg}
  \end{subfigure}
  \hfill
  \begin{subfigure}[t]{0.48\linewidth}
    \centering
    \includegraphics[width=\linewidth]{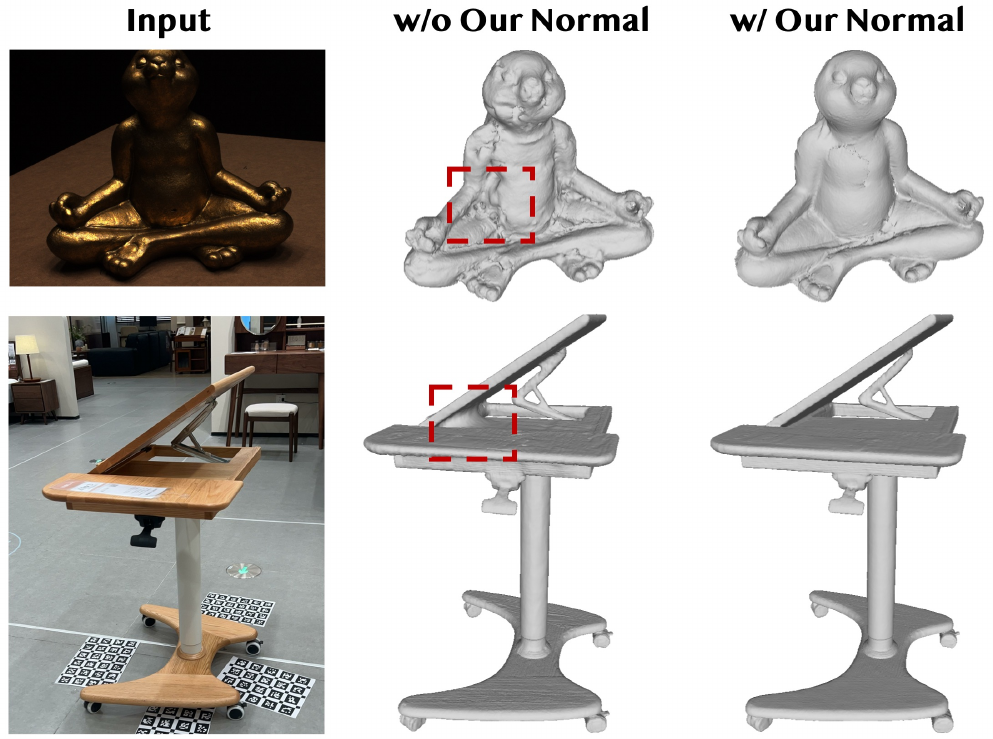}
    \caption{Normal supervision results. We demonstrate the effectiveness of using our normal as the supervision of surface reconstruction.}
    \label{fig:neus_sup}
  \end{subfigure}

  \caption{Ablation and downstream applications.}
    \vspace{-6mm}
  \label{fig:supp_all}
\end{figure*}

\noindent\textbf{{Position-Interpolated Rotary Positional Encoding.}}
Dens3R can support multi-resolution image inputs. With the position-interpolated rotary positional encoding and the coarse-to-fine training strategy, our method can prevent performance degradation when handling high-resolution inputs. As shown in Fig.~\ref{fig:infer_highres}, we can generate accurate and well-structured pointmaps with the position-interpolated RoPE, preventing the model from producing overlapping or inconsistent pointmaps at higher resolutions.

\noindent\textbf{{Intrinsic-Invariant Training.}}
Our approach first learns a scale-invariant pointmap, which is then transformed into an intrinsic-invariant pointmap via subsequent intrinsic-invariant training.
We find that jointly training the pointmap and normal at the initial scale-invariant stage leads to instability and poor convergence.
This is because pointmaps and normal maps lie in different data domains, and coupling their supervision potentially increases training complexity.
While GeoWizard~\cite{geowizard} addresses this domain gap with a task switcher, we adopt a two-stage training scheme to learn an intrinsic-invariant pointmap, ensuring stable learning.
As shown in Tab.~\ref{tab:ablation} and in Fig.~\ref{fig:ab}, the performance of the model will degrade without the intrinsic-invariant training.

\noindent\textbf{{Coarse-to-Fine Training.}}
Our model is trained on diverse training dataset of varying quality. To better utilize the full training set, we implement a coarse-to-fine training strategy that gradually increases resolution and data fidelity.
In the coarse stage, we set the max resolution of the training images as $512$ and enable all the training data. In the fine stage, we increase the image resolution to 1024 pixels and restrict training to the high-resolution data only. 
As demonstrated in Tab.~\ref{tab:ablation} and in Fig.~\ref{fig:ab}, this strategy improves prediction accuracy, particularly for high-resolution outputs.

\subsection{Downstream Applications}
\noindent\textbf{{Segmentation Head Training.}}
Dens3R serves as a visual foundation model that can be finetuned for several downstream tasks. We demonstrate this by training a new prediction head for segmentation task while keeping our backbone frozen. 
As shown in Fig.~\ref{fig:seg}, the segmentation head can generate accurate results, with much more effortless training than a large segmentation model.

\noindent\textbf{{Surface Reconstruction.}}
Dens3R can improve surface reconstruction quality by its sharp and accurate normals. We demonstrate this by utilizing our predicted normals as the supervision for NeuS~\cite{neus} training. The results are showcased in Fig.~\ref{fig:neus_sup}. It can be seen that the final reconstruction results are improved due to the strong normal prior provided by our Dens3R.

\section{Conclusion}
We propose Dens3R, a 3D foundation model for dense geometric prediction that jointly regresses multiple geometric quantities, including depth, surface normals, and pointmaps, from unconstrained image inputs. Unlike previous approaches that estimate geometry in isolation, Dens3R explicitly models the structural coupling among these properties to ensure consistency and improves overall accuracy. We utilize a two-stage training framework with coarse-to-fine strategy and build an accurate intrinsic-invariant pointmap representation. In addition, we design a lightweight encoder-decoder architecture and position-interpolated rotary positional encoding to enable scalable and high-fidelity inference for high-resolution inputs. Moreover, Dens3R incorporates a geometrically consistent post-processing pipeline for multi-view inputs. Extensive experiments demonstrate our superior performance across various 3D prediction benchmarks and highlight the potential as a versatile backbone for broader downstream applications.



\bibliographystyle{ACM-Reference-Format}
\bibliography{ours_cite}


\begin{thebibliography}{98}


\ifx \showCODEN    \undefined \def \showCODEN     #1{\unskip}     \fi
\ifx \showISBNx    \undefined \def \showISBNx     #1{\unskip}     \fi
\ifx \showISBNxiii \undefined \def \showISBNxiii  #1{\unskip}     \fi
\ifx \showISSN     \undefined \def \showISSN      #1{\unskip}     \fi
\ifx \showLCCN     \undefined \def \showLCCN      #1{\unskip}     \fi
\ifx \shownote     \undefined \def \shownote      #1{#1}          \fi
\ifx \showarticletitle \undefined \def \showarticletitle #1{#1}   \fi
\ifx \showURL      \undefined \def \showURL       {\relax}        \fi
\providecommand\bibfield[2]{#2}
\providecommand\bibinfo[2]{#2}
\providecommand\natexlab[1]{#1}
\providecommand\showeprint[2][]{arXiv:#2}

\bibitem[Arnold et~al\mbox{.}(2022)]%
        {arnold2022mapfree}
\bibfield{author}{\bibinfo{person}{Eduardo Arnold}, \bibinfo{person}{Jamie Wynn}, \bibinfo{person}{Sara Vicente}, \bibinfo{person}{Guillermo Garcia-Hernando}, \bibinfo{person}{{\'{A}}ron Monszpart}, \bibinfo{person}{Victor~Adrian Prisacariu}, \bibinfo{person}{Daniyar Turmukhambetov}, {and} \bibinfo{person}{Eric Brachmann}.} \bibinfo{year}{2022}\natexlab{}.
\newblock \showarticletitle{Map-free Visual Relocalization: Metric Pose Relative to a Single Image}. In \bibinfo{booktitle}{\emph{ECCV}}, Vol.~\bibinfo{volume}{13661}. \bibinfo{pages}{690--708}.
\newblock


\bibitem[Bae and Davison(2024)]%
        {DBLP:conf/cvpr/BaeD24}
\bibfield{author}{\bibinfo{person}{Gwangbin Bae} {and} \bibinfo{person}{Andrew~J. Davison}.} \bibinfo{year}{2024}\natexlab{}.
\newblock \showarticletitle{Rethinking Inductive Biases for Surface Normal Estimation}. In \bibinfo{booktitle}{\emph{{CVPR}}}. \bibinfo{pages}{9535--9545}.
\newblock


\bibitem[Bansal et~al\mbox{.}(2016)]%
        {DBLP:conf/cvpr/BansalRG16}
\bibfield{author}{\bibinfo{person}{Aayush Bansal}, \bibinfo{person}{Bryan~C. Russell}, {and} \bibinfo{person}{Abhinav Gupta}.} \bibinfo{year}{2016}\natexlab{}.
\newblock \showarticletitle{Marr Revisited: 2D-3D Alignment via Surface Normal Prediction}. In \bibinfo{booktitle}{\emph{{CVPR}}}. \bibinfo{pages}{5965--5974}.
\newblock


\bibitem[Barron et~al\mbox{.}(2021)]%
        {mipnerf}
\bibfield{author}{\bibinfo{person}{Jonathan~T. Barron}, \bibinfo{person}{Ben Mildenhall}, \bibinfo{person}{Matthew Tancik}, \bibinfo{person}{Peter Hedman}, \bibinfo{person}{Ricardo Martin{-}Brualla}, {and} \bibinfo{person}{Pratul~P. Srinivasan}.} \bibinfo{year}{2021}\natexlab{}.
\newblock \showarticletitle{Mip-NeRF: {A} Multiscale Representation for Anti-Aliasing Neural Radiance Fields}. In \bibinfo{booktitle}{\emph{ICCV}}. \bibinfo{pages}{5835--5844}.
\newblock


\bibitem[Barron et~al\mbox{.}(2023)]%
        {Barron_2023_ICCV}
\bibfield{author}{\bibinfo{person}{Jonathan~T. Barron}, \bibinfo{person}{Ben Mildenhall}, \bibinfo{person}{Dor Verbin}, \bibinfo{person}{Pratul~P. Srinivasan}, {and} \bibinfo{person}{Peter Hedman}.} \bibinfo{year}{2023}\natexlab{}.
\newblock \showarticletitle{Zip-NeRF: Anti-Aliased Grid-Based Neural Radiance Fields}. In \bibinfo{booktitle}{\emph{ICCV}}. \bibinfo{pages}{19697--19705}.
\newblock


\bibitem[Baruch et~al\mbox{.}(2021)]%
        {dehghan2021arkitscenes}
\bibfield{author}{\bibinfo{person}{Gilad Baruch}, \bibinfo{person}{Zhuoyuan Chen}, \bibinfo{person}{Afshin Dehghan}, \bibinfo{person}{Tal Dimry}, \bibinfo{person}{Yuri Feigin}, \bibinfo{person}{Peter Fu}, \bibinfo{person}{Thomas Gebauer}, \bibinfo{person}{Brandon Joffe}, \bibinfo{person}{Daniel Kurz}, \bibinfo{person}{Arik Schwartz}, {and} \bibinfo{person}{Elad Shulman}.} \bibinfo{year}{2021}\natexlab{}.
\newblock \showarticletitle{{ARK}itScenes - A Diverse Real-World Dataset for 3D Indoor Scene Understanding Using Mobile {RGB}-D Data}. In \bibinfo{booktitle}{\emph{NeurIPS Datasets and Benchmarks}}.
\newblock


\bibitem[Bhalgat et~al\mbox{.}(2023)]%
        {DBLP:conf/cvpr/BhalgatHZ23}
\bibfield{author}{\bibinfo{person}{Yash Bhalgat}, \bibinfo{person}{Jo{\~{a}}o~F. Henriques}, {and} \bibinfo{person}{Andrew Zisserman}.} \bibinfo{year}{2023}\natexlab{}.
\newblock \showarticletitle{A Light Touch Approach to Teaching Transformers Multi-view Geometry}. In \bibinfo{booktitle}{\emph{{CVPR}}}. \bibinfo{pages}{4958--4969}.
\newblock


\bibitem[Bhat et~al\mbox{.}(2021)]%
        {DBLP:conf/cvpr/BhatAW21}
\bibfield{author}{\bibinfo{person}{Shariq~Farooq Bhat}, \bibinfo{person}{Ibraheem Alhashim}, {and} \bibinfo{person}{Peter Wonka}.} \bibinfo{year}{2021}\natexlab{}.
\newblock \showarticletitle{AdaBins: Depth Estimation Using Adaptive Bins}. In \bibinfo{booktitle}{\emph{{CVPR}}}. \bibinfo{pages}{4009--4018}.
\newblock


\bibitem[Bhat et~al\mbox{.}(2023)]%
        {DBLP:journals/corr/abs-2302-12288}
\bibfield{author}{\bibinfo{person}{Shariq~Farooq Bhat}, \bibinfo{person}{Reiner Birkl}, \bibinfo{person}{Diana Wofk}, \bibinfo{person}{Peter Wonka}, {and} \bibinfo{person}{Matthias M{\"{u}}ller}.} \bibinfo{year}{2023}\natexlab{}.
\newblock \showarticletitle{ZoeDepth: Zero-shot Transfer by Combining Relative and Metric Depth}.
\newblock \bibinfo{journal}{\emph{arxiv preprint arXiv:2302.12288}} (\bibinfo{year}{2023}).
\newblock


\bibitem[Chang et~al\mbox{.}(2019)]%
        {Argoverse}
\bibfield{author}{\bibinfo{person}{Ming-Fang Chang}, \bibinfo{person}{John~W Lambert}, \bibinfo{person}{Patsorn Sangkloy}, \bibinfo{person}{Jagjeet Singh}, \bibinfo{person}{Slawomir Bak}, \bibinfo{person}{Andrew Hartnett}, \bibinfo{person}{De Wang}, \bibinfo{person}{Peter Carr}, \bibinfo{person}{Simon Lucey}, \bibinfo{person}{Deva Ramanan}, {and} \bibinfo{person}{James Hays}.} \bibinfo{year}{2019}\natexlab{}.
\newblock \showarticletitle{Argoverse: 3D Tracking and Forecasting with Rich Maps}. In \bibinfo{booktitle}{\emph{CVPR}}. \bibinfo{pages}{8748--8757}.
\newblock


\bibitem[Chen et~al\mbox{.}(2023)]%
        {ropellm}
\bibfield{author}{\bibinfo{person}{Shouyuan Chen}, \bibinfo{person}{Sherman Wong}, \bibinfo{person}{Liangjian Chen}, {and} \bibinfo{person}{Yuandong Tian}.} \bibinfo{year}{2023}\natexlab{}.
\newblock \showarticletitle{Extending Context Window of Large Language Models via Positional Interpolation}.
\newblock \bibinfo{journal}{\emph{arxiv preprint arXiv:2306.15595}} (\bibinfo{year}{2023}).
\newblock


\bibitem[Chen et~al\mbox{.}(2016)]%
        {DBLP:conf/nips/ChenFYD16}
\bibfield{author}{\bibinfo{person}{Weifeng Chen}, \bibinfo{person}{Zhao Fu}, \bibinfo{person}{Dawei Yang}, {and} \bibinfo{person}{Jia Deng}.} \bibinfo{year}{2016}\natexlab{}.
\newblock \showarticletitle{Single-Image Depth Perception in the Wild}. In \bibinfo{booktitle}{\emph{NeurIPS}}. \bibinfo{pages}{730--738}.
\newblock


\bibitem[Chen et~al\mbox{.}(2020)]%
        {DBLP:conf/cvpr/ChenQFKHD20}
\bibfield{author}{\bibinfo{person}{Weifeng Chen}, \bibinfo{person}{Shengyi Qian}, \bibinfo{person}{David Fan}, \bibinfo{person}{Noriyuki Kojima}, \bibinfo{person}{Max Hamilton}, {and} \bibinfo{person}{Jia Deng}.} \bibinfo{year}{2020}\natexlab{}.
\newblock \showarticletitle{{OASIS:} {A} Large-Scale Dataset for Single Image 3D in the Wild}. In \bibinfo{booktitle}{\emph{{CVPR}}}. \bibinfo{pages}{676--685}.
\newblock


\bibitem[Cho et~al\mbox{.}(2019)]%
        {cho2019large}
\bibfield{author}{\bibinfo{person}{Jaehoon Cho}, \bibinfo{person}{Dongbo Min}, \bibinfo{person}{Youngjung Kim}, {and} \bibinfo{person}{Kwanghoon Sohn}.} \bibinfo{year}{2019}\natexlab{}.
\newblock \showarticletitle{A large RGB-D dataset for semi-supervised monocular depth estimation}.
\newblock \bibinfo{journal}{\emph{arXiv preprint arXiv:1904.10230}} (\bibinfo{year}{2019}).
\newblock


\bibitem[Collins et~al\mbox{.}(2022)]%
        {collins2022abo}
\bibfield{author}{\bibinfo{person}{Jasmine Collins}, \bibinfo{person}{Shubham Goel}, \bibinfo{person}{Kenan Deng}, \bibinfo{person}{Achleshwar Luthra}, \bibinfo{person}{Leon Xu}, \bibinfo{person}{Erhan Gundogdu}, \bibinfo{person}{Xi Zhang}, \bibinfo{person}{Tomas~F Yago~Vicente}, \bibinfo{person}{Thomas Dideriksen}, \bibinfo{person}{Himanshu Arora}, \bibinfo{person}{Matthieu Guillaumin}, {and} \bibinfo{person}{Jitendra Malik}.} \bibinfo{year}{2022}\natexlab{}.
\newblock \showarticletitle{ABO: Dataset and Benchmarks for Real-World 3D Object Understanding}. In \bibinfo{booktitle}{\emph{CVPR}}. \bibinfo{pages}{21094--21104}.
\newblock


\bibitem[Deitke et~al\mbox{.}(2023)]%
        {objaverseXL}
\bibfield{author}{\bibinfo{person}{Matt Deitke}, \bibinfo{person}{Ruoshi Liu}, \bibinfo{person}{Matthew Wallingford}, \bibinfo{person}{Huong Ngo}, \bibinfo{person}{Oscar Michel}, \bibinfo{person}{Aditya Kusupati}, \bibinfo{person}{Alan Fan}, \bibinfo{person}{Christian Laforte}, \bibinfo{person}{Vikram Voleti}, \bibinfo{person}{Samir~Yitzhak Gadre}, \bibinfo{person}{Eli VanderBilt}, \bibinfo{person}{Aniruddha Kembhavi}, \bibinfo{person}{Carl Vondrick}, \bibinfo{person}{Georgia Gkioxari}, \bibinfo{person}{Kiana Ehsani}, \bibinfo{person}{Ludwig Schmidt}, {and} \bibinfo{person}{Ali Farhadi}.} \bibinfo{year}{2023}\natexlab{}.
\newblock \showarticletitle{Objaverse-XL: {A} Universe of 10M+ 3D Objects}. In \bibinfo{booktitle}{\emph{NeurIPS}}.
\newblock


\bibitem[Edstedt et~al\mbox{.}(2023)]%
        {DBLP:conf/cvpr/EdstedtAWF23}
\bibfield{author}{\bibinfo{person}{Johan Edstedt}, \bibinfo{person}{Ioannis Athanasiadis}, \bibinfo{person}{M{\aa}rten Wadenb{\"{a}}ck}, {and} \bibinfo{person}{Michael Felsberg}.} \bibinfo{year}{2023}\natexlab{}.
\newblock \showarticletitle{{DKM:} Dense Kernelized Feature Matching for Geometry Estimation}. In \bibinfo{booktitle}{\emph{{CVPR}}}. \bibinfo{pages}{17765--17775}.
\newblock


\bibitem[Edstedt et~al\mbox{.}(2024)]%
        {DBLP:conf/cvpr/EdstedtSBWF24}
\bibfield{author}{\bibinfo{person}{Johan Edstedt}, \bibinfo{person}{Qiyu Sun}, \bibinfo{person}{Georg B{\"{o}}kman}, \bibinfo{person}{M{\aa}rten Wadenb{\"{a}}ck}, {and} \bibinfo{person}{Michael Felsberg}.} \bibinfo{year}{2024}\natexlab{}.
\newblock \showarticletitle{RoMa: Robust Dense Feature Matching}. In \bibinfo{booktitle}{\emph{{CVPR}}}. \bibinfo{pages}{19790--19800}.
\newblock


\bibitem[Efe et~al\mbox{.}(2021)]%
        {DBLP:conf/cvpr/EfeIA21}
\bibfield{author}{\bibinfo{person}{Ufuk Efe}, \bibinfo{person}{Kutalmis~Gokalp Ince}, {and} \bibinfo{person}{A.~Aydin Alatan}.} \bibinfo{year}{2021}\natexlab{}.
\newblock \showarticletitle{{DFM:} {A} Performance Baseline for Deep Feature Matching}. In \bibinfo{booktitle}{\emph{{CVPRW}}}. \bibinfo{pages}{4284--4293}.
\newblock


\bibitem[Eftekhar et~al\mbox{.}(2021)]%
        {DBLP:conf/iccv/EftekharSMZ21}
\bibfield{author}{\bibinfo{person}{Ainaz Eftekhar}, \bibinfo{person}{Alexander Sax}, \bibinfo{person}{Jitendra Malik}, {and} \bibinfo{person}{Amir Zamir}.} \bibinfo{year}{2021}\natexlab{}.
\newblock \showarticletitle{Omnidata: {A} Scalable Pipeline for Making Multi-Task Mid-Level Vision Datasets from 3D Scans}. In \bibinfo{booktitle}{\emph{{ICCV}}}. \bibinfo{pages}{10766--10776}.
\newblock


\bibitem[Eigen et~al\mbox{.}(2014)]%
        {DBLP:conf/nips/EigenPF14}
\bibfield{author}{\bibinfo{person}{David Eigen}, \bibinfo{person}{Christian Puhrsch}, {and} \bibinfo{person}{Rob Fergus}.} \bibinfo{year}{2014}\natexlab{}.
\newblock \showarticletitle{Depth Map Prediction from a Single Image using a Multi-Scale Deep Network}. In \bibinfo{booktitle}{\emph{NeurIPS}}. \bibinfo{pages}{2366--2374}.
\newblock


\bibitem[Fu et~al\mbox{.}(2024)]%
        {geowizard}
\bibfield{author}{\bibinfo{person}{Xiao Fu}, \bibinfo{person}{Wei Yin}, \bibinfo{person}{Mu Hu}, \bibinfo{person}{Kaixuan Wang}, \bibinfo{person}{Yuexin Ma}, \bibinfo{person}{Ping Tan}, \bibinfo{person}{Shaojie Shen}, \bibinfo{person}{Dahua Lin}, {and} \bibinfo{person}{Xiaoxiao Long}.} \bibinfo{year}{2024}\natexlab{}.
\newblock \showarticletitle{GeoWizard: Unleashing the Diffusion Priors for 3D Geometry Estimation from a Single Image}. In \bibinfo{booktitle}{\emph{{ECCV}}}, Vol.~\bibinfo{volume}{15080}. \bibinfo{pages}{241--258}.
\newblock


\bibitem[Gaidon et~al\mbox{.}(2016)]%
        {gaidon2016virtual}
\bibfield{author}{\bibinfo{person}{Adrien Gaidon}, \bibinfo{person}{Qiao Wang}, \bibinfo{person}{Yohann Cabon}, {and} \bibinfo{person}{Eleonora Vig}.} \bibinfo{year}{2016}\natexlab{}.
\newblock \showarticletitle{Virtual worlds as proxy for multi-object tracking analysis}. In \bibinfo{booktitle}{\emph{CVPR}}. \bibinfo{pages}{4340--4349}.
\newblock


\bibitem[Godard et~al\mbox{.}(2019)]%
        {DBLP:conf/iccv/GodardAFB19}
\bibfield{author}{\bibinfo{person}{Cl{\'{e}}ment Godard}, \bibinfo{person}{Oisin~Mac Aodha}, \bibinfo{person}{Michael Firman}, {and} \bibinfo{person}{Gabriel~J. Brostow}.} \bibinfo{year}{2019}\natexlab{}.
\newblock \showarticletitle{Digging Into Self-Supervised Monocular Depth Estimation}. In \bibinfo{booktitle}{\emph{{ICCV}}}. \bibinfo{pages}{3827--3837}.
\newblock


\bibitem[Gui et~al\mbox{.}(2024)]%
        {DBLP:journals/corr/abs-2403-13788}
\bibfield{author}{\bibinfo{person}{Ming Gui}, \bibinfo{person}{Johannes~S. Fischer}, \bibinfo{person}{Ulrich Prestel}, \bibinfo{person}{Pingchuan Ma}, \bibinfo{person}{Dmytro Kotovenko}, \bibinfo{person}{Olga Grebenkova}, \bibinfo{person}{Stefan~Andreas Baumann}, \bibinfo{person}{Vincent~Tao Hu}, {and} \bibinfo{person}{Bj{\"{o}}rn Ommer}.} \bibinfo{year}{2024}\natexlab{}.
\newblock \showarticletitle{DepthFM: Fast Monocular Depth Estimation with Flow Matching}.
\newblock \bibinfo{journal}{\emph{arxiv preprint arXiv:2403.13788}} (\bibinfo{year}{2024}).
\newblock


\bibitem[Gómez et~al\mbox{.}(2025)]%
        {gomez2025}
\bibfield{author}{\bibinfo{person}{Jose~L. Gómez}, \bibinfo{person}{Manuel Silva}, \bibinfo{person}{Antonio Seoane}, \bibinfo{person}{Agnés Borràs}, \bibinfo{person}{Mario Noriega}, \bibinfo{person}{German Ros}, \bibinfo{person}{Jose~A. Iglesias-Guitian}, {and} \bibinfo{person}{Antonio~M. López}.} \bibinfo{year}{2025}\natexlab{}.
\newblock \showarticletitle{All for one, and one for all: UrbanSyn Dataset, the third Musketeer of synthetic driving scenes}.
\newblock \bibinfo{journal}{\emph{Neurocomputing}}  \bibinfo{volume}{637} (\bibinfo{year}{2025}), \bibinfo{pages}{130038}.
\newblock


\bibitem[He et~al\mbox{.}(2025)]%
        {lotus}
\bibfield{author}{\bibinfo{person}{Jing He}, \bibinfo{person}{Haodong Li}, \bibinfo{person}{Wei Yin}, \bibinfo{person}{Yixun Liang}, \bibinfo{person}{Leheng Li}, \bibinfo{person}{Kaiqiang Zhou}, \bibinfo{person}{Hongbo Liu}, \bibinfo{person}{Bingbing Liu}, {and} \bibinfo{person}{Ying-Cong Chen}.} \bibinfo{year}{2025}\natexlab{}.
\newblock \showarticletitle{Lotus: Diffusion-based Visual Foundation Model for High-quality Dense Prediction}. In \bibinfo{booktitle}{\emph{ICLR}}.
\newblock


\bibitem[He et~al\mbox{.}(2020)]%
        {DBLP:conf/cvpr/HeYFY20a}
\bibfield{author}{\bibinfo{person}{Yihui He}, \bibinfo{person}{Rui Yan}, \bibinfo{person}{Katerina Fragkiadaki}, {and} \bibinfo{person}{Shoou{-}I Yu}.} \bibinfo{year}{2020}\natexlab{}.
\newblock \showarticletitle{Epipolar Transformers}. In \bibinfo{booktitle}{\emph{{CVPR}}}. \bibinfo{pages}{7776--7785}.
\newblock


\bibitem[Hong et~al\mbox{.}(2024)]%
        {coponerf}
\bibfield{author}{\bibinfo{person}{Sunghwan Hong}, \bibinfo{person}{Jaewoo Jung}, \bibinfo{person}{Heeseong Shin}, \bibinfo{person}{Jiaolong Yang}, \bibinfo{person}{Seungryong Kim}, {and} \bibinfo{person}{Chong Luo}.} \bibinfo{year}{2024}\natexlab{}.
\newblock \showarticletitle{Unifying Correspondence, Pose and NeRF for Pose-Free Novel View Synthesis from Stereo Pairs}. In \bibinfo{booktitle}{\emph{{CVPR}}}.
\newblock


\bibitem[Hu et~al\mbox{.}(2024)]%
        {DBLP:journals/pami/HuYZCLCWYSS24}
\bibfield{author}{\bibinfo{person}{Mu Hu}, \bibinfo{person}{Wei Yin}, \bibinfo{person}{Chi Zhang}, \bibinfo{person}{Zhipeng Cai}, \bibinfo{person}{Xiaoxiao Long}, \bibinfo{person}{Hao Chen}, \bibinfo{person}{Kaixuan Wang}, \bibinfo{person}{Gang Yu}, \bibinfo{person}{Chunhua Shen}, {and} \bibinfo{person}{Shaojie Shen}.} \bibinfo{year}{2024}\natexlab{}.
\newblock \showarticletitle{Metric3D v2: {A} Versatile Monocular Geometric Foundation Model for Zero-Shot Metric Depth and Surface Normal Estimation}.
\newblock \bibinfo{journal}{\emph{{IEEE} Trans. Pattern Anal. Mach. Intell.}} \bibinfo{volume}{46}, \bibinfo{number}{12} (\bibinfo{year}{2024}), \bibinfo{pages}{10579--10596}.
\newblock


\bibitem[Jin et~al\mbox{.}(2025)]%
        {lvsm}
\bibfield{author}{\bibinfo{person}{Haian Jin}, \bibinfo{person}{Hanwen Jiang}, \bibinfo{person}{Hao Tan}, \bibinfo{person}{Kai Zhang}, \bibinfo{person}{Sai Bi}, \bibinfo{person}{Tianyuan Zhang}, \bibinfo{person}{Fujun Luan}, \bibinfo{person}{Noah Snavely}, {and} \bibinfo{person}{Zexiang Xu}.} \bibinfo{year}{2025}\natexlab{}.
\newblock \showarticletitle{LVSM: A Large View Synthesis Model with Minimal 3D Inductive Bias}. In \bibinfo{booktitle}{\emph{ICLR}}.
\newblock


\bibitem[Ke et~al\mbox{.}(2024)]%
        {DBLP:conf/cvpr/KeOHMDS24}
\bibfield{author}{\bibinfo{person}{Bingxin Ke}, \bibinfo{person}{Anton Obukhov}, \bibinfo{person}{Shengyu Huang}, \bibinfo{person}{Nando Metzger}, \bibinfo{person}{Rodrigo~Caye Daudt}, {and} \bibinfo{person}{Konrad Schindler}.} \bibinfo{year}{2024}\natexlab{}.
\newblock \showarticletitle{Repurposing Diffusion-Based Image Generators for Monocular Depth Estimation}. In \bibinfo{booktitle}{\emph{{CVPR}}}. \bibinfo{pages}{9492--9502}.
\newblock


\bibitem[Kerbl et~al\mbox{.}(2023)]%
        {3dgs}
\bibfield{author}{\bibinfo{person}{Bernhard Kerbl}, \bibinfo{person}{Georgios Kopanas}, \bibinfo{person}{Thomas Leimk{\"{u}}hler}, {and} \bibinfo{person}{George Drettakis}.} \bibinfo{year}{2023}\natexlab{}.
\newblock \showarticletitle{3D Gaussian Splatting for Real-Time Radiance Field Rendering}.
\newblock \bibinfo{journal}{\emph{{ACM} Trans. Graph.}} \bibinfo{volume}{42}, \bibinfo{number}{4} (\bibinfo{year}{2023}), \bibinfo{pages}{139:1--139:14}.
\newblock


\bibitem[Leroy et~al\mbox{.}(2024)]%
        {mast3r}
\bibfield{author}{\bibinfo{person}{Vincent Leroy}, \bibinfo{person}{Yohann Cabon}, {and} \bibinfo{person}{J{\'{e}}r{\^{o}}me Revaud}.} \bibinfo{year}{2024}\natexlab{}.
\newblock \showarticletitle{Grounding Image Matching in 3D with MASt3R}. In \bibinfo{booktitle}{\emph{{ECCV}}}, Vol.~\bibinfo{volume}{15130}. \bibinfo{pages}{71--91}.
\newblock


\bibitem[Li et~al\mbox{.}(2022)]%
        {conf/corl/0002ZWGSMWLLSAH22}
\bibfield{author}{\bibinfo{person}{Chengshu Li}, \bibinfo{person}{Ruohan Zhang}, \bibinfo{person}{Josiah Wong}, \bibinfo{person}{Cem Gokmen}, \bibinfo{person}{Sanjana Srivastava}, \bibinfo{person}{Roberto Martín-Martín}, \bibinfo{person}{Chen Wang}, \bibinfo{person}{Gabrael Levine}, \bibinfo{person}{Michael Lingelbach}, \bibinfo{person}{Jiankai Sun}, \bibinfo{person}{Mona Anvari}, \bibinfo{person}{Minjune Hwang}, \bibinfo{person}{Manasi Sharma}, \bibinfo{person}{Arman Aydin}, \bibinfo{person}{Dhruva Bansal}, \bibinfo{person}{Samuel Hunter}, \bibinfo{person}{Kyu-Young Kim}, \bibinfo{person}{Alan Lou}, \bibinfo{person}{Caleb~R. Matthews}, \bibinfo{person}{Ivan Villa-Renteria}, \bibinfo{person}{Jerry~Huayang Tang}, \bibinfo{person}{Claire Tang}, \bibinfo{person}{Fei Xia}, \bibinfo{person}{Silvio Savarese}, \bibinfo{person}{Hyowon Gweon}, \bibinfo{person}{C.~Karen Liu}, \bibinfo{person}{Jiajun Wu}, {and} \bibinfo{person}{Li Fei-Fei}.} \bibinfo{year}{2022}\natexlab{}.
\newblock \showarticletitle{BEHAVIOR-1K: A Benchmark for Embodied AI with 1, 000 Everyday Activities and Realistic Simulation.}. In \bibinfo{booktitle}{\emph{CORL}}, Vol.~\bibinfo{volume}{205}. \bibinfo{pages}{80--93}.
\newblock


\bibitem[Li et~al\mbox{.}(2023)]%
        {li2023matrixcity}
\bibfield{author}{\bibinfo{person}{Yixuan Li}, \bibinfo{person}{Lihan Jiang}, \bibinfo{person}{Linning Xu}, \bibinfo{person}{Yuanbo Xiangli}, \bibinfo{person}{Zhenzhi Wang}, \bibinfo{person}{Dahua Lin}, {and} \bibinfo{person}{Bo Dai}.} \bibinfo{year}{2023}\natexlab{}.
\newblock \showarticletitle{Matrixcity: A large-scale city dataset for city-scale neural rendering and beyond}. In \bibinfo{booktitle}{\emph{ICCV}}. \bibinfo{pages}{3205--3215}.
\newblock


\bibitem[Li and Snavely(2018a)]%
        {DBLP:conf/cvpr/LiS18}
\bibfield{author}{\bibinfo{person}{Zhengqi Li} {and} \bibinfo{person}{Noah Snavely}.} \bibinfo{year}{2018}\natexlab{a}.
\newblock \showarticletitle{MegaDepth: Learning Single-View Depth Prediction From Internet Photos}. In \bibinfo{booktitle}{\emph{{CVPR}}}. \bibinfo{pages}{2041--2050}.
\newblock


\bibitem[Li and Snavely(2018b)]%
        {MegaDepthLi18}
\bibfield{author}{\bibinfo{person}{Zhengqi Li} {and} \bibinfo{person}{Noah Snavely}.} \bibinfo{year}{2018}\natexlab{b}.
\newblock \showarticletitle{MegaDepth: Learning Single-View Depth Prediction from Internet Photos}. In \bibinfo{booktitle}{\emph{CVPR}}. \bibinfo{pages}{2041--2050}.
\newblock


\bibitem[Ling et~al\mbox{.}(2024)]%
        {ling2024dl3dv}
\bibfield{author}{\bibinfo{person}{Lu Ling}, \bibinfo{person}{Yichen Sheng}, \bibinfo{person}{Zhi Tu}, \bibinfo{person}{Wentian Zhao}, \bibinfo{person}{Cheng Xin}, \bibinfo{person}{Kun Wan}, \bibinfo{person}{Lantao Yu}, \bibinfo{person}{Qianyu Guo}, \bibinfo{person}{Zixun Yu}, \bibinfo{person}{Yawen Lu}, {et~al\mbox{.}}} \bibinfo{year}{2024}\natexlab{}.
\newblock \showarticletitle{Dl3dv-10k: A large-scale scene dataset for deep learning-based 3d vision}. In \bibinfo{booktitle}{\emph{CVPR}}. \bibinfo{pages}{22160--22169}.
\newblock


\bibitem[Long et~al\mbox{.}(2024)]%
        {DBLP:conf/cvpr/LongGLLDLMZHTW24}
\bibfield{author}{\bibinfo{person}{Xiaoxiao Long}, \bibinfo{person}{Yuan{-}Chen Guo}, \bibinfo{person}{Cheng Lin}, \bibinfo{person}{Yuan Liu}, \bibinfo{person}{Zhiyang Dou}, \bibinfo{person}{Lingjie Liu}, \bibinfo{person}{Yuexin Ma}, \bibinfo{person}{Song{-}Hai Zhang}, \bibinfo{person}{Marc Habermann}, \bibinfo{person}{Christian Theobalt}, {and} \bibinfo{person}{Wenping Wang}.} \bibinfo{year}{2024}\natexlab{}.
\newblock \showarticletitle{Wonder3D: Single Image to 3D Using Cross-Domain Diffusion}. In \bibinfo{booktitle}{\emph{{CVPR}}}. \bibinfo{pages}{9970--9980}.
\newblock


\bibitem[Lu et~al\mbox{.}(2024)]%
        {scaffoldgs}
\bibfield{author}{\bibinfo{person}{Tao Lu}, \bibinfo{person}{Mulin Yu}, \bibinfo{person}{Linning Xu}, \bibinfo{person}{Yuanbo Xiangli}, \bibinfo{person}{Limin Wang}, \bibinfo{person}{Dahua Lin}, {and} \bibinfo{person}{Bo Dai}.} \bibinfo{year}{2024}\natexlab{}.
\newblock \showarticletitle{Scaffold-GS: Structured 3D Gaussians for View-Adaptive Rendering}.
\newblock \bibinfo{journal}{\emph{CVPR}} (\bibinfo{year}{2024}).
\newblock


\bibitem[Martin{-}Brualla et~al\mbox{.}(2021)]%
        {Martin-BruallaR21}
\bibfield{author}{\bibinfo{person}{Ricardo Martin{-}Brualla}, \bibinfo{person}{Noha Radwan}, \bibinfo{person}{Mehdi S.~M. Sajjadi}, \bibinfo{person}{Jonathan~T. Barron}, \bibinfo{person}{Alexey Dosovitskiy}, {and} \bibinfo{person}{Daniel Duckworth}.} \bibinfo{year}{2021}\natexlab{}.
\newblock \showarticletitle{NeRF in the Wild: Neural Radiance Fields for Unconstrained Photo Collections}. In \bibinfo{booktitle}{\emph{{CVPR}}}. \bibinfo{pages}{7210--7219}.
\newblock


\bibitem[Mehl et~al\mbox{.}(2023)]%
        {Mehl2023_Spring}
\bibfield{author}{\bibinfo{person}{Lukas Mehl}, \bibinfo{person}{Jenny Schmalfuss}, \bibinfo{person}{Azin Jahedi}, \bibinfo{person}{Yaroslava Nalivayko}, {and} \bibinfo{person}{Andr\'es Bruhn}.} \bibinfo{year}{2023}\natexlab{}.
\newblock \showarticletitle{Spring: A High-Resolution High-Detail Dataset and Benchmark for Scene Flow, Optical Flow and Stereo}. In \bibinfo{booktitle}{\emph{CVPR}}. \bibinfo{pages}{4981--4991}.
\newblock


\bibitem[Melekhov et~al\mbox{.}(2019)]%
        {DBLP:conf/wacv/MelekhovTSPRK19}
\bibfield{author}{\bibinfo{person}{Iaroslav Melekhov}, \bibinfo{person}{Aleksei Tiulpin}, \bibinfo{person}{Torsten Sattler}, \bibinfo{person}{Marc Pollefeys}, \bibinfo{person}{Esa Rahtu}, {and} \bibinfo{person}{Juho Kannala}.} \bibinfo{year}{2019}\natexlab{}.
\newblock \showarticletitle{DGC-Net: Dense Geometric Correspondence Network}. In \bibinfo{booktitle}{\emph{{WACV}}}. \bibinfo{pages}{1034--1042}.
\newblock


\bibitem[Mildenhall et~al\mbox{.}(2020)]%
        {nerf}
\bibfield{author}{\bibinfo{person}{Ben Mildenhall}, \bibinfo{person}{Pratul~P. Srinivasan}, \bibinfo{person}{Matthew Tancik}, \bibinfo{person}{Jonathan~T. Barron}, \bibinfo{person}{Ravi Ramamoorthi}, {and} \bibinfo{person}{Ren Ng}.} \bibinfo{year}{2020}\natexlab{}.
\newblock \showarticletitle{NeRF: Representing Scenes as Neural Radiance Fields for View Synthesis}. In \bibinfo{booktitle}{\emph{{ECCV}}}.
\newblock


\bibitem[Niklaus et~al\mbox{.}(2019)]%
        {Niklaus_TOG_2019}
\bibfield{author}{\bibinfo{person}{Simon Niklaus}, \bibinfo{person}{Long Mai}, \bibinfo{person}{Jimei Yang}, {and} \bibinfo{person}{Feng Liu}.} \bibinfo{year}{2019}\natexlab{}.
\newblock \showarticletitle{3D Ken Burns Effect from a Single Image}.
\newblock \bibinfo{journal}{\emph{{ACM} Trans. Graph.}} \bibinfo{volume}{38}, \bibinfo{number}{6} (\bibinfo{year}{2019}), \bibinfo{pages}{184:1--184:15}.
\newblock


\bibitem[Oord et~al\mbox{.}(2018)]%
        {Oord_Li_Vinyals_2018}
\bibfield{author}{\bibinfo{person}{Aaronvanden Oord}, \bibinfo{person}{Yazhe Li}, {and} \bibinfo{person}{Oriol Vinyals}.} \bibinfo{year}{2018}\natexlab{}.
\newblock \showarticletitle{Representation Learning with Contrastive Predictive Coding}.
\newblock \bibinfo{journal}{\emph{arXiv preprint arXiv:1807.03748}} (\bibinfo{year}{2018}).
\newblock


\bibitem[Piccinelli et~al\mbox{.}(2024)]%
        {DBLP:conf/cvpr/PiccinelliYSSLG24}
\bibfield{author}{\bibinfo{person}{Luigi Piccinelli}, \bibinfo{person}{Yung{-}Hsu Yang}, \bibinfo{person}{Christos Sakaridis}, \bibinfo{person}{Mattia Seg{\`{u}}}, \bibinfo{person}{Siyuan Li}, \bibinfo{person}{Luc~Van Gool}, {and} \bibinfo{person}{Fisher Yu}.} \bibinfo{year}{2024}\natexlab{}.
\newblock \showarticletitle{UniDepth: Universal Monocular Metric Depth Estimation}. In \bibinfo{booktitle}{\emph{CVPR}}. \bibinfo{pages}{10106--10116}.
\newblock


\bibitem[Raistrick et~al\mbox{.}(2024)]%
        {infinigen2024indoors}
\bibfield{author}{\bibinfo{person}{Alexander Raistrick}, \bibinfo{person}{Lingjie Mei}, \bibinfo{person}{Karhan Kayan}, \bibinfo{person}{David Yan}, \bibinfo{person}{Yiming Zuo}, \bibinfo{person}{Beining Han}, \bibinfo{person}{Hongyu Wen}, \bibinfo{person}{Meenal Parakh}, \bibinfo{person}{Stamatis Alexandropoulos}, \bibinfo{person}{Lahav Lipson}, \bibinfo{person}{Zeyu Ma}, {and} \bibinfo{person}{Jia Deng}.} \bibinfo{year}{2024}\natexlab{}.
\newblock \showarticletitle{Infinigen Indoors: Photorealistic Indoor Scenes using Procedural Generation}. In \bibinfo{booktitle}{\emph{CVPR}}. \bibinfo{pages}{21783--21794}.
\newblock


\bibitem[Ranftl et~al\mbox{.}(2021)]%
        {DBLP:conf/iccv/RanftlBK21}
\bibfield{author}{\bibinfo{person}{Ren{\'{e}} Ranftl}, \bibinfo{person}{Alexey Bochkovskiy}, {and} \bibinfo{person}{Vladlen Koltun}.} \bibinfo{year}{2021}\natexlab{}.
\newblock \showarticletitle{Vision Transformers for Dense Prediction}. In \bibinfo{booktitle}{\emph{{ICCV}}}. \bibinfo{pages}{12159--12168}.
\newblock


\bibitem[Ranftl et~al\mbox{.}(2022)]%
        {DBLP:journals/pami/RanftlLHSK22}
\bibfield{author}{\bibinfo{person}{Ren{\'{e}} Ranftl}, \bibinfo{person}{Katrin Lasinger}, \bibinfo{person}{David Hafner}, \bibinfo{person}{Konrad Schindler}, {and} \bibinfo{person}{Vladlen Koltun}.} \bibinfo{year}{2022}\natexlab{}.
\newblock \showarticletitle{Towards Robust Monocular Depth Estimation: Mixing Datasets for Zero-Shot Cross-Dataset Transfer}.
\newblock \bibinfo{journal}{\emph{{IEEE} Trans. Pattern Anal. Mach. Intell.}} \bibinfo{volume}{44}, \bibinfo{number}{3} (\bibinfo{year}{2022}), \bibinfo{pages}{1623--1637}.
\newblock


\bibitem[Reizenstein et~al\mbox{.}(2021)]%
        {Reizenstein_Shapovalov_Henzler_Sbordone_Labatut_Novotny_2021}
\bibfield{author}{\bibinfo{person}{Jeremy Reizenstein}, \bibinfo{person}{Roman Shapovalov}, \bibinfo{person}{Philipp Henzler}, \bibinfo{person}{Luca Sbordone}, \bibinfo{person}{Patrick Labatut}, {and} \bibinfo{person}{David Novotny}.} \bibinfo{year}{2021}\natexlab{}.
\newblock \showarticletitle{Common Objects in 3D: Large-Scale Learning and Evaluation of Real-Life 3D Category Reconstruction}. In \bibinfo{booktitle}{\emph{ICCV}}. \bibinfo{pages}{10881--10891}.
\newblock


\bibitem[Richter et~al\mbox{.}(2016)]%
        {richter2016playing}
\bibfield{author}{\bibinfo{person}{Stephan~R Richter}, \bibinfo{person}{Vibhav Vineet}, \bibinfo{person}{Stefan Roth}, {and} \bibinfo{person}{Vladlen Koltun}.} \bibinfo{year}{2016}\natexlab{}.
\newblock \showarticletitle{Playing for data: Ground truth from computer games}. In \bibinfo{booktitle}{\emph{ECCV}}, Vol.~\bibinfo{volume}{9906}. \bibinfo{pages}{102--118}.
\newblock


\bibitem[Roberts et~al\mbox{.}(2021)]%
        {roberts:2021}
\bibfield{author}{\bibinfo{person}{Mike Roberts}, \bibinfo{person}{Jason Ramapuram}, \bibinfo{person}{Anurag Ranjan}, \bibinfo{person}{Atulit Kumar}, \bibinfo{person}{Miguel~Angel Bautista}, \bibinfo{person}{Nathan Paczan}, \bibinfo{person}{Russ Webb}, {and} \bibinfo{person}{Joshua~M. Susskind}.} \bibinfo{year}{2021}\natexlab{}.
\newblock \showarticletitle{{Hypersim}: {A} Photorealistic Synthetic Dataset for Holistic Indoor Scene Understanding}. In \bibinfo{booktitle}{\emph{ICCV}}. \bibinfo{pages}{10892--10902}.
\newblock


\bibitem[Ros et~al\mbox{.}(2016)]%
        {Ros_2016_CVPR}
\bibfield{author}{\bibinfo{person}{German Ros}, \bibinfo{person}{Laura Sellart}, \bibinfo{person}{Joanna Materzynska}, \bibinfo{person}{David Vazquez}, {and} \bibinfo{person}{Antonio~M. Lopez}.} \bibinfo{year}{2016}\natexlab{}.
\newblock \showarticletitle{The SYNTHIA Dataset: A Large Collection of Synthetic Images for Semantic Segmentation of Urban Scenes}. In \bibinfo{booktitle}{\emph{CVPR}}. \bibinfo{pages}{3234--3243}.
\newblock


\bibitem[Sarlin et~al\mbox{.}(2020)]%
        {DBLP:conf/cvpr/SarlinDMR20}
\bibfield{author}{\bibinfo{person}{Paul{-}Edouard Sarlin}, \bibinfo{person}{Daniel DeTone}, \bibinfo{person}{Tomasz Malisiewicz}, {and} \bibinfo{person}{Andrew Rabinovich}.} \bibinfo{year}{2020}\natexlab{}.
\newblock \showarticletitle{SuperGlue: Learning Feature Matching With Graph Neural Networks}. In \bibinfo{booktitle}{\emph{{CVPR}}}. \bibinfo{pages}{4937--4946}.
\newblock


\bibitem[Savva et~al\mbox{.}(2019)]%
        {Savva_Kadian_Maksymets_Zhao_Wijmans_Jain_Straub_Liu_Koltun_Malik_et_al._2019}
\bibfield{author}{\bibinfo{person}{Manolis Savva}, \bibinfo{person}{Abhishek Kadian}, \bibinfo{person}{Oleksandr Maksymets}, \bibinfo{person}{Yili Zhao}, \bibinfo{person}{Erik Wijmans}, \bibinfo{person}{Bhavana Jain}, \bibinfo{person}{Julian Straub}, \bibinfo{person}{Jia Liu}, \bibinfo{person}{Vladlen Koltun}, \bibinfo{person}{Jitendra Malik}, \bibinfo{person}{Devi Parikh}, {and} \bibinfo{person}{Dhruv Batra}.} \bibinfo{year}{2019}\natexlab{}.
\newblock \showarticletitle{Habitat: A Platform for Embodied AI Research}. In \bibinfo{booktitle}{\emph{ICCV}}. \bibinfo{pages}{9338--9346}.
\newblock


\bibitem[Schr{\"{o}}ppel et~al\mbox{.}(2022)]%
        {staticthings3d}
\bibfield{author}{\bibinfo{person}{Philipp Schr{\"{o}}ppel}, \bibinfo{person}{Jan Bechtold}, \bibinfo{person}{Artemij Amiranashvili}, {and} \bibinfo{person}{Thomas Brox}.} \bibinfo{year}{2022}\natexlab{}.
\newblock \showarticletitle{A Benchmark and a Baseline for Robust Multi-view Depth Estimation}. In \bibinfo{booktitle}{\emph{3DV}}. \bibinfo{pages}{637--645}.
\newblock


\bibitem[Smart et~al\mbox{.}(2024)]%
        {smart2024splatt3r}
\bibfield{author}{\bibinfo{person}{Brandon Smart}, \bibinfo{person}{Chuanxia Zheng}, \bibinfo{person}{Iro Laina}, {and} \bibinfo{person}{Victor~Adrian Prisacariu}.} \bibinfo{year}{2024}\natexlab{}.
\newblock \showarticletitle{Splatt3R: Zero-shot Gaussian Splatting from Uncalibrated Image Pairs}.
\newblock \bibinfo{journal}{\emph{arxiv preprint arXiv:2408.13912}} (\bibinfo{year}{2024}).
\newblock


\bibitem[Sun et~al\mbox{.}(2021)]%
        {DBLP:conf/cvpr/SunSWBZ21}
\bibfield{author}{\bibinfo{person}{Jiaming Sun}, \bibinfo{person}{Zehong Shen}, \bibinfo{person}{Yuang Wang}, \bibinfo{person}{Hujun Bao}, {and} \bibinfo{person}{Xiaowei Zhou}.} \bibinfo{year}{2021}\natexlab{}.
\newblock \showarticletitle{LoFTR: Detector-Free Local Feature Matching With Transformers}. In \bibinfo{booktitle}{\emph{{CVPR}}}. \bibinfo{pages}{8922--8931}.
\newblock


\bibitem[Sun et~al\mbox{.}(2020)]%
        {Sun_2020_CVPR}
\bibfield{author}{\bibinfo{person}{Pei Sun}, \bibinfo{person}{Henrik Kretzschmar}, \bibinfo{person}{Xerxes Dotiwalla}, \bibinfo{person}{Aurelien Chouard}, \bibinfo{person}{Vijaysai Patnaik}, \bibinfo{person}{Paul Tsui}, \bibinfo{person}{James Guo}, \bibinfo{person}{Yin Zhou}, \bibinfo{person}{Yuning Chai}, \bibinfo{person}{Benjamin Caine}, \bibinfo{person}{Vijay Vasudevan}, \bibinfo{person}{Wei Han}, \bibinfo{person}{Jiquan Ngiam}, \bibinfo{person}{Hang Zhao}, \bibinfo{person}{Aleksei Timofeev}, \bibinfo{person}{Scott Ettinger}, \bibinfo{person}{Maxim Krivokon}, \bibinfo{person}{Amy Gao}, \bibinfo{person}{Aditya Joshi}, \bibinfo{person}{Yu Zhang}, \bibinfo{person}{Jonathon Shlens}, \bibinfo{person}{Zhifeng Chen}, {and} \bibinfo{person}{Dragomir Anguelov}.} \bibinfo{year}{2020}\natexlab{}.
\newblock \showarticletitle{Scalability in Perception for Autonomous Driving: Waymo Open Dataset}. In \bibinfo{booktitle}{\emph{CVPR}}. \bibinfo{pages}{2443--2451}.
\newblock


\bibitem[Toft et~al\mbox{.}(2020)]%
        {DBLP:conf/eccv/ToftTSKB20}
\bibfield{author}{\bibinfo{person}{Carl Toft}, \bibinfo{person}{Daniyar Turmukhambetov}, \bibinfo{person}{Torsten Sattler}, \bibinfo{person}{Fredrik Kahl}, {and} \bibinfo{person}{Gabriel~J. Brostow}.} \bibinfo{year}{2020}\natexlab{}.
\newblock \showarticletitle{Single-Image Depth Prediction Makes Feature Matching Easier}. In \bibinfo{booktitle}{\emph{{ECCV}}}, Vol.~\bibinfo{volume}{12361}. \bibinfo{pages}{473--492}.
\newblock


\bibitem[Tosi et~al\mbox{.}(2021)]%
        {Tosi2021CVPR}
\bibfield{author}{\bibinfo{person}{Fabio Tosi}, \bibinfo{person}{Yiyi Liao}, \bibinfo{person}{Carolin Schmitt}, {and} \bibinfo{person}{Andreas Geiger}.} \bibinfo{year}{2021}\natexlab{}.
\newblock \showarticletitle{SMD-Nets: Stereo Mixture Density Networks}. In \bibinfo{booktitle}{\emph{CVPR}}. \bibinfo{pages}{8942--8952}.
\newblock


\bibitem[Truong et~al\mbox{.}(2021)]%
        {DBLP:conf/cvpr/TruongDGT21}
\bibfield{author}{\bibinfo{person}{Prune Truong}, \bibinfo{person}{Martin Danelljan}, \bibinfo{person}{Luc~Van Gool}, {and} \bibinfo{person}{Radu Timofte}.} \bibinfo{year}{2021}\natexlab{}.
\newblock \showarticletitle{Learning Accurate Dense Correspondences and When To Trust Them}. In \bibinfo{booktitle}{\emph{{CVPR}}}. \bibinfo{pages}{5714--5724}.
\newblock


\bibitem[Truong et~al\mbox{.}(2020)]%
        {DBLP:conf/cvpr/TruongDT20}
\bibfield{author}{\bibinfo{person}{Prune Truong}, \bibinfo{person}{Martin Danelljan}, {and} \bibinfo{person}{Radu Timofte}.} \bibinfo{year}{2020}\natexlab{}.
\newblock \showarticletitle{GLU-Net: Global-Local Universal Network for Dense Flow and Correspondences}. In \bibinfo{booktitle}{\emph{{CVPR}}}. \bibinfo{pages}{6257--6267}.
\newblock


\bibitem[Truong et~al\mbox{.}(2023)]%
        {DBLP:journals/pami/TruongDTG23}
\bibfield{author}{\bibinfo{person}{Prune Truong}, \bibinfo{person}{Martin Danelljan}, \bibinfo{person}{Radu Timofte}, {and} \bibinfo{person}{Luc~Van Gool}.} \bibinfo{year}{2023}\natexlab{}.
\newblock \showarticletitle{PDC-Net+: Enhanced Probabilistic Dense Correspondence Network}.
\newblock \bibinfo{journal}{\emph{{IEEE} Trans. Pattern Anal. Mach. Intell.}} \bibinfo{volume}{45}, \bibinfo{number}{8} (\bibinfo{year}{2023}), \bibinfo{pages}{10247--10266}.
\newblock


\bibitem[Wan et~al\mbox{.}(2023)]%
        {DBLP:journals/corr/abs-2312-14733}
\bibfield{author}{\bibinfo{person}{Qiang Wan}, \bibinfo{person}{Zilong Huang}, \bibinfo{person}{Bingyi Kang}, \bibinfo{person}{Jiashi Feng}, {and} \bibinfo{person}{Li Zhang}.} \bibinfo{year}{2023}\natexlab{}.
\newblock \showarticletitle{Harnessing Diffusion Models for Visual Perception with Meta Prompts}.
\newblock \bibinfo{journal}{\emph{arxiv preprint arXiv:2312.14733}} (\bibinfo{year}{2023}).
\newblock


\bibitem[Wang and Agapito(2025)]%
        {wang2024spann3r}
\bibfield{author}{\bibinfo{person}{Hengyi Wang} {and} \bibinfo{person}{Lourdes Agapito}.} \bibinfo{year}{2025}\natexlab{}.
\newblock \showarticletitle{3D Reconstruction with Spatial Memory}. In \bibinfo{booktitle}{\emph{{3DV}}}.
\newblock


\bibitem[Wang et~al\mbox{.}(2025a)]%
        {vggt}
\bibfield{author}{\bibinfo{person}{Jianyuan Wang}, \bibinfo{person}{Minghao Chen}, \bibinfo{person}{Nikita Karaev}, \bibinfo{person}{Andrea Vedaldi}, \bibinfo{person}{Christian Rupprecht}, {and} \bibinfo{person}{David Novotny}.} \bibinfo{year}{2025}\natexlab{a}.
\newblock \showarticletitle{VGGT: Visual Geometry Grounded Transformer}. In \bibinfo{booktitle}{\emph{CVPR}}.
\newblock


\bibitem[Wang et~al\mbox{.}(2023b)]%
        {DBLP:conf/iccv/Wang0N23}
\bibfield{author}{\bibinfo{person}{Jianyuan Wang}, \bibinfo{person}{Christian Rupprecht}, {and} \bibinfo{person}{David Novotn{\'{y}}}.} \bibinfo{year}{2023}\natexlab{b}.
\newblock \showarticletitle{PoseDiffusion: Solving Pose Estimation via Diffusion-aided Bundle Adjustment}. In \bibinfo{booktitle}{\emph{{ICCV}}}. \bibinfo{pages}{9739--9749}.
\newblock


\bibitem[Wang and Shen(2020)]%
        {wang2020flow}
\bibfield{author}{\bibinfo{person}{Kaixuan Wang} {and} \bibinfo{person}{Shaojie Shen}.} \bibinfo{year}{2020}\natexlab{}.
\newblock \showarticletitle{Flow-motion and depth network for monocular stereo and beyond}.
\newblock \bibinfo{journal}{\emph{IEEE Robotics and Automation Letters}} \bibinfo{volume}{5}, \bibinfo{number}{2} (\bibinfo{year}{2020}), \bibinfo{pages}{3307--3314}.
\newblock


\bibitem[Wang et~al\mbox{.}(2021a)]%
        {neus}
\bibfield{author}{\bibinfo{person}{Peng Wang}, \bibinfo{person}{Lingjie Liu}, \bibinfo{person}{Yuan Liu}, \bibinfo{person}{Christian Theobalt}, \bibinfo{person}{Taku Komura}, {and} \bibinfo{person}{Wenping Wang}.} \bibinfo{year}{2021}\natexlab{a}.
\newblock \showarticletitle{NeuS: Learning Neural Implicit Surfaces by Volume Rendering for Multi-view Reconstruction}. In \bibinfo{booktitle}{\emph{NeurIPS}}. \bibinfo{pages}{27171--27183}.
\newblock


\bibitem[Wang et~al\mbox{.}(2021b)]%
        {wang2021irs}
\bibfield{author}{\bibinfo{person}{Qiang Wang}, \bibinfo{person}{Shizhen Zheng}, \bibinfo{person}{Qingsong Yan}, \bibinfo{person}{Fei Deng}, \bibinfo{person}{Kaiyong Zhao}, {and} \bibinfo{person}{Xiaowen Chu}.} \bibinfo{year}{2021}\natexlab{b}.
\newblock \showarticletitle{Irs: A large naturalistic indoor robotics stereo dataset to train deep models for disparity and surface normal estimation}. In \bibinfo{booktitle}{\emph{ICME}}. \bibinfo{pages}{1--6}.
\newblock


\bibitem[Wang et~al\mbox{.}(2020a)]%
        {DBLP:conf/eccv/WangZHS20}
\bibfield{author}{\bibinfo{person}{Qianqian Wang}, \bibinfo{person}{Xiaowei Zhou}, \bibinfo{person}{Bharath Hariharan}, {and} \bibinfo{person}{Noah Snavely}.} \bibinfo{year}{2020}\natexlab{a}.
\newblock \showarticletitle{Learning Feature Descriptors Using Camera Pose Supervision}. In \bibinfo{booktitle}{\emph{{ECCV}}}, Vol.~\bibinfo{volume}{12346}. \bibinfo{pages}{757--774}.
\newblock


\bibitem[Wang et~al\mbox{.}(2025b)]%
        {moge}
\bibfield{author}{\bibinfo{person}{Ruicheng Wang}, \bibinfo{person}{Sicheng Xu}, \bibinfo{person}{Cassie Dai}, \bibinfo{person}{Jianfeng Xiang}, \bibinfo{person}{Yu Deng}, \bibinfo{person}{Xin Tong}, {and} \bibinfo{person}{Jiaolong Yang}.} \bibinfo{year}{2025}\natexlab{b}.
\newblock \showarticletitle{MoGe: Unlocking Accurate Monocular Geometry Estimation for Open-Domain Images with Optimal Training Supervision}. In \bibinfo{booktitle}{\emph{CVPR}}.
\newblock


\bibitem[Wang et~al\mbox{.}(2024)]%
        {dust3r}
\bibfield{author}{\bibinfo{person}{Shuzhe Wang}, \bibinfo{person}{Vincent Leroy}, \bibinfo{person}{Yohann Cabon}, \bibinfo{person}{Boris Chidlovskii}, {and} \bibinfo{person}{J{\'{e}}r{\^{o}}me Revaud}.} \bibinfo{year}{2024}\natexlab{}.
\newblock \showarticletitle{DUSt3R: Geometric 3D Vision Made Easy}. In \bibinfo{booktitle}{\emph{{CVPR}}}. \bibinfo{pages}{20697--20709}.
\newblock


\bibitem[Wang et~al\mbox{.}(2020b)]%
        {tartanair2020iros}
\bibfield{author}{\bibinfo{person}{Wenshan Wang}, \bibinfo{person}{Delong Zhu}, \bibinfo{person}{Xiangwei Wang}, \bibinfo{person}{Yaoyu Hu}, \bibinfo{person}{Yuheng Qiu}, \bibinfo{person}{Chen Wang}, \bibinfo{person}{Yafei Hu}, \bibinfo{person}{Ashish Kapoor}, {and} \bibinfo{person}{Sebastian Scherer}.} \bibinfo{year}{2020}\natexlab{b}.
\newblock \showarticletitle{TartanAir: A Dataset to Push the Limits of Visual SLAM}.
\newblock  (\bibinfo{year}{2020}), \bibinfo{pages}{4909--4916}.
\newblock


\bibitem[Wang et~al\mbox{.}(2015)]%
        {DBLP:conf/cvpr/WangFG15}
\bibfield{author}{\bibinfo{person}{Xiaolong Wang}, \bibinfo{person}{David~F. Fouhey}, {and} \bibinfo{person}{Abhinav Gupta}.} \bibinfo{year}{2015}\natexlab{}.
\newblock \showarticletitle{Designing deep networks for surface normal estimation}. In \bibinfo{booktitle}{\emph{{CVPR}}}. \bibinfo{pages}{539--547}.
\newblock


\bibitem[Wang et~al\mbox{.}(2023a)]%
        {neus2}
\bibfield{author}{\bibinfo{person}{Yiming Wang}, \bibinfo{person}{Qin Han}, \bibinfo{person}{Marc Habermann}, \bibinfo{person}{Kostas Daniilidis}, \bibinfo{person}{Christian Theobalt}, {and} \bibinfo{person}{Lingjie Liu}.} \bibinfo{year}{2023}\natexlab{a}.
\newblock \showarticletitle{NeuS2: Fast Learning of Neural Implicit Surfaces for Multi-view Reconstruction}. In \bibinfo{booktitle}{\emph{ICCV}}.
\newblock


\bibitem[Xia et~al\mbox{.}(2024)]%
        {xia2024rgbd}
\bibfield{author}{\bibinfo{person}{Hongchi Xia}, \bibinfo{person}{Yang Fu}, \bibinfo{person}{Sifei Liu}, {and} \bibinfo{person}{Xiaolong Wang}.} \bibinfo{year}{2024}\natexlab{}.
\newblock \showarticletitle{{RGBD} Objects in the Wild: Scaling Real-World 3D Object Learning from {RGB-D} Videos}. In \bibinfo{booktitle}{\emph{CVPR}}. \bibinfo{pages}{22378--22389}.
\newblock


\bibitem[Xu et~al\mbox{.}(2025)]%
        {genpercept}
\bibfield{author}{\bibinfo{person}{Guangkai Xu}, \bibinfo{person}{Yongtao Ge}, \bibinfo{person}{Mingyu Liu}, \bibinfo{person}{Chengxiang Fan}, \bibinfo{person}{Kangyang Xie}, \bibinfo{person}{Zhiyue Zhao}, \bibinfo{person}{Hao Chen}, {and} \bibinfo{person}{Chunhua Shen}.} \bibinfo{year}{2025}\natexlab{}.
\newblock \showarticletitle{What Matters When Repurposing Diffusion Models for General Dense Perception Tasks?}. In \bibinfo{booktitle}{\emph{ICLR}}.
\newblock


\bibitem[Yang et~al\mbox{.}(2025)]%
        {fast3r}
\bibfield{author}{\bibinfo{person}{Jianing Yang}, \bibinfo{person}{Alexander Sax}, \bibinfo{person}{Kevin~J. Liang}, \bibinfo{person}{Mikael Henaff}, \bibinfo{person}{Hao Tang}, \bibinfo{person}{Ang Cao}, \bibinfo{person}{Joyce Chai}, \bibinfo{person}{Franziska Meier}, {and} \bibinfo{person}{Matt Feiszli}.} \bibinfo{year}{2025}\natexlab{}.
\newblock \showarticletitle{Fast3R: Towards 3D Reconstruction of 1000+ Images in One Forward Pass}. In \bibinfo{booktitle}{\emph{CVPR}}.
\newblock


\bibitem[Yang et~al\mbox{.}(2024a)]%
        {DBLP:conf/cvpr/YangKHXFZ24}
\bibfield{author}{\bibinfo{person}{Lihe Yang}, \bibinfo{person}{Bingyi Kang}, \bibinfo{person}{Zilong Huang}, \bibinfo{person}{Xiaogang Xu}, \bibinfo{person}{Jiashi Feng}, {and} \bibinfo{person}{Hengshuang Zhao}.} \bibinfo{year}{2024}\natexlab{a}.
\newblock \showarticletitle{Depth Anything: Unleashing the Power of Large-Scale Unlabeled Data}. In \bibinfo{booktitle}{\emph{{CVPR}}}. \bibinfo{pages}{10371--10381}.
\newblock


\bibitem[Yang et~al\mbox{.}(2024b)]%
        {depth_anything_v2}
\bibfield{author}{\bibinfo{person}{Lihe Yang}, \bibinfo{person}{Bingyi Kang}, \bibinfo{person}{Zilong Huang}, \bibinfo{person}{Zhen Zhao}, \bibinfo{person}{Xiaogang Xu}, \bibinfo{person}{Jiashi Feng}, {and} \bibinfo{person}{Hengshuang Zhao}.} \bibinfo{year}{2024}\natexlab{b}.
\newblock \showarticletitle{Depth Anything V2}. In \bibinfo{booktitle}{\emph{NeurIPS}}.
\newblock


\bibitem[Yao et~al\mbox{.}(2019)]%
        {DBLP:conf/iccv/YaoJP19}
\bibfield{author}{\bibinfo{person}{Yuan Yao}, \bibinfo{person}{Yasamin Jafarian}, {and} \bibinfo{person}{Hyun~Soo Park}.} \bibinfo{year}{2019}\natexlab{}.
\newblock \showarticletitle{{MONET:} Multiview Semi-Supervised Keypoint Detection via Epipolar Divergence}. In \bibinfo{booktitle}{\emph{{ICCV}}}. \bibinfo{pages}{753--762}.
\newblock


\bibitem[Yao et~al\mbox{.}(2020)]%
        {yao2020blendedmvs}
\bibfield{author}{\bibinfo{person}{Yao Yao}, \bibinfo{person}{Zixin Luo}, \bibinfo{person}{Shiwei Li}, \bibinfo{person}{Jingyang Zhang}, \bibinfo{person}{Yufan Ren}, \bibinfo{person}{Lei Zhou}, \bibinfo{person}{Tian Fang}, {and} \bibinfo{person}{Long Quan}.} \bibinfo{year}{2020}\natexlab{}.
\newblock \showarticletitle{BlendedMVS: {A} Large-Scale Dataset for Generalized Multi-View Stereo Networks}. In \bibinfo{booktitle}{\emph{CVPR}}. \bibinfo{pages}{1787--1796}.
\newblock


\bibitem[Yariv et~al\mbox{.}(2021)]%
        {DBLP:conf/nips/YarivGKL21}
\bibfield{author}{\bibinfo{person}{Lior Yariv}, \bibinfo{person}{Jiatao Gu}, \bibinfo{person}{Yoni Kasten}, {and} \bibinfo{person}{Yaron Lipman}.} \bibinfo{year}{2021}\natexlab{}.
\newblock \showarticletitle{Volume Rendering of Neural Implicit Surfaces}. In \bibinfo{booktitle}{\emph{NeurIPS}}. \bibinfo{pages}{4805--4815}.
\newblock


\bibitem[Ye et~al\mbox{.}(2025)]%
        {noposplat}
\bibfield{author}{\bibinfo{person}{Botao Ye}, \bibinfo{person}{Sifei Liu}, \bibinfo{person}{Haofei Xu}, \bibinfo{person}{Li Xueting}, \bibinfo{person}{Marc Pollefeys}, \bibinfo{person}{Ming-Hsuan Yang}, {and} \bibinfo{person}{Peng Songyou}.} \bibinfo{year}{2025}\natexlab{}.
\newblock \showarticletitle{No Pose, No Problem: Surprisingly Simple 3D Gaussian Splats from Sparse Unposed Images}. In \bibinfo{booktitle}{\emph{{ICLR}}}.
\newblock


\bibitem[Ye et~al\mbox{.}(2024)]%
        {stablenormal}
\bibfield{author}{\bibinfo{person}{Chongjie Ye}, \bibinfo{person}{Lingteng Qiu}, \bibinfo{person}{Xiaodong Gu}, \bibinfo{person}{Qi Zuo}, \bibinfo{person}{Yushuang Wu}, \bibinfo{person}{Zilong Dong}, \bibinfo{person}{Liefeng Bo}, \bibinfo{person}{Yuliang Xiu}, {and} \bibinfo{person}{Xiaoguang Han}.} \bibinfo{year}{2024}\natexlab{}.
\newblock \showarticletitle{StableNormal: Reducing Diffusion Variance for Stable and Sharp Normal}.
\newblock \bibinfo{journal}{\emph{{ACM} Trans. Graph.}} \bibinfo{volume}{43}, \bibinfo{number}{6} (\bibinfo{year}{2024}), \bibinfo{pages}{250:1--250:18}.
\newblock


\bibitem[Yeshwanth et~al\mbox{.}(2023)]%
        {yeshwanthliu2023scannetpp}
\bibfield{author}{\bibinfo{person}{Chandan Yeshwanth}, \bibinfo{person}{Yueh-Cheng Liu}, \bibinfo{person}{Matthias Nie{\ss}ner}, {and} \bibinfo{person}{Angela Dai}.} \bibinfo{year}{2023}\natexlab{}.
\newblock \showarticletitle{ScanNet++: A High-Fidelity Dataset of 3D Indoor Scenes}. In \bibinfo{booktitle}{\emph{ICCV}}. \bibinfo{pages}{12--22}.
\newblock


\bibitem[Yifan et~al\mbox{.}(2022)]%
        {DBLP:conf/cvpr/YifanDACZ22}
\bibfield{author}{\bibinfo{person}{Wang Yifan}, \bibinfo{person}{Carl Doersch}, \bibinfo{person}{Relja Arandjelovic}, \bibinfo{person}{Jo{\~{a}}o Carreira}, {and} \bibinfo{person}{Andrew Zisserman}.} \bibinfo{year}{2022}\natexlab{}.
\newblock \showarticletitle{Input-level Inductive Biases for 3D Reconstruction}. In \bibinfo{booktitle}{\emph{{CVPR}}}. \bibinfo{pages}{6166--6176}.
\newblock


\bibitem[Yin et~al\mbox{.}(2023)]%
        {DBLP:conf/iccv/000600CYWCS23}
\bibfield{author}{\bibinfo{person}{Wei Yin}, \bibinfo{person}{Chi Zhang}, \bibinfo{person}{Hao Chen}, \bibinfo{person}{Zhipeng Cai}, \bibinfo{person}{Gang Yu}, \bibinfo{person}{Kaixuan Wang}, \bibinfo{person}{Xiaozhi Chen}, {and} \bibinfo{person}{Chunhua Shen}.} \bibinfo{year}{2023}\natexlab{}.
\newblock \showarticletitle{Metric3D: Towards Zero-shot Metric 3D Prediction from {A} Single Image}. In \bibinfo{booktitle}{\emph{{ICCV}}}. \bibinfo{pages}{9009--9019}.
\newblock


\bibitem[Yu et~al\mbox{.}(2024)]%
        {Yu2023MipSplatting}
\bibfield{author}{\bibinfo{person}{Zehao Yu}, \bibinfo{person}{Anpei Chen}, \bibinfo{person}{Binbin Huang}, \bibinfo{person}{Torsten Sattler}, {and} \bibinfo{person}{Andreas Geiger}.} \bibinfo{year}{2024}\natexlab{}.
\newblock \showarticletitle{Mip-Splatting: Alias-free 3D Gaussian Splatting}.
\newblock \bibinfo{journal}{\emph{CVPR}} (\bibinfo{year}{2024}).
\newblock


\bibitem[Zamir et~al\mbox{.}(2018)]%
        {zamir2018taskonomy}
\bibfield{author}{\bibinfo{person}{Amir~R Zamir}, \bibinfo{person}{Alexander Sax}, \bibinfo{person}{}, \bibinfo{person}{William~B Shen}, \bibinfo{person}{Leonidas Guibas}, \bibinfo{person}{Jitendra Malik}, {and} \bibinfo{person}{Silvio Savarese}.} \bibinfo{year}{2018}\natexlab{}.
\newblock \showarticletitle{Taskonomy: Disentangling Task Transfer Learning}. In \bibinfo{booktitle}{\emph{CVPR}}. \bibinfo{pages}{3712--3722}.
\newblock


\bibitem[Zhang et~al\mbox{.}(2024)]%
        {DBLP:conf/iclr/0001LKYRT24}
\bibfield{author}{\bibinfo{person}{Jason~Y. Zhang}, \bibinfo{person}{Amy Lin}, \bibinfo{person}{Moneish Kumar}, \bibinfo{person}{Tzu{-}Hsuan Yang}, \bibinfo{person}{Deva Ramanan}, {and} \bibinfo{person}{Shubham Tulsiani}.} \bibinfo{year}{2024}\natexlab{}.
\newblock \showarticletitle{Cameras as Rays: Pose Estimation via Ray Diffusion}. In \bibinfo{booktitle}{\emph{{ICLR}}}.
\newblock


\bibitem[Zheng et~al\mbox{.}(2020)]%
        {Structured3D}
\bibfield{author}{\bibinfo{person}{Jia Zheng}, \bibinfo{person}{Junfei Zhang}, \bibinfo{person}{Jing Li}, \bibinfo{person}{Rui Tang}, \bibinfo{person}{Shenghua Gao}, {and} \bibinfo{person}{Zihan Zhou}.} \bibinfo{year}{2020}\natexlab{}.
\newblock \showarticletitle{Structured3D: A Large Photo-realistic Dataset for Structured 3D Modeling}. In \bibinfo{booktitle}{\emph{ECCV}}, Vol.~\bibinfo{volume}{12354}. \bibinfo{pages}{519--535}.
\newblock


\bibitem[Zhou et~al\mbox{.}(2021)]%
        {DBLP:conf/cvpr/ZhouSL21}
\bibfield{author}{\bibinfo{person}{Qunjie Zhou}, \bibinfo{person}{Torsten Sattler}, {and} \bibinfo{person}{Laura Leal{-}Taix{\'{e}}}.} \bibinfo{year}{2021}\natexlab{}.
\newblock \showarticletitle{Patch2Pix: Epipolar-Guided Pixel-Level Correspondences}. In \bibinfo{booktitle}{\emph{{CVPR}}}. \bibinfo{pages}{4669--4678}.
\newblock


\bibitem[Zhu and Liu(2023)]%
        {DBLP:conf/cvpr/Zhu023a}
\bibfield{author}{\bibinfo{person}{Shengjie Zhu} {and} \bibinfo{person}{Xiaoming Liu}.} \bibinfo{year}{2023}\natexlab{}.
\newblock \showarticletitle{PMatch: Paired Masked Image Modeling for Dense Geometric Matching}. In \bibinfo{booktitle}{\emph{{CVPR}}}. \bibinfo{pages}{21909--21918}.
\newblock


\end{thebibliography}

\clearpage

\begin{figure*}
    \centering
    \includegraphics[width=0.85\linewidth]{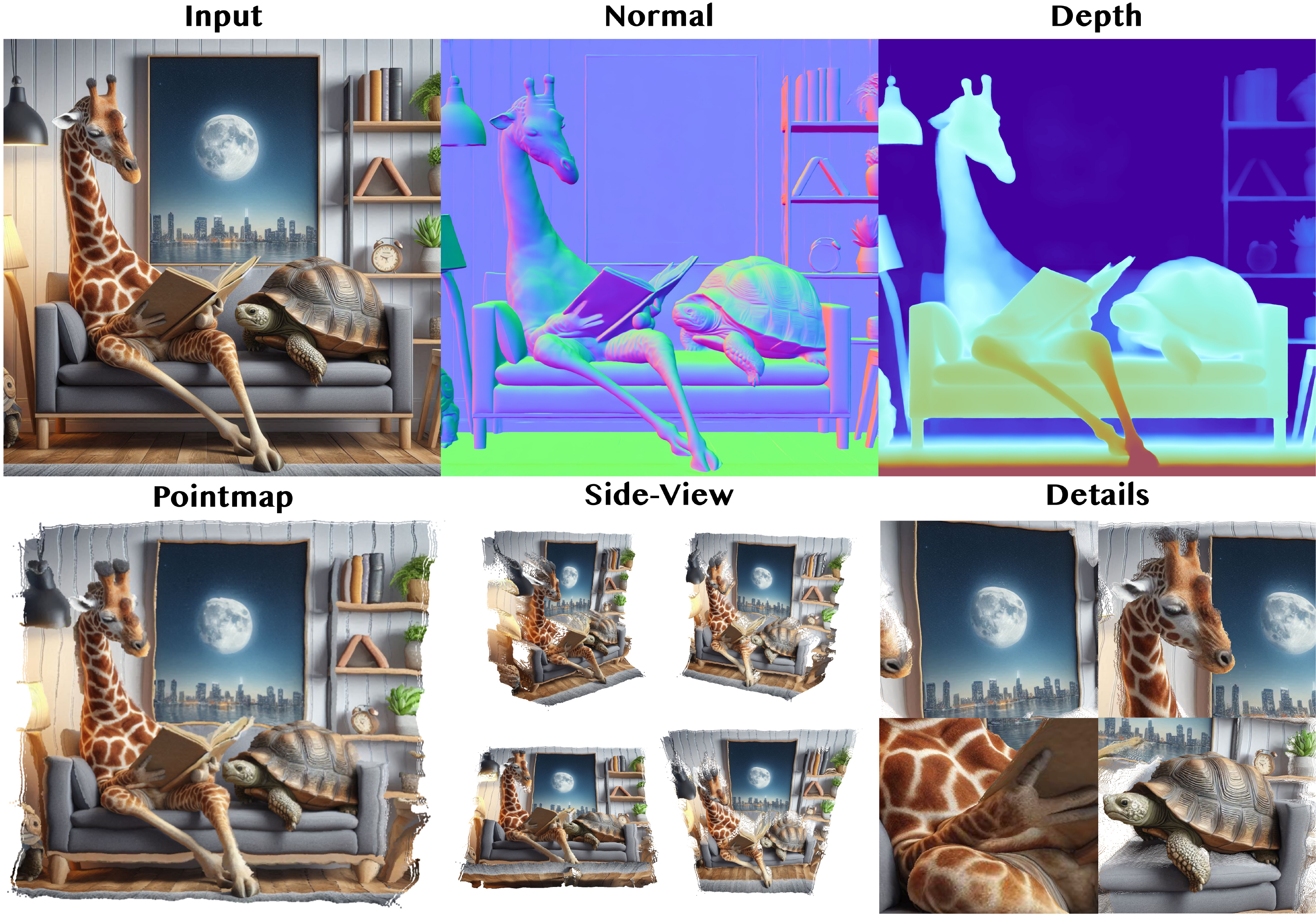}
    \caption{High-quality geometric predictions for high-resolution (2K) inputs. Please zoom in to better observe the fine-grained details.}
    \label{fig:2k}
    \vspace{-6mm}
\end{figure*}

\clearpage

\begin{figure*}
    \centering
    \includegraphics[width=0.95\linewidth]{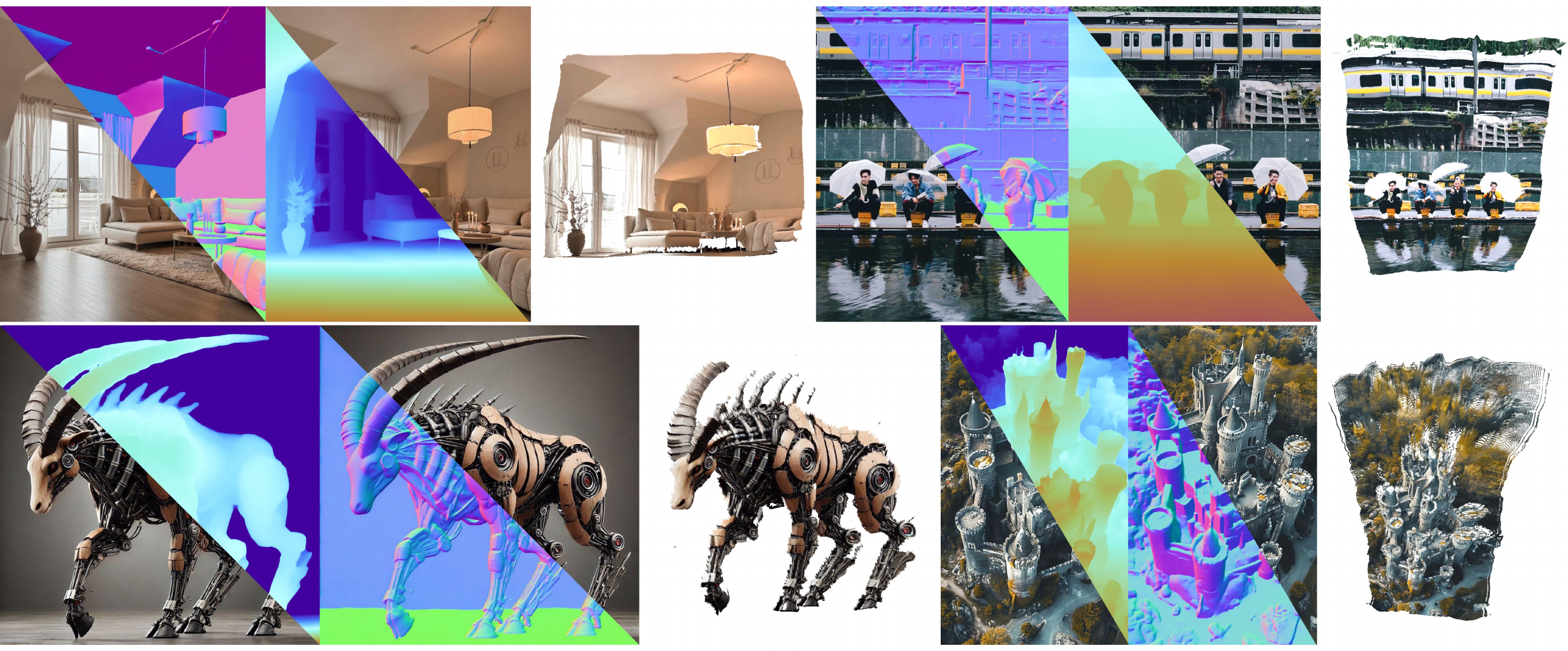}
    \caption{High-quality unified geometric predictions for various scenarios. We demonstrate accurate normal and depth predictions with high-quality 3D pointmaps for challenging object-centric, indoor and outdoor scenes.}
    \label{fig:sup_vis}
\end{figure*}

\clearpage

\appendix

\section{Implementation Details}
\begin{figure*}
    \centering
    \includegraphics[width=0.95\linewidth]{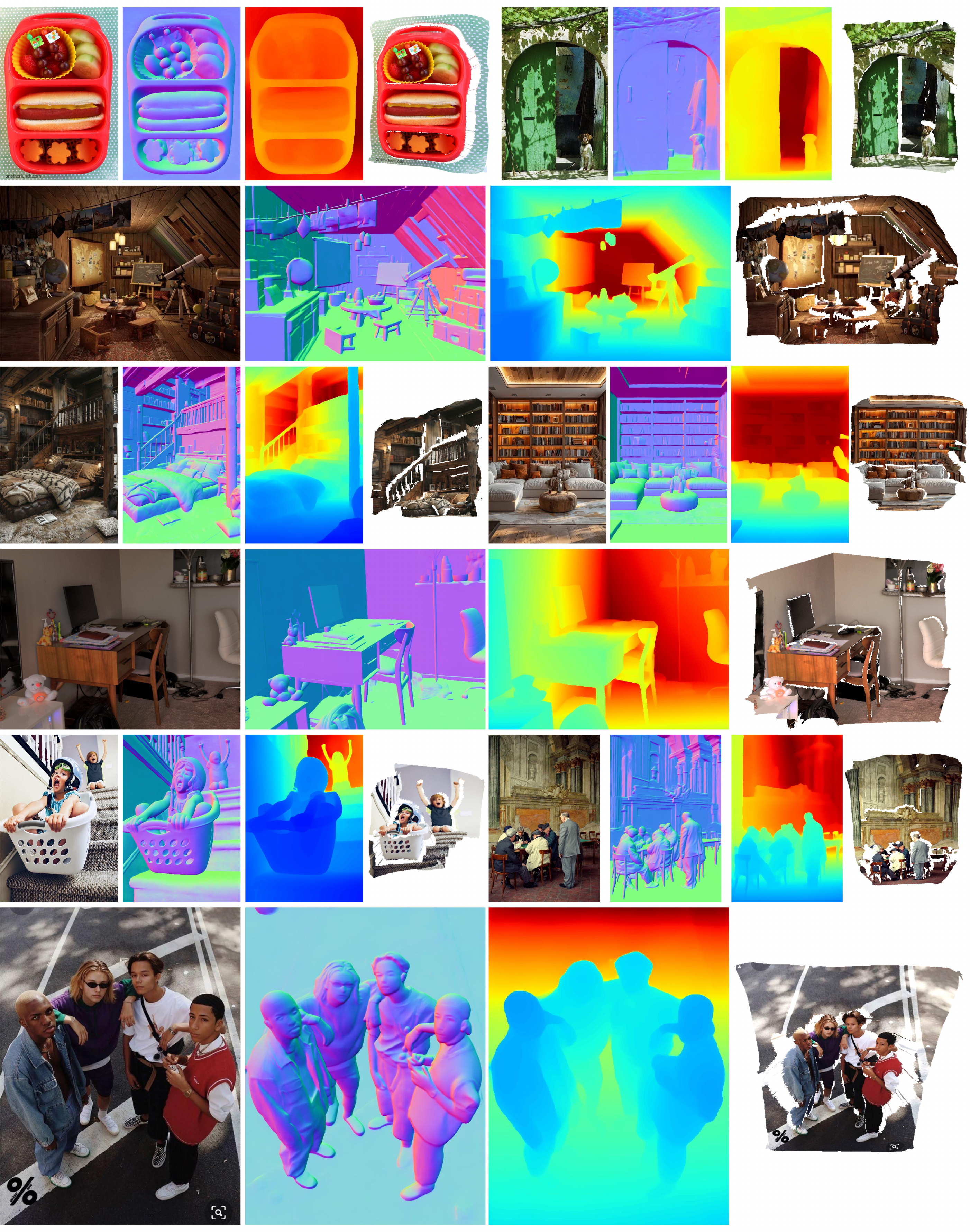}
    \caption{Additional visual results for unified geometric predictions of monocular inputs by our methods}
    \label{fig:more1}
\end{figure*}

\begin{figure*}
    \centering
    \includegraphics[width=0.95\linewidth]{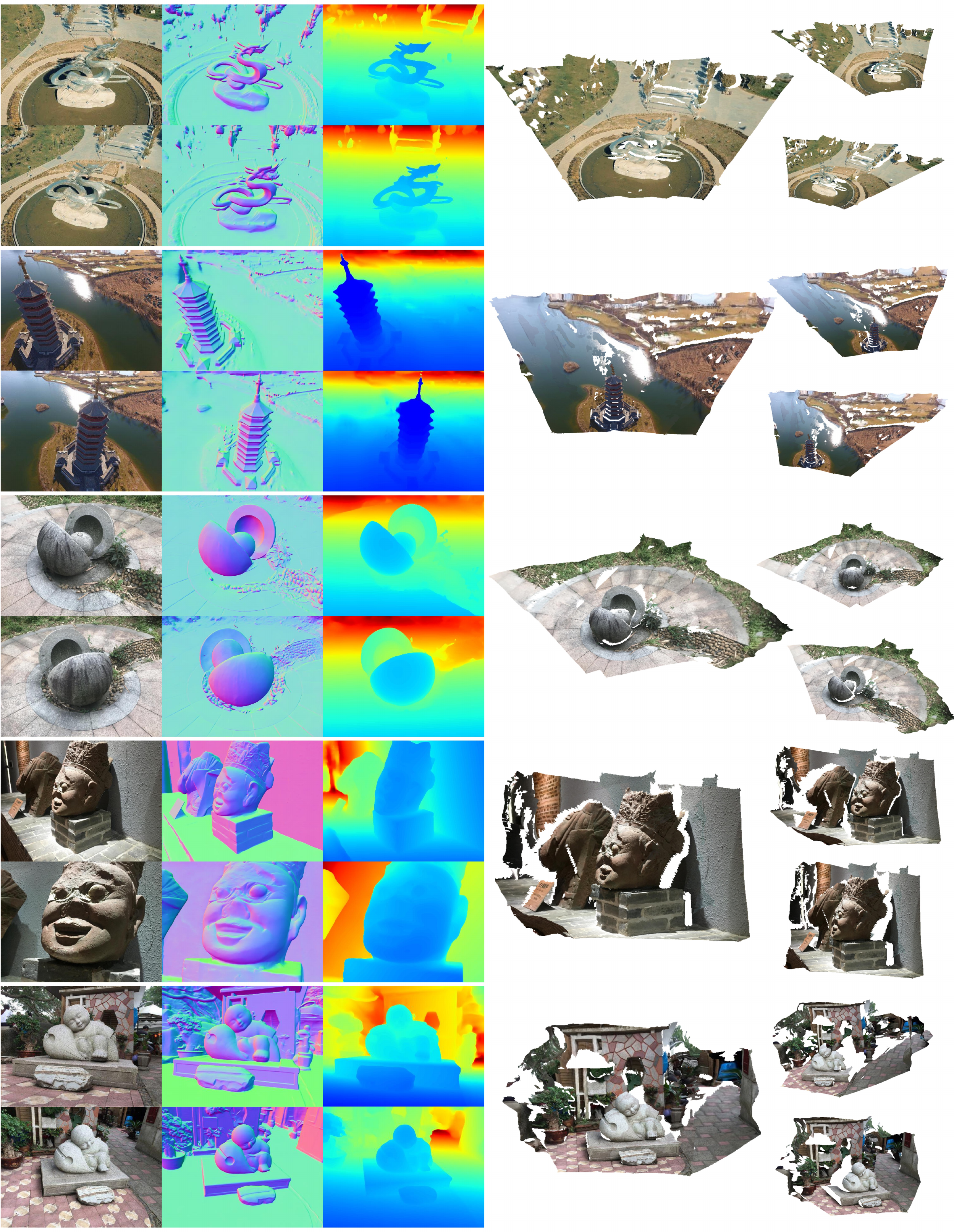}
    \caption{Additional visual results for unified geometric predictions of 2-view images inputs by our method.}
    \label{fig:more2}
\end{figure*}
\noindent\textbf{Datasets.}
To train the visual foundation model, we collect and reorganize a large-scale training dataset containing various data types. The dataset includes indoor scenes, outdoor scenes, and object-level data. It is noteworthy that the quality of training data has a substantial impact on model performance. We then make the most of high-quality synthetic data in the training process for more accurate and robust predictions. We divide all the data into three types based on their quality. Data of type A is collected from synthetic rendering process with the highest quality. Data of type B also originates from synthetic rendering, but they possess certain quality issues like insufficient resolution or absence of background or imprecise original 3D models, \textit{etc}. Data of type C is obtained from the real world using cameras and depth sensors. We also carefully allocate the proportions of each dataset to attain the optimal model training performance. We summarize and present this dataset information in Tab.~\ref{tab:dataset}.

\begin{table*}[h]
  \begin{tabular}{l|c|cccc|c|c}
    \toprule
    \multirow{2}*{\textbf{Dataset}} & \multirow{2}*{\textbf{Type}} & \multicolumn{4}{c|}{\textbf{Applied Losses}}& \textbf{Image} & \multirow{2}*{\textbf{Ratio}}\\

    ~ & ~ & $\mathcal{L}_{\text{pts\_loc}}$, $\mathcal{L}_{\text{pts\_glb}}$ & $\mathcal{L}_{\text{pts\_n}}$ & $\mathcal{L}_{\text{match}}$ & $\mathcal{L}_{\text{n}}$ & \textbf{Pairs} & ~ \\

    \midrule
    
    Hypersim\cite{roberts:2021} & A & \checkmark & \checkmark & \checkmark & \checkmark & 1.8M & 6.77\%\\

    UnrealStereo4K\cite{Tosi2021CVPR} & A & \checkmark & \checkmark & \checkmark & \checkmark & 0.9M & 6.77\%\\

    MatrixCity\cite{li2023matrixcity} & A & \checkmark & \checkmark & \checkmark & \checkmark & 0.7M & 6.77\%\\

    Infinigen\cite{infinigen2024indoors} & A & \checkmark & \checkmark & \checkmark & \checkmark & 2.8M & 6.77\%\\

    Behavior \cite{conf/corl/0002ZWGSMWLLSAH22} & A & \checkmark & \checkmark & \checkmark & \checkmark & 6.8M & 6.77\%\\

    Structure3D\cite{Structured3D} & A & \checkmark & \checkmark & \checkmark & \checkmark & 0.2M & 4.06\%\\

    GTASFM\cite{wang2020flow} & A & \checkmark & \checkmark & \checkmark & \checkmark & 0.2M & 13.53\%\\

    GTAV\cite{richter2016playing} & A & \checkmark & \checkmark & \checkmark & \checkmark & 0.6M & 13.53\%\\

    VirtualKitti\cite{gaidon2016virtual} & A & \checkmark & \checkmark & \checkmark & \checkmark & 4.0M & 13.53\%\\

    IRS\cite{wang2021irs} & A & \checkmark & \checkmark & \checkmark & \checkmark & 74K & 0.41\%\\

    UrbanSyn\cite{gomez2025} & A & \checkmark & \checkmark & \checkmark & \checkmark & 7.0K & 0.41\%\\

    Spring\cite{Mehl2023_Spring} & A & \checkmark & \checkmark & \checkmark & \checkmark & 10K & 0.41\%\\

    \midrule

    ScanNet++\cite{yeshwanthliu2023scannetpp} & B & \checkmark & \checkmark & \checkmark & \checkmark & 3.5M & 1.35\%\\

    ABO\cite{collins2022abo} & B & \checkmark & \checkmark & \checkmark & \checkmark & 2.0M & 1.35\%\\

    GObjaverseXL\cite{objaverseXL} & B & \checkmark & \checkmark & \checkmark & \checkmark & 6.8M & 1.35\%\\

    StaticThings3D\cite{staticthings3d} & B & \checkmark & \checkmark & \checkmark & \checkmark & 0.3M & 1.35\%\\

    BlendedMVS\cite{yao2020blendedmvs} & B & \checkmark & \checkmark & \checkmark & \checkmark & 1.1M & 1.35\%\\
    Habitat\cite{Savva_Kadian_Maksymets_Zhao_Wijmans_Jain_Straub_Liu_Koltun_Malik_et_al._2019} & B & \checkmark & \checkmark & \checkmark & \checkmark & 1.3M & 0.68\%\\

    Taskonomy\cite{zamir2018taskonomy} & B & \checkmark & \checkmark & \checkmark & \checkmark & 1.8M & 0.68\%\\
    ARKitScenes\cite{dehghan2021arkitscenes} & B & \checkmark & \checkmark & \checkmark & \checkmark & 2.2M & 0.68\%\\

    Tartanair\cite{tartanair2020iros} & B & \checkmark & \checkmark & \checkmark & \checkmark & 4.5M & 0.68\%\\

    Synthia\cite{Ros_2016_CVPR} & B & \checkmark & \checkmark & \checkmark & \checkmark & 2.6M & 0.68\%\\

    KenBurns\cite{Niklaus_TOG_2019} & B & \checkmark & \checkmark & \checkmark & \checkmark & 0.3M & 0.68\%\\

    \midrule

    MegaDepth\cite{MegaDepthLi18} & C & \checkmark & \checkmark & \checkmark & ~ & 1.8M & 1.35\%\\

    Waymo\cite{Sun_2020_CVPR} & C & \checkmark & \checkmark & \checkmark & ~ & 1.1M & 1.35\%\\
    Co3dv2\cite{Reizenstein_Shapovalov_Henzler_Sbordone_Labatut_Novotny_2021} & C & \checkmark & \checkmark & \checkmark & ~ & 1.2M & 1.35\%\\

    WildRGBD\cite{xia2024rgbd} & C & \checkmark & \checkmark & \checkmark & ~ & 1.1M & 1.35\%\\

    NianticMapFree\cite{arnold2022mapfree} & C & \checkmark & \checkmark & \checkmark & ~ & 3.7M & 1.35\%\\

    DL3DV\cite{ling2024dl3dv} & C & \checkmark & \checkmark & \checkmark & ~ & 1.2M & 1.35\%\\

    DIMLIndoor\cite{cho2019large} & C & \checkmark & \checkmark & \checkmark & ~ & 0.9M & 0.68\%\\

    ArgoverseStereo\cite{Argoverse} & C & \checkmark & \checkmark & \checkmark & ~ & 4.0K & 0.68\%\\

    \bottomrule
  \end{tabular}
    \caption{Training dataset information. We reorganize a large-scale training dataset and divide the data into three types based on their quality. We also showcase the training objectives we apply during training, the number of image pairs and the corresponding dataset ratio.}
  \label{tab:dataset}
\end{table*}

\noindent\textbf{{Training Details.}}
During our coarse-to-fine training, we first utilize all the images with 512 resolution and train our model for about two weeks in the coarse-stage training. Then we only utilize the images from type A dataset and a minor portion of type B dataset and set the image resolution to 1024 for the fine-stage training. We utilize 32 Nvidia A800 GPUs for both the coarse and fine stage training. As for model inference, our model only requires a single Nvidia RTX3090 GPU for 1024-resolution image inputs.

\section{Additional Visualization}
We provide additional unified geometric prediction results of monocular inputs in Fig.~\ref{fig:more1} and 2-view inputs in Fig.~\ref{fig:more2}. It can be seen that our model achieves robust and high-quality unified geometric predictions across several scenarios.

\section{Normal and Depth Comparison}
\begin{figure}
    \centering
    \includegraphics[width=\linewidth]{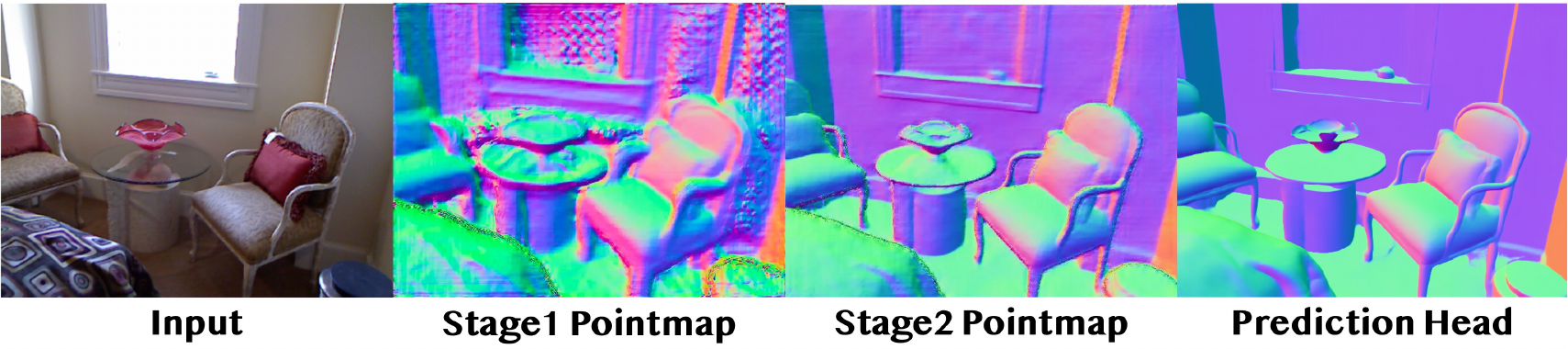}
    \caption{Normal prediction comparison of different training stages.}
    \label{fig:sup_nexp}
\end{figure}

Dens3R predicts robust and accurate normal and depth for various scenarios. As shown in Fig.~\ref{fig:sup_nexp}, we demonstrate that the intrinsic-invariant training assists the pointmap to capture the geometric information from normal. Then the normal prediction head further predicts sharper edges and more accurate results.

\begin{figure}
    \centering
    \includegraphics[width=\linewidth]{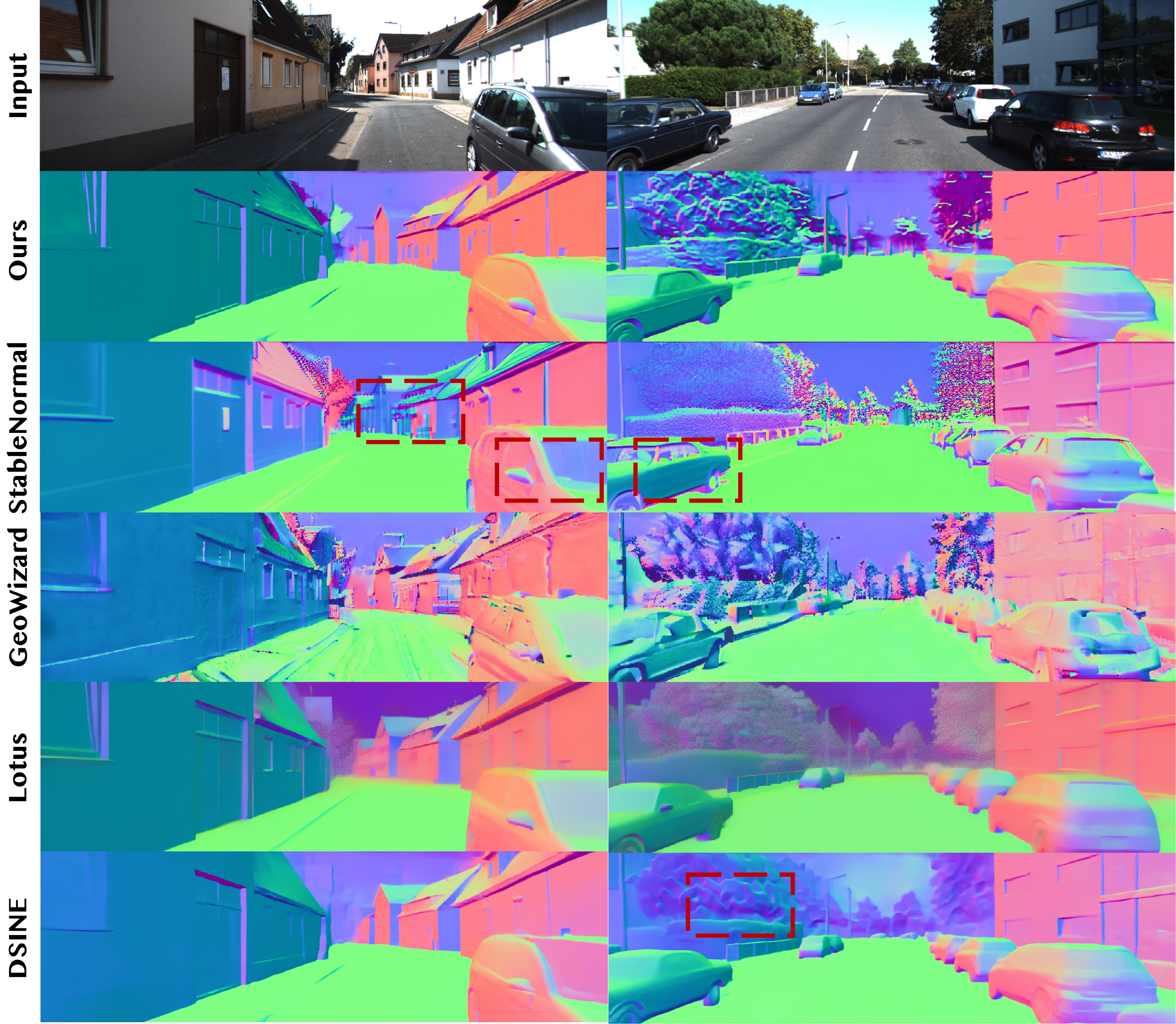}
    \caption{Normal comparison of Kitti dataset. We present more normal comparison of outdoor scenes, our method produces more accurate and sharper normals than previous methods.}
    \label{fig:kitti_normal}
\end{figure}

\begin{figure}
    \centering
    \includegraphics[width=\linewidth]{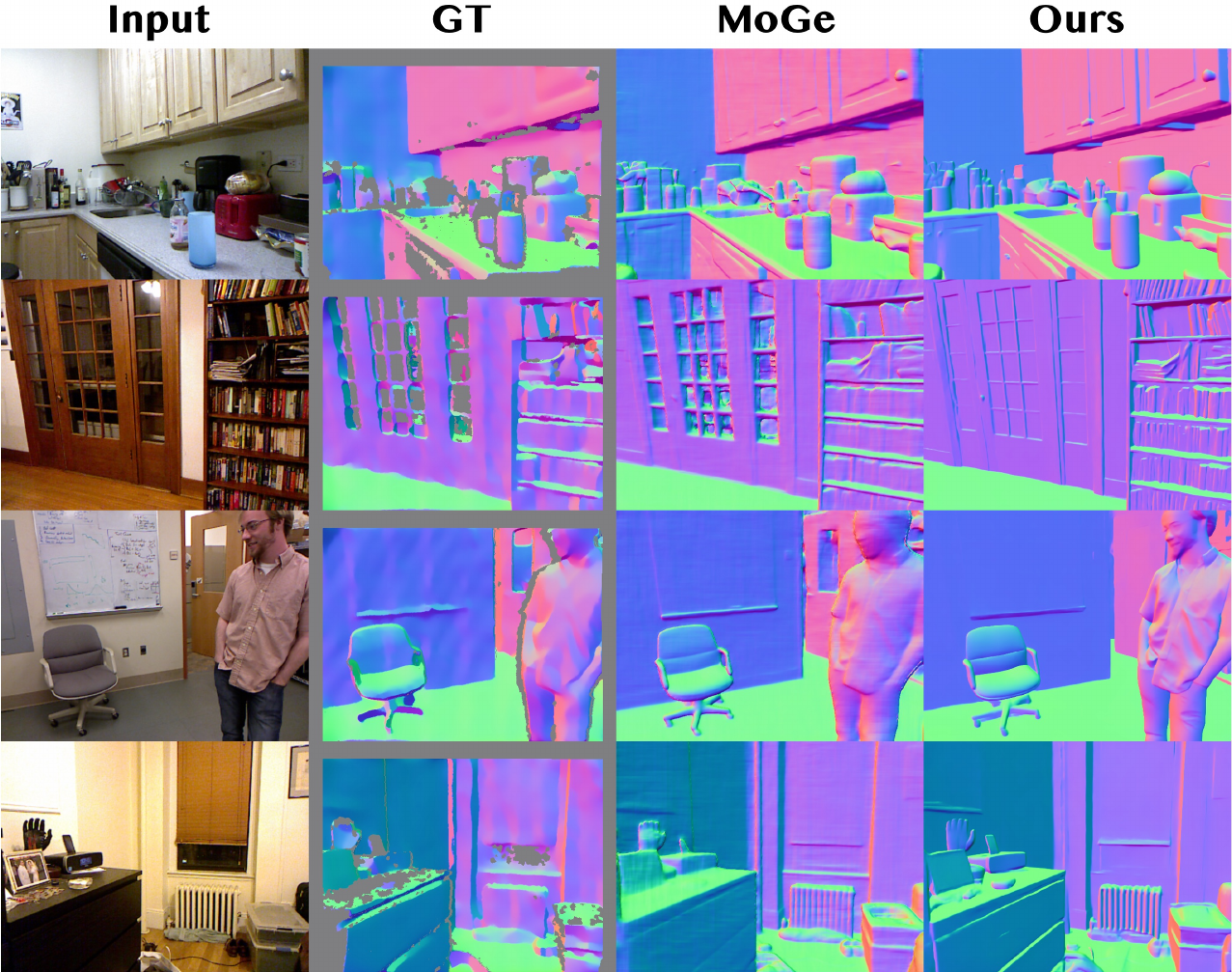}
    \caption{Normal comparison with MoGe. We provide more normal comparison with the normal maps derived from the depth map of MoGe~\cite{moge}. Our method yields sharper and more accurate predictions.}
    \label{fig:moge_ours}
\end{figure}

\begin{figure*}
    \centering
    \includegraphics[width=0.85\linewidth]{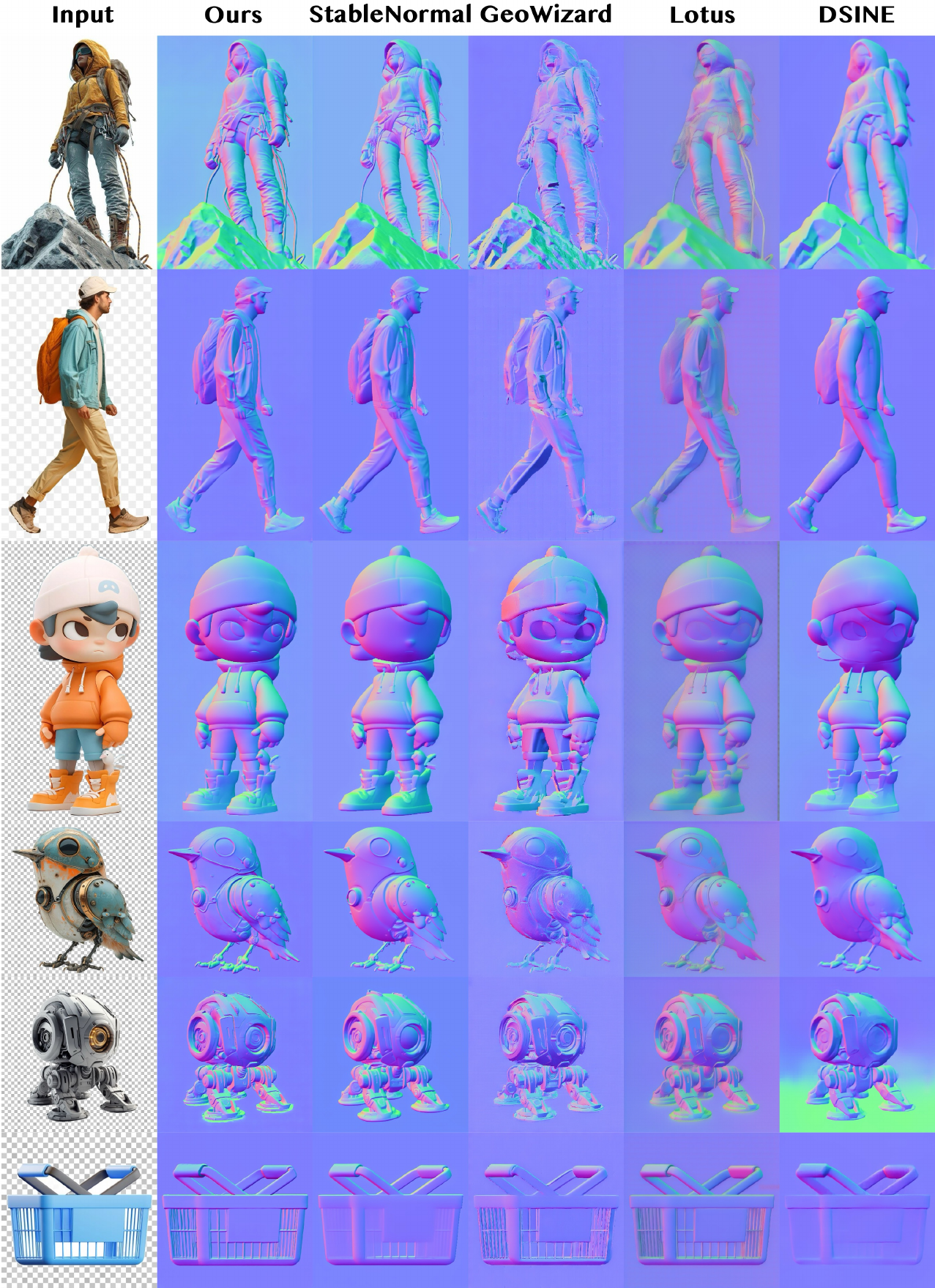}
    \caption{More qualitative comparison of normal map. We provide more normal comparison of both object-centric and human scenes. Dens3R is able to produce more accurate and sharper results}
    \label{fig:normal_sup2}
\end{figure*}

\begin{figure*}
    \centering
    \includegraphics[width=\linewidth]{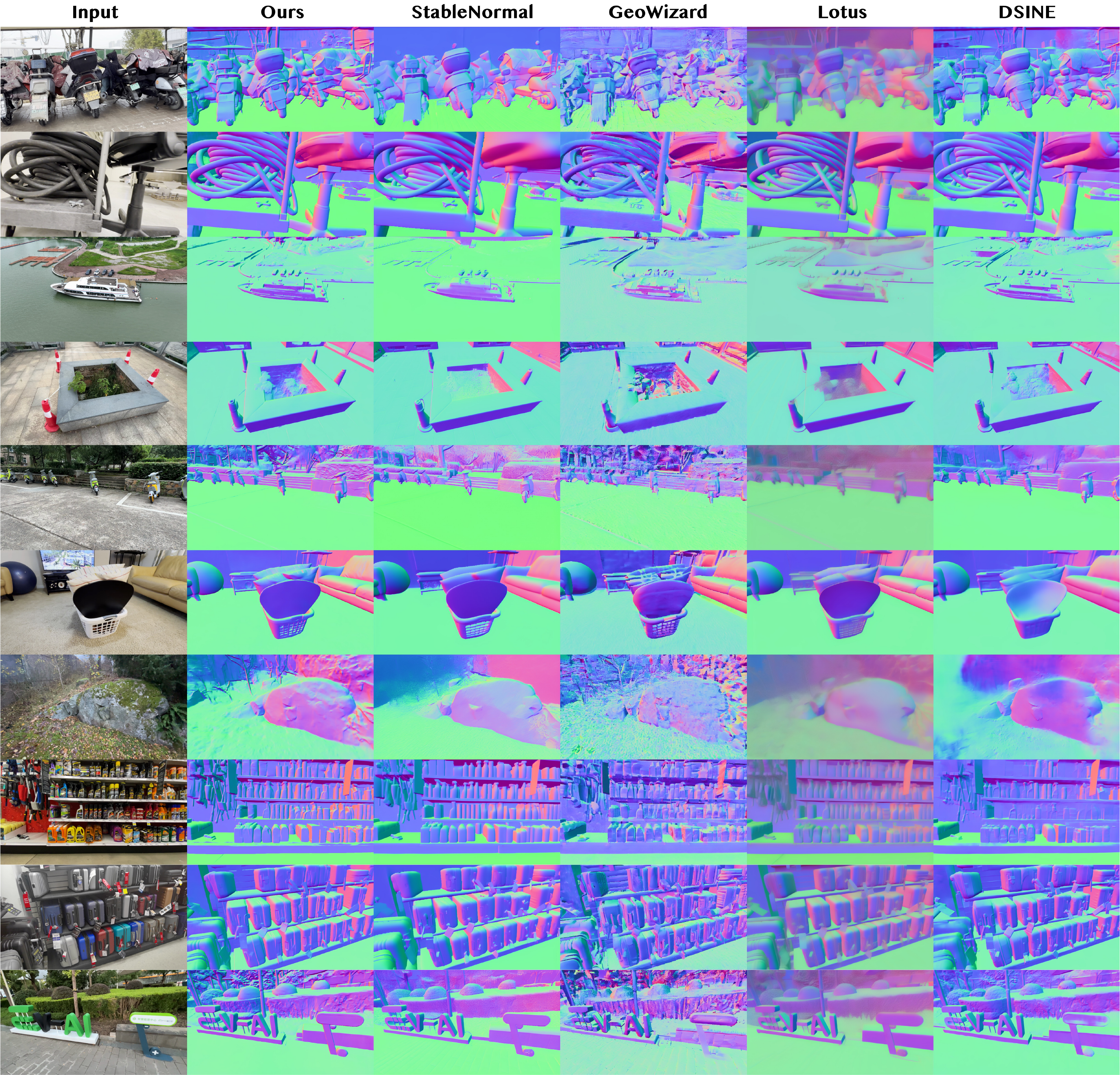}
    \caption{More qualitative comparison of normal map. We provide more normal comparison of both indoor and outdoor scenes. Dens3R is able to produce sharper and more accurate results and surpasses previous methods.}
    \label{fig:normal_sup}
\end{figure*}

We provide the normal prediction comparison of the Kitti dataset in Fig.~\ref{fig:kitti_normal}. It can be seen that our method generates the most accurate and sharp normals. We also provide more comparison of normal map prediction in Fig~\ref{fig:normal_sup2} using in-the-wild images and in Fig.~\ref{fig:normal_sup} using DL3DV dataset. It can be seen that our method predicts sharper and more accurate normal across various scenarios. We also compare our method with the normal map derived from the predicted depth map of MoGe~\cite{moge}, the results are shown in Fig.~\ref{fig:moge_ours}. It can be seen that Dens3R can handle normal predictions for reflective surfaces and accomplishes to generate richer details. We also provide the full quantitative comparison in Tab.~\ref{tab:full_n} and full ablation comparison in Tab.~\ref{tab:full_ab}, which are partly shown in Tab. 1 and Tab. 4 in the main paper correspondingly. 

\begin{figure*}
    \centering
    \includegraphics[width=\linewidth]{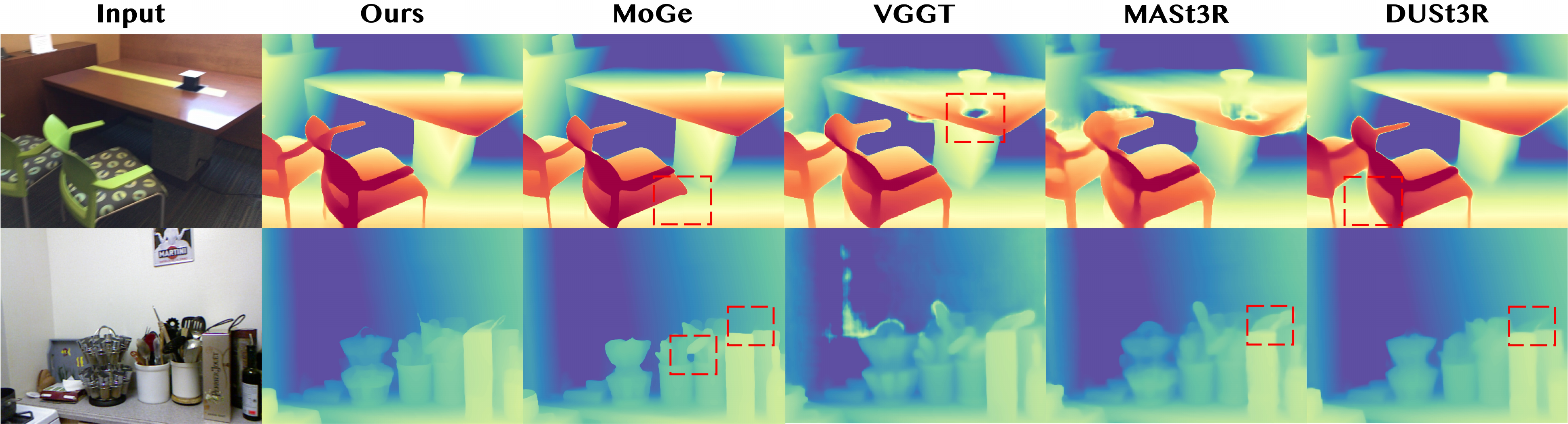}
    \caption{Additional depth comparison. We provide more depth comparison with previous methods and our method can predict more accurate and detailed results.}
    \label{fig:depth2}
\end{figure*}

\begin{figure*}
    \centering
    \includegraphics[width=\linewidth]{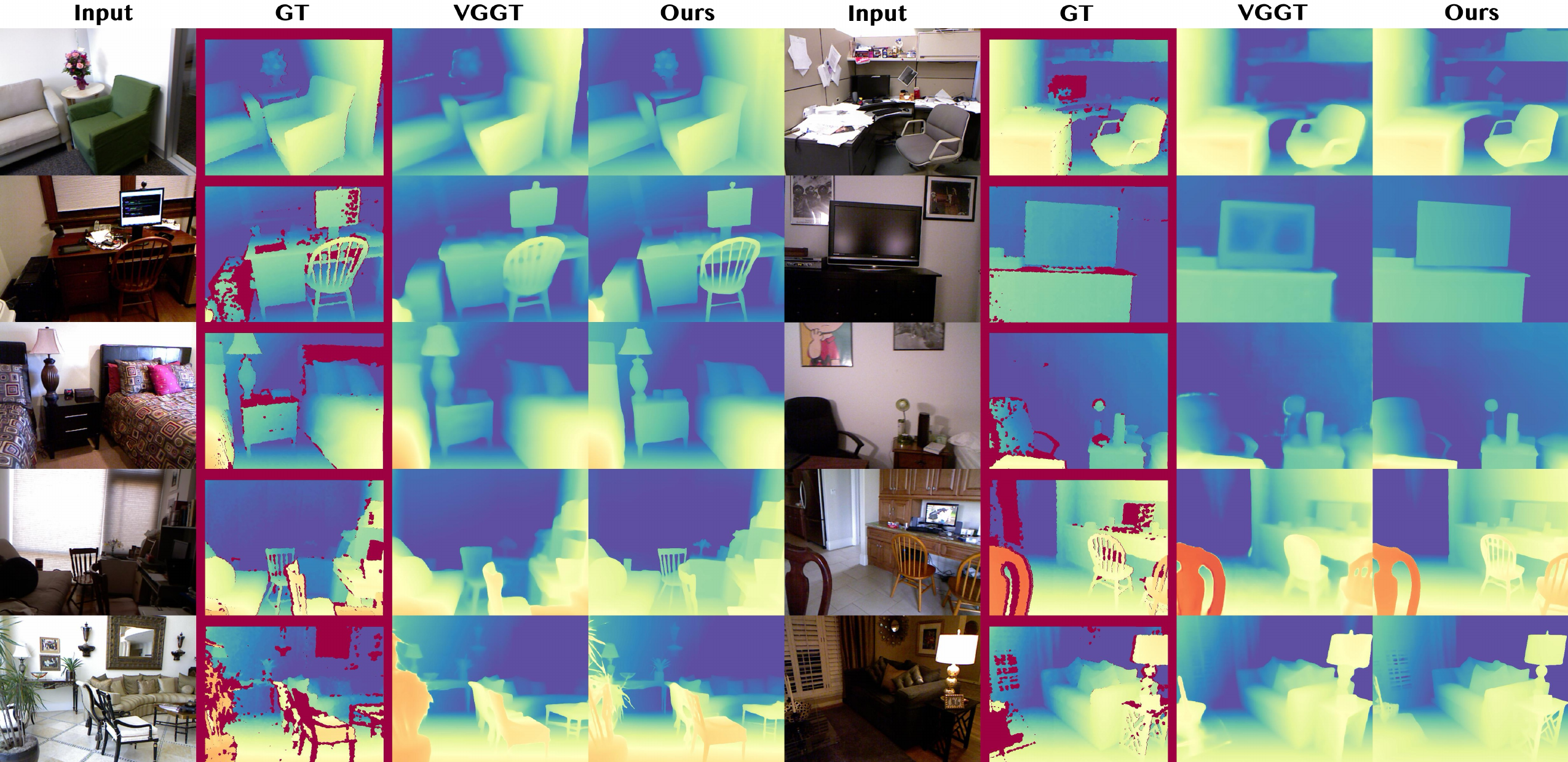}
    \caption{Additional depth comparison with VGGT. We compare our depth prediction results with VGGT and Dens3R demonstrates more robust and accurate predictions.}
    \label{fig:nyu_sup}
\end{figure*}

\begin{figure*}
    \centering
    \includegraphics[width=0.95\linewidth]{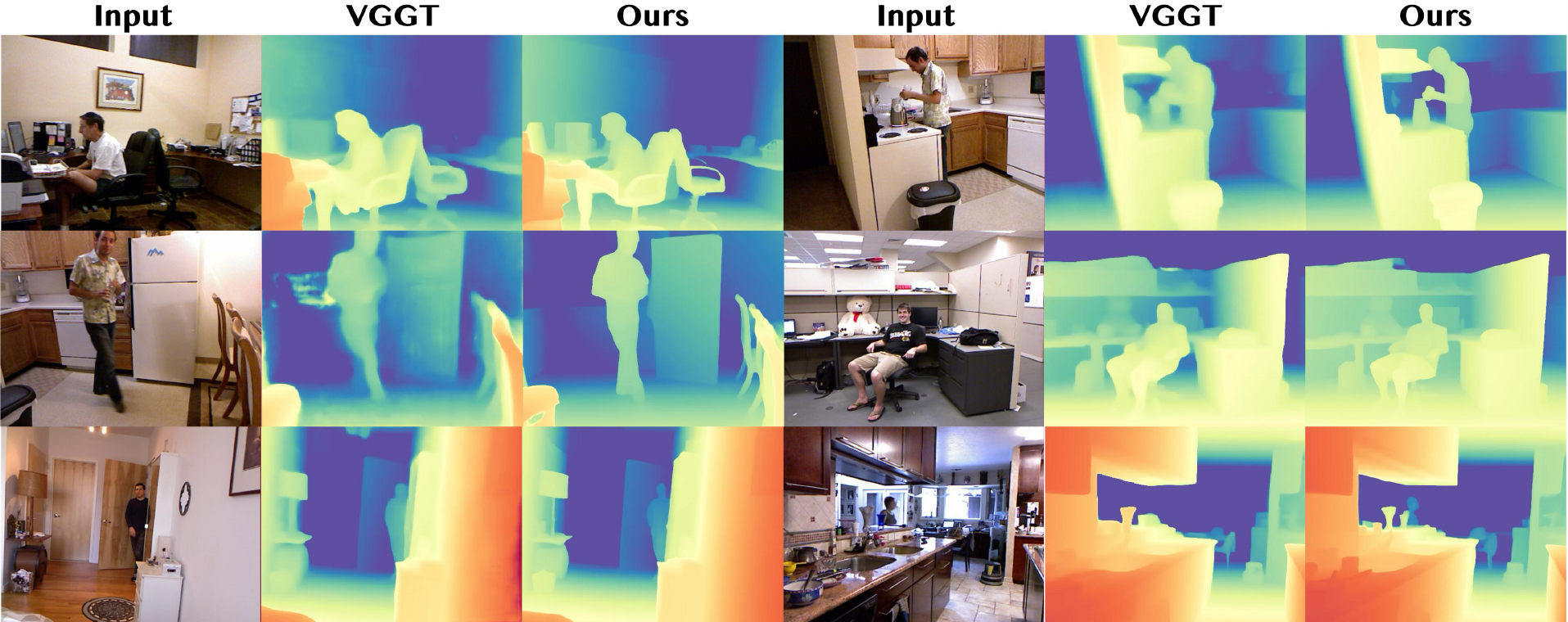}
    \caption{Additional depth comparison of indoor scenes with VGGT. Dens3R demonstrates more accurate results for human depth estimation.}
    \label{fig:person_sup}
\end{figure*}

\begin{figure*}
    \centering
    \includegraphics[width=0.95\linewidth]{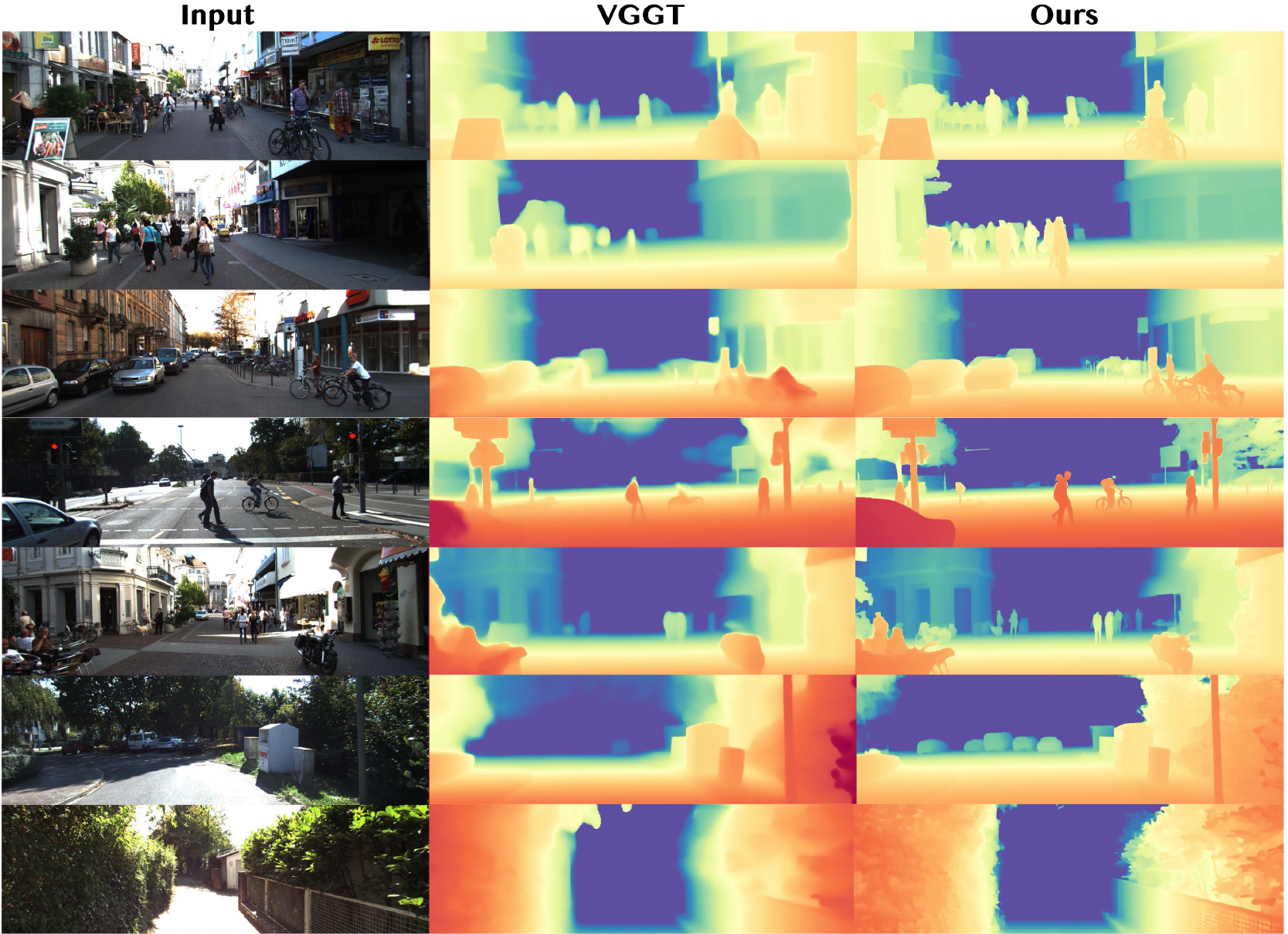}
    \caption{Additional depth comparison of outdoor scenes with VGGT. We compare our depth prediction results of autonomous driving dataset. Our methods achieves much more accurate predictions.}
    \label{fig:kitti_sup}
\end{figure*}

We provide the depth prediction comparison in Fig.~\ref{fig:depth2}, it can be seen that our method generates the most accurate depth maps even for reflective surfaces. Since VGGT~\cite{vggt} also predicts multiple quantities including depth and matching, we further compare our predicted depth map with VGGT. We demonstrate more accurate depth predictions of NYUv2 dataset in Fig.~\ref{fig:nyu_sup}. We also showcase the accurate prediction of both indoor scenes of NYUv2 dataset and outdoor scenes of Kitti dataset. It can be seen in Fig.~\ref{fig:person_sup} and Fig.~\ref{fig:kitti_sup} that our model also achieves accurate human depth estimation that can be further utilized for detection and autonomous driving.

\begin{table}
  \renewcommand\arraystretch{0.85}
  \begin{tabular}{c|cc|ccc}
    \toprule
    Method & Mean $\downarrow$ & Med $\downarrow$ & $\delta_{11.25 \degree}\uparrow $ & $\delta_{22.5 \degree}\uparrow $ & $\delta_{30 \degree}\uparrow $\\
    \midrule
    \multicolumn{6}{c}{NYUv2 (indoor)} \\
    \midrule
    DSINE & 18.6 & 9.9 & 56.1 & \cellcolor{second}76.9 & \cellcolor{second}82.6 \\
    Lotus & \cellcolor{second}17.5 & \cellcolor{second}8.6 & \cellcolor{second}58.7 & 76.4 & 82.0 \\
    GeoWizard & 20.4 & 11.9 & 47.0 & 73.8 & 80.8 \\
    StableNormal & 19.7 & 10.5 & 53.0 & 75.9 & 81.7 \\
    Ours & \cellcolor{best}{16.1} & \cellcolor{best}{7.4} & \cellcolor{best}{62.5} & \cellcolor{best}{78.8} & \cellcolor{best}{84.0}\\
    \midrule
    \multicolumn{6}{c}{ScanNet (indoor)} \\
    \midrule
    DSINE & 18.6 & 9.9 & 56.1 & 76.9 & 82 \\
    Lotus &  18.1 & \cellcolor{second}8.8 & \cellcolor{second}58.2 & 75.3 & 80.8 \\
    GeoWizard & 21.4 & 13.9 & 37.1 & 71.7 & 79.7 \\
    StableNormal & \cellcolor{second}18.1 & 10.1 & 56.0 & \cellcolor{best}78.8 & \cellcolor{best}84.1 \\
    Ours & \cellcolor{best}16.9 & \cellcolor{best}{7.1} & \cellcolor{best}{64.0} & \cellcolor{second}78.1 & \cellcolor{second}82.7\\
    \midrule
    \multicolumn{6}{c}{IBims-1 (indoor)} \\
    \midrule
    DSINE & 18.8 & 8.3 & 64.1 & 78.6 & 82.2 \\
    Lotus &  19.2 & \cellcolor{second}5.6 & 66.2 & 74.9 & 78.1 \\
    GeoWizard & 19.7 & 9.7 & 58.4 & 77.6 & 81.6 \\
    StableNormal & \cellcolor{second}17.2 & 8.1 & \cellcolor{second}66.7 & \cellcolor{best}81.1 & \cellcolor{best}84.6 \\
    Ours & \cellcolor{best}16.0 & \cellcolor{best}{4.3} & \cellcolor{best}{72.2} & \cellcolor{second}80.1 & \cellcolor{second}83.0\\
    \midrule
    
    \multicolumn{6}{c}{Sintel (outdoor)} \\
    \midrule
    DSINE & \cellcolor{second}34.9 & 28.1 & \cellcolor{second}21.5 & 41.5 & 52.7 \\
    Lotus & 35.7 & 28.0 & 20.5 & 41.8 & 52.8 \\
    GeoWizard & 41.6 & 34.3 & 11.8 & 31.8 & 43.9 \\
    StableNormal & 35.0 & \cellcolor{second}27.0 & 19.5 & \cellcolor{second}42.4 & \cellcolor{second}54.6 \\
    Ours & \cellcolor{best}{30.7} & \cellcolor{best}{21.4} & \cellcolor{best}{28.9} & \cellcolor{best}{51.9} & \cellcolor{best}{62.2}\\
    \midrule
    
    \multicolumn{6}{c}{DIODE-outdoor (outdoor)} \\
    \midrule
    DSINE & \cellcolor{second}22.0 & \cellcolor{second}14.5 & \cellcolor{second}39.6 & \cellcolor{second}67.5 & \cellcolor{second}75.4 \\
    Lotus & 24.7 & 15.9 & 32.9 & 63.9 & 71.9 \\ 
    GeoWizard & 27.0 & 19.8 & 24.0 & 56.6 & 68.9 \\
    StableNormal & 26.9 & 16.1 & 36.1 & 60.6 & 67.5 \\
    Ours & \cellcolor{best}20.8 & \cellcolor{best}12.8 & \cellcolor{best}43.0 & \cellcolor{best}70.7 & \cellcolor{best}77.0\\
   \bottomrule
\end{tabular}
  \caption{Full quantitative comparison of normal prediction. We report the mean and median angular errors with each cell colored to indicate the \colorbox{best}{best} and the \colorbox{second}{second}.}
  \label{tab:full_n}
\end{table}

\begin{table}[h]
    \centering
\resizebox{0.8\linewidth}{!}{%
 \begin{tabular}{c|c|ccc}
\toprule
Dataset                  & Metrics                           & w/o IIT & w/o C2F & Ours          \\ \hline
\multirow{2}{*}{NYUv2}  & Mean $\downarrow$                  & 17.8    & 17.6    & \textbf{16.1} \\
                         & $\delta_{11.25 \degree}\uparrow $ & 50.6    & 50.5    & \textbf{62.5} \\ \cline{1-5} 
\multirow{2}{*}{ScanNet} & Mean $\downarrow$                  & 18.6    & 17.8    & \textbf{16.9} \\
                         & $\delta_{11.25 \degree}\uparrow $ & 49.4    & 58.8    & \textbf{64.0} \\ \cline{1-5} 
\multirow{2}{*}{IBims}   & Mean $\downarrow$                  & 20.2    & 18.6    & \textbf{16.0} \\
                         & $\delta_{11.25 \degree}\uparrow $ & 56.8    & 63.9    & \textbf{72.2} \\ \cline{1-5} 
\multirow{2}{*}{Sintel}  & Mean $\downarrow$                  & 35.9    & 35.8    & \textbf{30.7} \\
                         & $\delta_{11.25 \degree}\uparrow $ & 18.9    & 22.3    & \textbf{28.9} \\ 
                         \cline{1-5}
 \multirow{2}{*}{DIODE-outdoor}  & Mean $\downarrow$                  & 23.5    & 21.6   & \textbf{20.8} \\
 & $\delta_{11.25 \degree}\uparrow $ & 33.7    & 40.2    & \textbf{43.0} \\ 
 \bottomrule
\end{tabular}}
    \caption{Full normal quantitative metrics for ablation. We demonstrate that both the intrinsic-invariant training and coarse-to-fine strategy contributes to accurate normal predictions.}
  \label{tab:full_ab}
\end{table}

\section{Camera Pose Estimation Comparison}
Dens3R can also perform accurate camera pose estimation through a single feed-forward pass. We conduct extended experiments to demonstrate its accuracy. We utilize the map-free benchmark~\cite{arnold2022mapfree} following the MASt3R protocol~\cite{mast3r}, which is a challenging dataset aiming at localizing the camera in metric space given a single reference image without any map. We present the camera pose estimation (Map-free relocalization) comparison in Tab.~\ref{tab:camera_map_relloc}. It can be seen that Dens3R outperforms previous methods in nearly all the metrics, demonstrating highly accurate camera pose estimation results.

\begin{table*}
\centering
\begin{tabular}{cccccccc}
\toprule
Method & Reproj. Error $\downarrow$ & Precision $\uparrow$ & AUC $\uparrow$   & Median Error (m) $\downarrow$ & Median Error (°) $\downarrow$ & Pose Precision $\uparrow$ & Pose AUC $\uparrow$ \\
\midrule
DUSt3R & 125.8 px      & 45.2\%    & 0.704 & 1.10 m & 9.4° & 17.0\%         & 0.344    \\
MASt3R & 57.2 px       & 75.9\%    & \cellcolor{second}0.934 & 0.46 m & \cellcolor{best}{3.0°}       & 51.7\%         & \cellcolor{second}{0.746}    \\
VGGT   & \cellcolor{second}{48.8 px}       & \cellcolor{second}{78.9\%}    & 0.789 & \cellcolor{second}{0.36 m} & 3.6°       & \cellcolor{second}{57.7\%}         & 0.577    \\
Ours   & \cellcolor{best}{30.4 px}       & \cellcolor{best}{82.1\%}    & \cellcolor{best}0.944 & \cellcolor{best}{0.24 m} & \cellcolor{second}{3.4°}       & \cellcolor{best}65.5\%         & \cellcolor{best}0.852   \\
\bottomrule
\end{tabular}
\caption{Camera pose estimation results of the Map-free dataset. We report the metrics with each cell colored to indicate the \colorbox{best}{best} and the \colorbox{second}{second}.}
\label{tab:camera_map_relloc}
\end{table*}

\section{Image Matching Visualization Results}
For image-matching, apart from the ZEB dataset, we also provide the quantitative comparison of the Scannet-1500 dataset in Tab.~\ref{tab:matching_result_scannet} and the MegaDepth-1500 dataset in Tab.~\ref{tab:matching_result_megadepth}. The comparisons on the ScanNet-1500 and the MegaDepth-1500 benchmarks further demonstrate our superior performance over pervious DUSt3R-based method MASt3R~\cite{mast3r} and VGGT~\cite{vggt}.
\begin{table}
    \centering
    \begin{tabular}{c|ccc}
    \toprule
         Method &  AUC$@5\degree\uparrow$ & AUC$@10\degree\uparrow$ & AUC$@20\degree\uparrow$\\
         \midrule
          ROMA  & 31.8 & 53.4 &  70.9 \\
          VGGT & 33.9 & 55.2 & 73.4 \\
          MASt3R & \cellcolor{second}62.4 & \cellcolor{second}77.4 & \cellcolor{second}86.9 \\
          Ours & \cellcolor{best}65.6 & \cellcolor{best}80.3 & \cellcolor{best}89.2 \\
          \bottomrule
    \end{tabular}
    \caption{Two-view matching comparison on ScanNet-1500 Dataset. We report the AUC values with each cell colored to indicate the \colorbox{best}{best} and the \colorbox{second}{second}. Our method achieves state-of-the-art for two-view matching, surpassing all the previous methods.}
  \label{tab:matching_result_scannet}
\end{table}
\begin{table}
    \centering
    \begin{tabular}{c|ccc}
    \toprule
         Method &  AUC$@5\degree\uparrow$ & AUC$@10\degree\uparrow$ & AUC$@20\degree\uparrow$\\
         \midrule
          SP+SG & 42.2 & 61.2 & 76.0 \\
          SP+LG & 49.9 & 67.0 & 80.1 \\
          LoFTR & 52.8 & 69.2 & 81.2 \\
          MASt3R & \cellcolor{second}73.3 & \cellcolor{second}84.1 & \cellcolor{second}90.9 \\
          Ours & \cellcolor{best}73.9 & \cellcolor{best}84.4 & \cellcolor{best}91.2 \\
          \bottomrule
    \end{tabular}
    \caption{Two-view matching comparison on MegaDepth-1500 Dataset. We report the AUC values with each cell colored to indicate the \colorbox{best}{best} and the \colorbox{second}{second}. Our method also achieves state-of-the-art for the two-view matching using the MegaDepth-1500 Dataset.}
  \label{tab:matching_result_megadepth}
\end{table}

\begin{figure*}
    \centering
    \includegraphics[width=0.8\linewidth]{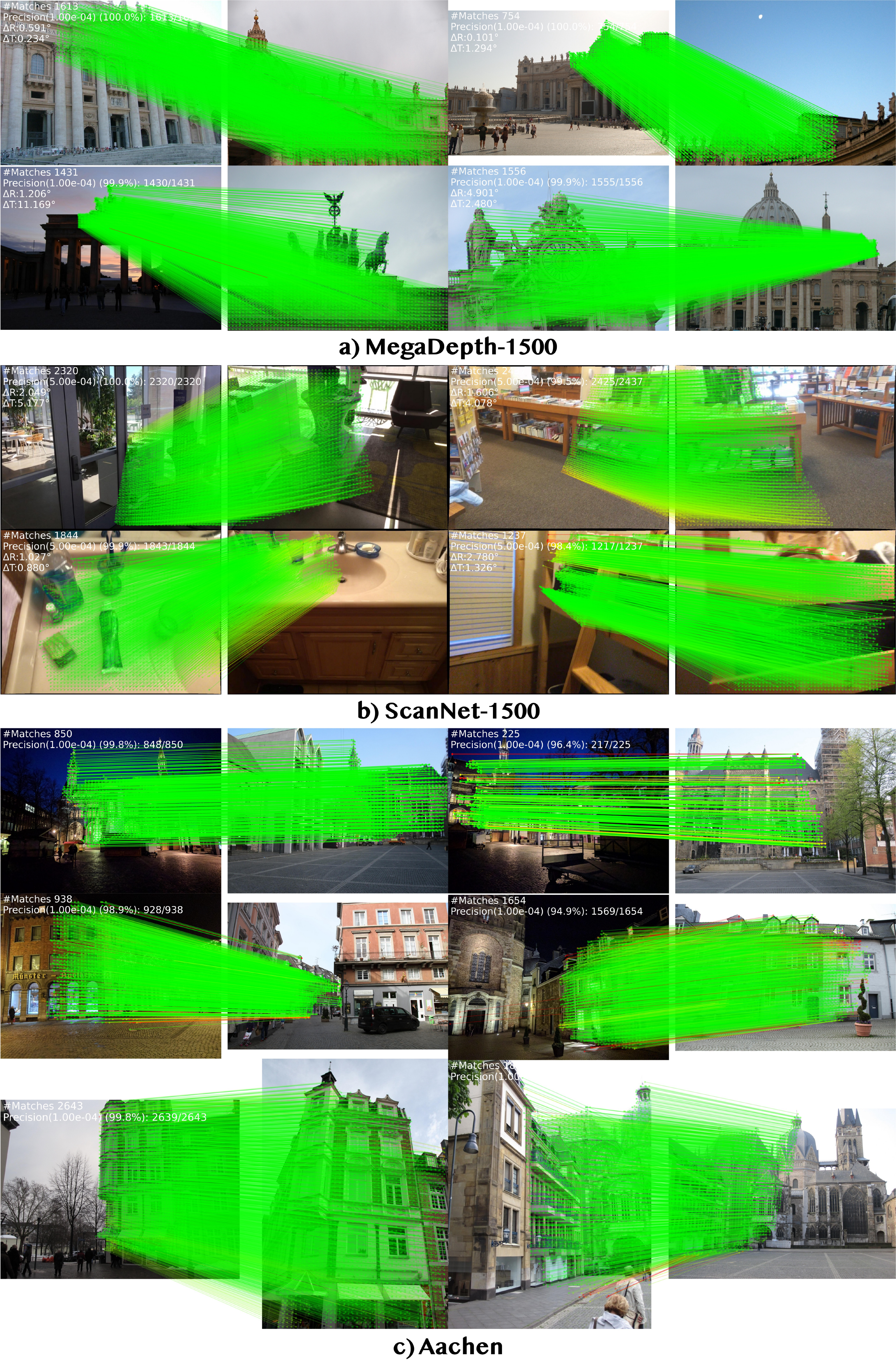}
    \caption{Image-matching visualization. We provide the visualization of our dense and accurate image-matching results. The accuracy is also presented on the upper left of each matching image-pair.}
    \label{fig:matching}
\end{figure*}

We also demonstrate our dense and accurate matching results of several challenging cases in Fig.~\ref{fig:matching}. We visualize the matching results from MegaDepth-1500 dataset, ScanNet-1500 dataset and Aachen dataset. It can be seen that our method can handle image-matching for \textbf{1)} inputs with different lighting conditions, \textbf{2)} inputs taken from different views with large-angle difference, \textbf{3)} inputs with small overlapping regions. The accuracy of our method is shown on the upper left of each matching image-pair.

\section{High-Resolution Inference Comparison}
\begin{figure*}
    \centering
    \includegraphics[width=\linewidth]{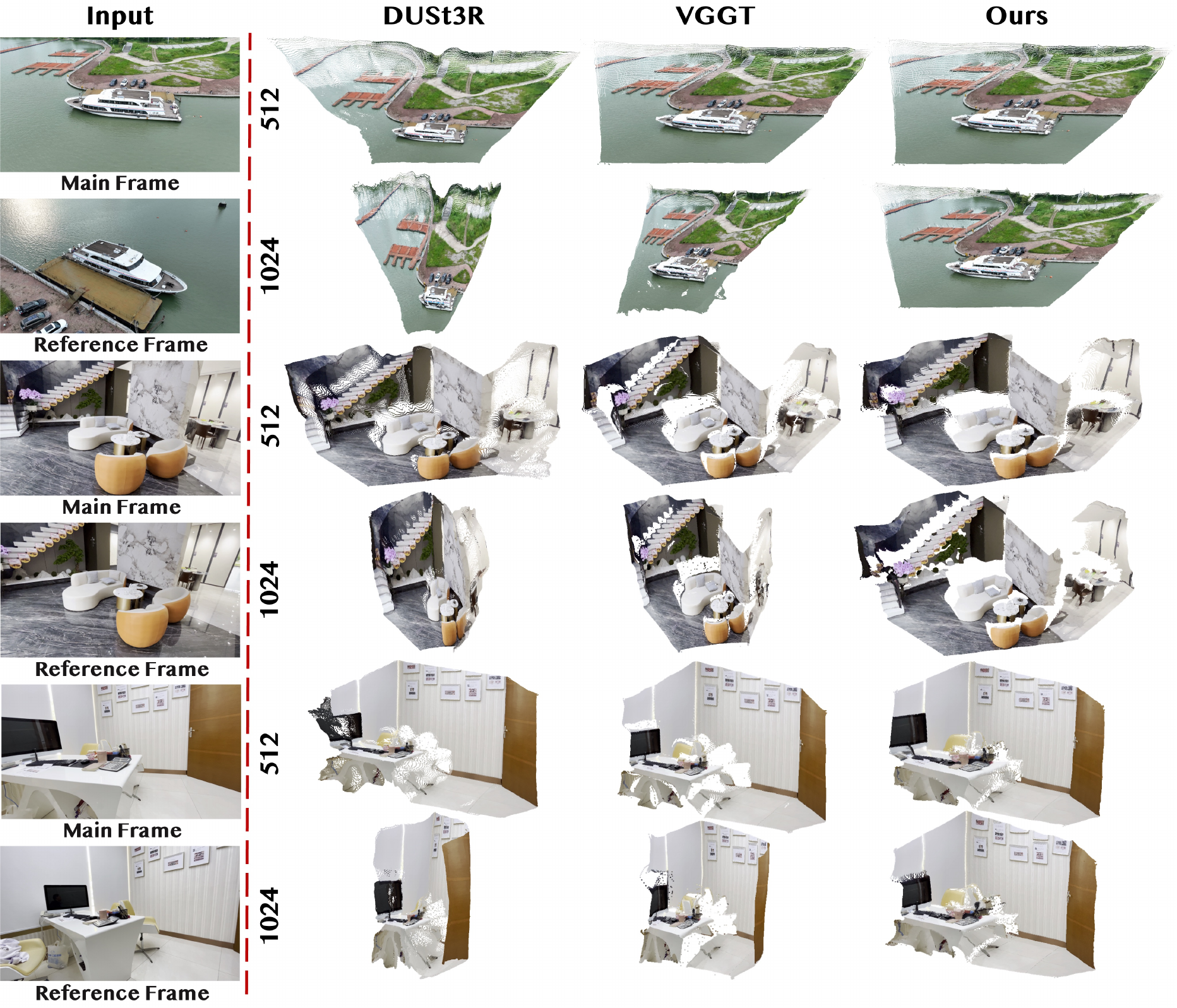}
    \caption{Additional high-resolution inference comparison. We provide more high-resolution inference results to demonstrate the effectiveness of the proposed position-interpolated rotary positional encoding. We present the pointmap of the main frame and our method accomplishes to prevent the degeneration problem that occured in previous methods.}
    \label{fig:hq_sup}
\end{figure*}

We showcase more comparison of high-resolution inputs with DUSt3R \cite{dust3r} and VGGT~\cite{vggt} in Fig.~\ref{fig:hq_sup}. It can be seen that our method can handle higher-resolution inputs without causing degenerated predictions like previous methods with our proposed position-interpolated rotary positional encoding.

\section{Shared Encoder-Decoder Backbone Ablation}
Dens3R employs a dense visual transformer backbone designed to capture spatial relationships across viewpoints and capture the global 3D geometric information of scenes. Different from previous methods, both the encoder and decoder components in our architecture share weights. The comparison of the network parameters and the memory cost is shown in Tab.~\ref{tab:network_params}. Since our model deals with more 3D quantities than previous methods, the framework initially requires a higher memory cost. Employing the shared encoder-decoder structure also resolves this issue, reducing the memory cost and network parameters without losing the prediction quality.

\begin{table}
    \centering
    \begin{tabular}{c|ccc}
    \toprule
         Setting &  Compute Cost & Memory Cost & Network Params \\
         \midrule
         w/o Shared & 1.362 TFlops & 4.6 GB & 737.591 M \\
          w/ Shared & 1.362 TFlops & \textbf{4.1 GB} & \textbf{624.152 M}\\
          \bottomrule
    \end{tabular}
    \caption{Ablation on shared encoder-decoder structure. We conduct experiments for both of the model on image pairs with 512 resolution. With the shared encoder-decoder structure, our model yields lower memory cost and less network parameters.}
  \label{tab:network_params}
\end{table}

\section{Multi-view Image Inputs}
\begin{figure*}
    \centering
    \includegraphics[width=\linewidth]{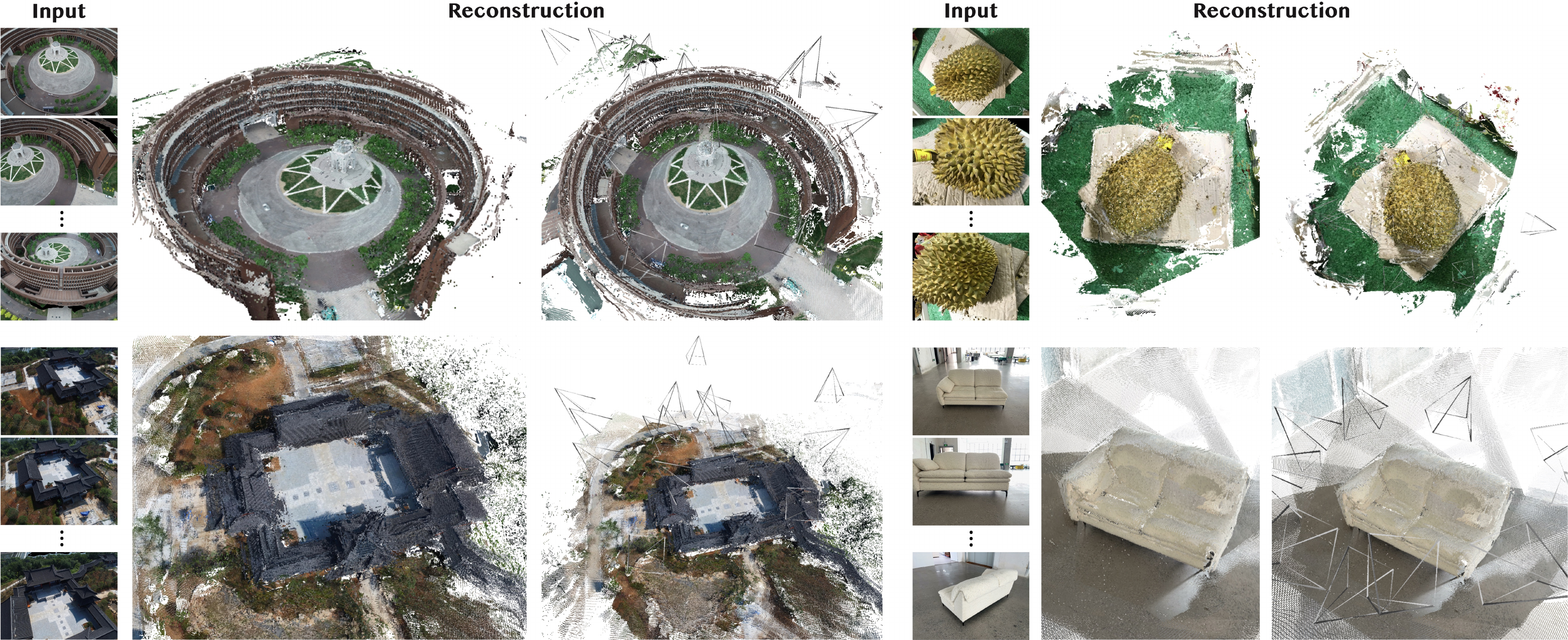}
    \caption{Multi-view reconstruction results. We demonstrate high-quality 3D reconstruction for various scenarios without known camera poses. The predicted camera poses are also shown on the right of the reconstruction results.}
    \label{fig:mv}
\end{figure*}

Apart from multi-resolution image inputs, our model also design a simple yet effective post-processing pipeline and supports multi-view inputs. The results are demonstrate in Fig.~\ref{fig:mv}. It can be seen that our method enables high-quality 3D reconstruction even without known camera poses. In addition, the predictions of Dens3R can be used to initialize robust and accurate 3D reconstructions by integrating the predicted attributes with modern structure-from-motion (SfM) pipelines such as Glomap. 

\section{Limitation}
\begin{figure}
    \centering
    \includegraphics[width=\linewidth]{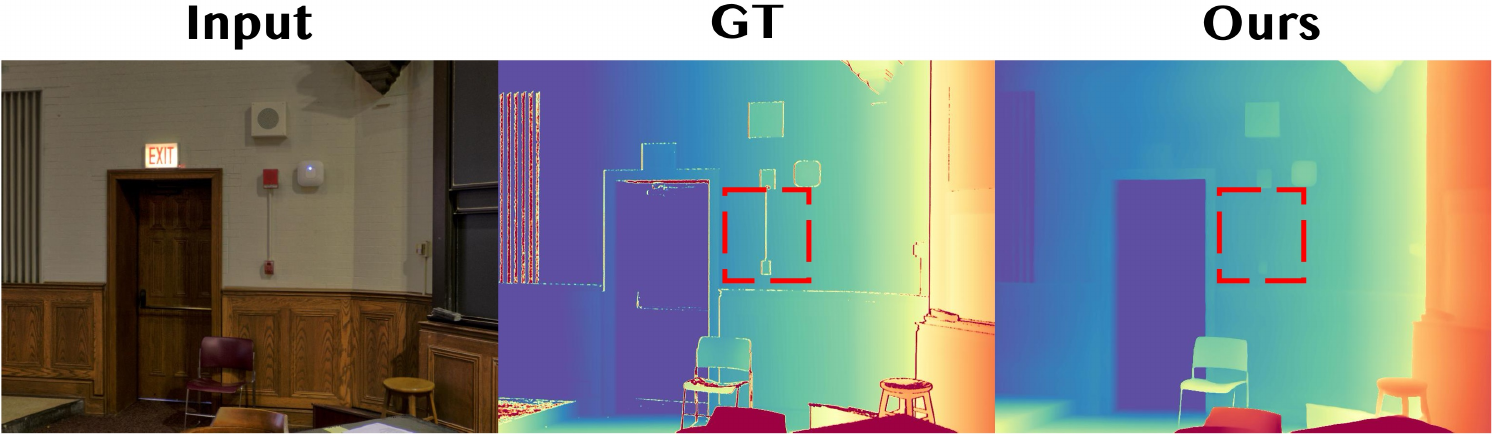}
    \caption{Limitations. Despite that our method outperforms previous methods in geometric predictions, the prediction quality for thin structures still require further improvement.}
    \label{fig:thin}
\end{figure}

Although Dens3R outperforms previous methods in geometric predictions, predicting accurate results for inputs with thin structures remains a significant challenge. Restricted by the network’s limited capacity and the presence of noisy training data, our method may predict inaccurate results for these inputs. As shown in Fig.~\ref{fig:thin}, the prediction quality for thin structures still require further improvement.

\clearpage

\end{document}